\RequirePackage{amsthm}
\documentclass[pdflatex,sn-mathphys-num]{sn-jnl}%

\usepackage{graphicx}%
\usepackage{multirow}%
\usepackage{amsmath,amssymb,amsfonts}%
\usepackage[title]{appendix}%
\usepackage{xcolor}%
\usepackage{textcomp}%
\usepackage{manyfoot}%
\usepackage{booktabs}%
\usepackage{algorithm}%
\usepackage{algorithmicx}%
\usepackage{algpseudocode}%
\usepackage{listings}%
\usepackage{hyperref}
\usepackage{cleveref}
\usepackage{lmodern}

\usepackage{underscore}

\DeclareMathOperator*{\Var}{\text{Var}}

\newcommand{\PDC}{\text{PDC}}
\newcommand{\TU}{\text{TU}}
\newcommand{\EU}{\text{EU}}
\newcommand{\AU}{\text{AU}}

\newcommand{\given}{\, | \,}
\newcommand{\fromto}{\longrightarrow}

\graphicspath{{img/}}
\Crefname{equation}{Eq.}{Eqs.}
\Crefname{figure}{Fig.}{Figs.}
\Crefname{tabular}{Tab.}{Tabs.}
\Crefname{table}{Tab.}{Tabs.}
\crefname{equation}{Eq.}{Eqs.}
\crefname{figure}{Fig.}{Figs.}
\crefname{tabular}{Tab.}{Tabs.}
\crefname{table}{Tab.}{Tabs.}
\crefname{section}{Section}{Sections}
\crefname{annex}{Annex}{Annexes}

\theoremstyle{thmstyleone}%

\theoremstyle{thmstyletwo}%

\theoremstyle{thmstylethree}%

\usepackage{eso-pic}
\begin{document}
\AddToShipoutPictureFG*{\AtPageLowerLeft{\put(\LenToUnit{\dimexpr(\paperwidth-\textwidth)/2\relax},\LenToUnit{0.55in}){\parbox{\textwidth}{\footnotesize\centering Preprint (version submitted before peer review). Published in \emph{Machine Learning} 115, 12 (2026), \url{https://doi.org/10.1007/s10994-025-06949-0}}}}}

\title[Uncertainty Quantification in Pairwise Difference Learning]{Uncertainty Quantification in Pairwise Difference Learning for Classification}

\author*[1,2]{\fnm{Mohamed Karim} \sur{Belaid}}\email{karim.belaid@idiada.com}

\author[2]{\fnm{Maximilian} \sur{Rabus}}\email{maximilian.rabus2@porsche.de}

\author[3,4,5]{\fnm{Eyke} \sur{H\"ullermeier}}\email{eyke@lmu.de}

\affil*[1]{\orgname{IDIADA Fahrzeugtechnik GmbH}, \orgaddress{\city{Munich}, \country{Germany}}}

\affil[2]{\orgname{Dr. Ing. h.c. F. Porsche AG}, \orgaddress{\city{Stuttgart}, \country{Germany}}}

\affil[3]{\orgdiv{Institute of Informatics}, \orgname{University of Munich (LMU)}, \orgaddress{\country{Germany}}}

\affil[4]{\orgname{Munich Center for Machine Learning (MCML)}, \orgaddress{\country{Munich, Germany}}}

\affil[5]{\orgname{German Centre for Artificial Intelligence (DFKL, DSA)}, \orgaddress{\country{Kaiserslautern, Germany}}}

\abstract{
Instead of learning a mapping from instances to outcomes in the standard way, the key idea of pairwise difference learning (PDL) is to learn a function that takes two instances as input and predicts the difference between the respective outcomes. Given a function of this kind, predictions for a query instance are derived from every training example and then averaged. We propose a version of PDL for the task of classification and propose a suitable meta-learning technique for this purpose. Moreover, we address the problem of uncertainty quantification, i.e., of representing and quantifying the uncertainty of a PDL classifier's predictions in a faithful manner. Building on recent work on the decomposition of total uncertainty into aleatoric and epistemic components, we propose different ways of extending corresponding measures to the setting of PDL, depending on how and to what extent uncertainty is represented by the underlying base learner. In large-scale empirical studies, we analyze both the predictive accuracy and the uncertainty awareness of our PDL classifier.
}


\maketitle

\section{Introduction}

Pairwise Difference Learning (PDL) is a versatile meta-learning technique that solves supervised learning tasks by learning to predict the difference $\Delta(x, x') = f(x) - f(x')$ between the outcomes of pairs of inputs\,---\, rather than outcomes of individual inputs directly, as ``conventional'' learners do. Point predictions can then be obtained from observed data $(x',f(x'))$ thanks to the relationship $f(x) = f(x') + \Delta(x, x')$. Averaging over many predictions of that kind turns out to significantly enhance predictive accuracy, especially in small data regimes \cite{wetzel2020twin,tynes2021pairwise,xie2022automated,corbara2023same,belaid2024pairwise}. This is also because, by pairing training examples, the number of training examples is inflated to the square of the original size (albeit at the cost of losing statistical independence).

PDL has originally been introduced for regression tasks, i.e., for predicting (differences between) continuous outcomes. This paper is an extension of our recent work \cite{belaid2024pairwise}, in which we generalized PDL to the case of multiclass classification, essentially specifying the ``difference'' between outcomes in terms of class (in)equality.
Like in the case of regression, significantly improved accuracy (compared to conventional classification) could be shown in many cases.
However, while predictive accuracy is critical, capturing a model's reliability in its predictions is not less important, particularly for high-stakes decisions in domains such as healthcare, autonomous systems, and risk management. In this article, we therefore shift our focus towards uncertainty quantification (UQ) within the PDL framework and extend the latter by methods for representing and quantifying aleatoric uncertainty (which arises from inherent randomness and noise in the data) as well as epistemic uncertainty (which reflects the learner's lack of knowledge due to limited or incomplete training data).
Our key contributions are as follows:
\begin{itemize}
    \item To realize the idea of PDL for the task of classification, we propose a meta-learning technique that transforms any multiclass classification problem into a single binary problem (\cref{sec:pdc}).
    \item We establish a comprehensive taxonomy of uncertainty representations that aligns with the PDL framework, incorporating measures for total, aleatoric, and epistemic uncertainty (\cref{sec:method}).

    \item We conduct extensive empirical evaluations on both real-world benchmark datasets and controlled synthetic datasets, analyzing both the predictive accuracy and the uncertainty awareness of our PDL classifier. Our results also offer insights into the interactions between the choice of the base learner, the uncertainty representation, and the UQ measures (\cref{sec:evaluation}).

    \item We provide a theoretical and empirical analysis of UQ within the context of the pairwise transformation induced by PDL, critically assessing the suitability and effectiveness of various UQ methods. Based on this analysis, we propose a principled selection of UQ techniques best suited for PDL classifiers (\cref{sec:discussion}).

    \item We make a Python implementation of the PDL classifier available\footnote{Link: \url{https://github.com/Karim-53/pdll}}. Our ``pairwise difference learning library'' (pdll) adheres to the Scikit-learn standards and ensures compatibility with any Sklearn ML model.

\end{itemize}

\section{Related Work}

Pairwise Difference Learning (PDL) has evolved into a versatile AI framework. Initially developed for regression tasks, PDL is a meta-algorithm that can be applied on top of neural networks \cite{wetzel2020twin} or classical machine learning regressors \cite{tynes2021pairwise,xie2022automated}. Subsequent research extended PDL to classification, demonstrating its utility in domains such as natural language processing \cite{corbara2023same} and tabular data \cite{belaid2024pairwise}.

PDL has demonstrated remarkable performance in various domains, outperforming popular machine learning algorithms. Benchmark evaluations have used more than 400 supervised learning datasets \cite{belaid2024pairwise,belaid2024weighted,wetzel2023how}. In particular, PDL was applied to large-scale datasets, showcasing its ability to handle high-dimensional data, such as the Fe-DFTB dataset, with 11,656 features and 12,726 samples \cite{tynes2021pairwise}.

Beyond benchmark datasets, PDL has achieved notable results in chemistry \cite{fralish2023deepdelta,spiers2023physicochemical}, image processing \cite{hu2023exploring}, quantum reactions \cite{chen2023benchmark}, and drug activity ranking \cite{wang2023extrapolation}. This foundation has evolved PDL into a versatile tool with applications in active and contrastive learning. Moreover, learning differences is central to related domains such as Siamese networks and OpenAI's CLIP model \cite{radford2021learning}.

Uncertainty quantification (UQ) measures the confidence level of machine learning models in their predictions, assessing their reliability and trustworthiness.
In deep learning, uncertainty can be estimated by incorporating specific neurons to output uncertainty estimations \cite{sensoy2018evidential}. Alternatively, model-agnostic approaches perturb training data points or model parameters to infer prediction distributions and derive uncertainty estimates \cite{sale2023secondorder,yu2022uncertainty}.
A more convenient method involves exploiting intermediate predictions or results within ensemble models or PDL frameworks \cite{tynes2021pairwise,belaid2024pairwise}. Some authors have explored combining multiple perturbation methods, but these approaches remain insufficiently mature for real-world applications \cite{valdenegro2024unified}.

UQ research distinguishes between aleatoric uncertainty, arising from inherent data randomness, and epistemic uncertainty, resulting from a model's incomplete knowledge of the true data-generating process \cite{hullermeier2021aleatoric}. This distinction is critical in applications such as medical decision-making, where understanding uncertainty sources enhances model accuracy, robustness, and trustworthiness \cite{lohr2024aleatoric}.
UQ also contributed significantly in the field of active learning, Virtual screening, and Drug Discovery \cite{yu2022uncertainty}.
The decomposition of total uncertainty introduces additional complexity in selecting high-performing models and identifying an effective UQ measure \cite{wimmer2023quantifying}. For instance, \citeauthor{hou2024decomposing} demonstrated that methods for estimating epistemic uncertainty can also estimate aleatoric uncertainty in large language models (LLMs).

However, UQ has been sparsely explored within the context of PDL.
\citeauthor{kramer2014matched} emphasized reducing aleatoric uncertainty caused by experimental measures on models accepting pairs of inputs.
In regression, \citeauthor{jimenez-luna2019deltadelta} introduced a PDL regressor model called "DeltaDelta", which replaces the absolute activity of each ligand with its difference from a known reference compound, using the variance across multiple predictions as an uncertainty estimate \cite{jimenez-luna2019deltadelta}.
\citeauthor{tynes2021pairwise} transferred this idea to molecular property prediction and observed similar results. They used variation over pair predictions to measure the total uncertainty (though without decomposing it into aleatoric and epistemic components).

\section{Pairwise Difference Classifier}\label{sec:pdc}

We follow the mathematical notation introduced by \citeauthor{belaid2024pairwise} \cite{belaid2024pairwise,belaid2024weighted}.
Consider a conventional supervised learning scenario for classification with training data
$$
\mathcal{D} = \{ (x_i , y_i ) \}_{i=1}^N \subset \mathbb{R}^d \times \mathcal{Y} \, ,
$$
where each data point consists of a feature vector $x_i \in \mathbb{R}^d$ and its corresponding discrete label $y \in \mathcal{Y} = \{ 1, \ldots, K \}$. The data is assumed to be independently and identically distributed (i.i.d.) according to an unknown joint probability distribution. The objective is to obtain a classifier
that yields accurate predictions by minimizing a prespecified loss function (in expectation).

The Pairwise Difference Classifier (PDC) transforms the original dataset as follows:
\begin{equation}\label{eq:ttd}
\mathcal{D}_{pair} = \big\{ (z_{i,j} , y_{i,j}) \, \vert \, 1 \leq i , j \leq N \big\} \, ,
\end{equation}
where each pair $(x_i,x_j)$ is mapped to a joint feature representation $z_{i,j} = \phi(x_i,x_j)$, and the label $y_{i,j}$ is defined as
\begin{equation}\label{eq:Kronecker_delta}
y_{i,j}  = \begin{cases}
0 & \text{for } y_i \neq y_j \\
1 & \text{for } y_i = y_j
\end{cases} \, .
\end{equation}
Thus, $\mathcal{D}_{pair}$ suggests the task of learning a binary classifier  that predicts whether a given pair of inputs shares the same class label or not. More generally, we consider probabilistic predictors \begin{equation}\label{eq:ppm}
\gamma: \, \mathbb{R}^{d} \times \mathbb{R}^{d} \rightarrow [0, 1 ] \, ,
\end{equation}
where $\gamma(x,x')$ is the probability that the class labels of $x$ and $x'$ coincide.
To this end, $\gamma$ is modeled as $\gamma(x_i,x_j) = h(\phi(x_i,x_j))$,
where $h \in \mathcal{H}$ is a probabilistic binary classifier from a suitable hypothesis space $\mathcal{H}$ trained on the transformed data (\ref{eq:ttd}). %

Given a query $x_q$, PDC estimates the probability of class labels $y \in \mathcal{Y}$ as follows: Considering each training example $(x_i, y_i)$ as a piece of evidence for the unknown class $y_q$, the semantics of the above prediction suggests that the probability of the event $y_q = y_i$ is given by $\gamma(x_i , x_j)$. More formally, $P(E) = \gamma(x_q , x_i)$, where $E$ denotes the event $y_q = y_i$ (and hence $P(\neg E) = 1 - \gamma(x_q , x_i)$). Let $p$ denote the prior distribution on the class labels $\mathcal{Y}$ (which can easily be estimated by relative frequencies on the training data). This distribution is then updated by conditioning it on the (uncertain) event $E$, which yields the following posterior suggested by $(x_i, y_i)$:
\begin{equation}\label{eq:pdc_p_post}
p_{post,i} (y \given x_q ) = \left\{ \begin{array}{cl}
 \gamma(x_q , x_i) & \text{ if } y = y_i \\[2mm]
 \dfrac{p(y) \cdot (1 - \gamma(x_q , x_i))}{1- p(y_i)} & \text{ otherwise}
\end{array} \right.
\end{equation}
Thus, the (posterior) probability of $y_i$ is fixed to $\gamma(x_q , x_i)$, and all other probabilities are rescaled in a proportional way to guarantee that the sum of posterior probabilities adds to 1.
Finally, we average the evidence from all training examples to obtain
\begin{equation}\label{eq:bma}
p_{post}(y \given x_q ) = \frac{1}{N} \sum_{i=1}^N p_{post,i} (y \given x_q ) \, .
\end{equation}
Note that PDL combines properties of both ``eager'' and ``lazy'' learning \cite{aha_ll}. Like standard inductive learning methods, it uses the entire training data to induce a (global) predictive model (\ref{eq:ppm}), and like (lazy) instance-based learning methods, it uses the individual training examples to obtain local predictions for a new query point. In a sense, it can be seen as a sophisticated extension of nearest neighbor classification, using a learned instead of a predefined transfer function $\gamma(x,x') \equiv 1$ (nearest neighbor classification, each training instance simply predicts it's ``own'' label, i.e., answers ``yes'' to the question of class equality) and aggregating over the entire data instead of restricting to local neighborhoods.

Also note that the averaging (\ref{eq:bma}) over training examples is very much akin to averaging predictions in ensemble learning\,---\,except that conventional methods such as Random Forests or bagging aggregate predictions over multiple (weak) models (each produced through randomization of the training data data or the learning process), whereas PDL averages predictions of the same (difference) model (\ref{eq:ppm}) ``anchored'' on different training examples.\footnote{Interestingly, the predictor $\gamma$ in (\ref{eq:ppm}) can be realized as a model ensemble, thereby effectively combining both types of ensembling.} Nevertheless, the hope in both cases is to improve accuracy by reducing variance through averaging. Besides, as already mentioned, ensemble predictions are commonly used for UQ, with the basic intuition that diverse predictions suggest a high level of (epistemic) uncertainty \cite{laks_sa17}. Similarly, the individual predictions $p_{post,i} (y)$ in (\ref{eq:bma}) provide a natural point of departure for UQ in PDL.

\section{Uncertainty Quantification for PDL Classifiers} \label{sec:method}

The common assumption in machine learning is that the ground-truth dependence between inputs $x$ and outcomes $y$ is stochastic and characterized in terms of conditional probability distributions $p(\cdot \given x)$, i.e., $p(y \given x)$ denotes the probability to observe label $y \in \mathcal{Y}$ as an outcome given $x \in \mathcal{X}$ as input information. Since $\mathcal{Y}$ is finite, a distribution $p(\cdot \given x)$ can be identified with a vector $\theta = (\theta_1, \dots, \theta_K)^\top \in \Delta_K$, and the set $\mathbb{P}(\mathcal{Y})$ of all probability distributions on $\mathcal{Y}$ with the $(K-1)$-dimensional probability simplex $\Delta_K$. We denote the ground-truth distribution $p(\cdot \given x)$ by $\theta^*_{x}$ and a prediction of this distribution by $\theta_{x}$. In the literature on UQ in machine learning, one commonly distinguishes three levels of ``uncertainty awareness'' of predictive models \cite{hullermeier2021aleatoric}:
\begin{itemize}
\item Deterministic predictors $\mathcal{X} \fromto \mathcal{Y}$ deliver pure class predictions and are not uncertainty aware at all; their predictions can also be seen as degenerate probabilistic predictions $\theta \in \{0,1\}^K \cap \Delta_K$ assigning probability 1 to one class and 0 to all others (a Dirac distribution).
\item Probabilistic predictors $\mathcal{X} \fromto \Delta_K$ deliver (first-order) probability distributions and are thus able to represent aleatoric uncertainty.
\item Bayesian (second-order probabilistic) predictors $\mathcal{X} \fromto \mathbb{P}(\Delta_K)$ deliver second-order distributions as predictions, i.e., probability distributions over probability distributions, which allows them to express uncertainty about the ``right'' prediction $\theta_x \in \Delta_K$; second-order predictors are thus able to capture epistemic uncertainty.
\end{itemize}
In PDL, we assumed the pairwise predictor $\gamma$, which we subsequently refer to as the base learner, to be a first-order predictor. In principle, however, the base learner can be of any type. Moreover, instead of merely averaging the predictions of the base learner according to (\ref{eq:bma}), its predictions can also be ``integrated'' in order to reach a higher-level of uncertainty awareness. For example, from $N$ first order distributions $p_{post,i} (\cdot \given x )$, a second-order distribution can be estimated, e.g., in the form of a relative frequency distribution (histogram) or by estimating the parameters of a parametric family of distributions.

\begin{table}[ht]
\centering
\caption{Different types of predictors obtained from different types of base learners in PDL through averaging or integrating their predictions.}
\label{tab:pdcp}
\begin{tabular}{lll}
\hline
Base learner $\gamma$ & Averaging & Integrating \\
  \hline
Deterministic & First-order predictor & (Second-order predictor) \\
              & $\text{pdc}: \mathcal{X} \to \Delta_K$ &  \\[2mm]
Probabilistic  & First-order predictor  & Second-order predictor \\
              & $\text{pdc}: \mathcal{X} \to \Delta_K$ & $\text{PDC}: \mathcal{X} \to \mathbb{P}(\Delta_K)$ \\[2mm]
Bayesian     & Second-order predictor & (Third-order predictor) \\
             & $\text{PDC}: \mathcal{X} \to \mathbb{P}(\Delta_K)$ & \\
\hline
\end{tabular}
\end{table}

Table \ref{tab:pdcp} summarizes the different types of predictors that are obtained by either averaging or integrating the predictions of the base learners\,---\, we denote first-order predictors by $\text{pdc}$ (lower-case) and second-order predictors by $\text{PDC}$ (upper-case), i.e., $\theta_x = \text{pdc}(x) \in \Delta_K$ is a first-order prediction produced for input $x$, and  $Q_x = \text{PDC}(x) \in \mathbb{P}(\Delta_K)$ is a second-order distribution (assigning probability mass/density $Q_x(\theta)$ to first-order distribution $\theta$). There are two cases (in brackets) that do not appear to be very meaningful: Integrating deterministic predictions yields a second-order distribution that assign probability mass only to degenerate first-order distributions (upper-right case in the table). Moreover, integrating second-order predictions yields a third-order prediction that will be difficult to interpret semantically and hard to handle computationally (lower-right case). We elaborate on the different types of predictors further below and illustrate them by means of examples.

\subsection{Deterministic Predictors}

Deterministic base learners output ``hard'' class predictions, i.e., degenerate probabilistic predictions,  %
and therefore do not capture any uncertainty. However, within the PDL framework, aggregating such deterministic predictions across multiple input pairs using (\ref{eq:pdc_p_post}) produces a non-degenerate probability distribution \(\theta \in \Delta_K\), effectively transforming the deterministic base learner into a probabilistic predictor capable of capturing first-order uncertainty.

\begin{figure}[htb]
    \centering
    \includegraphics[width=0.24\linewidth]{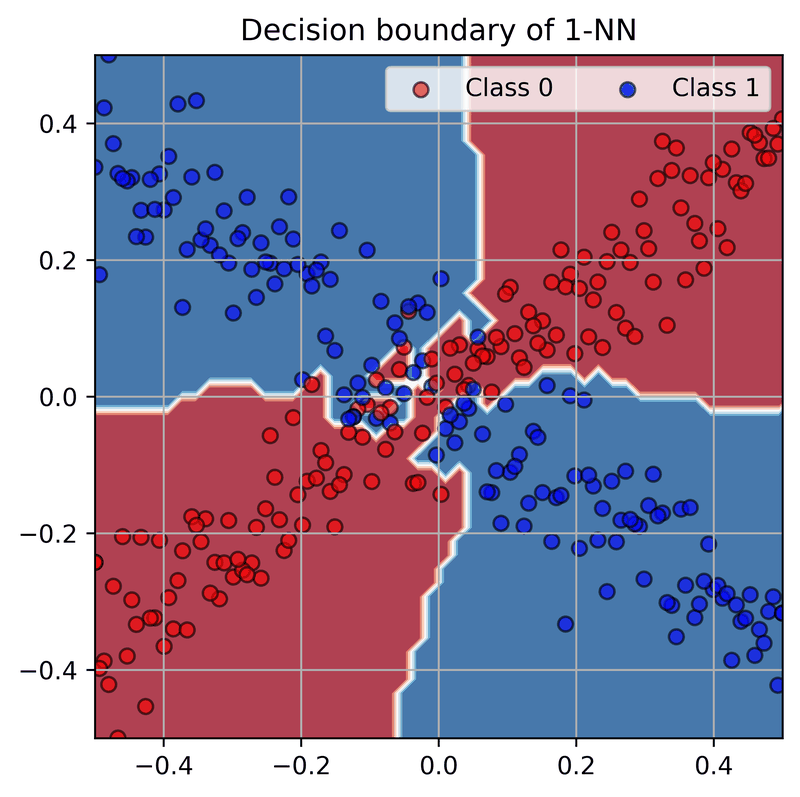}
    \includegraphics[width=0.24\linewidth]{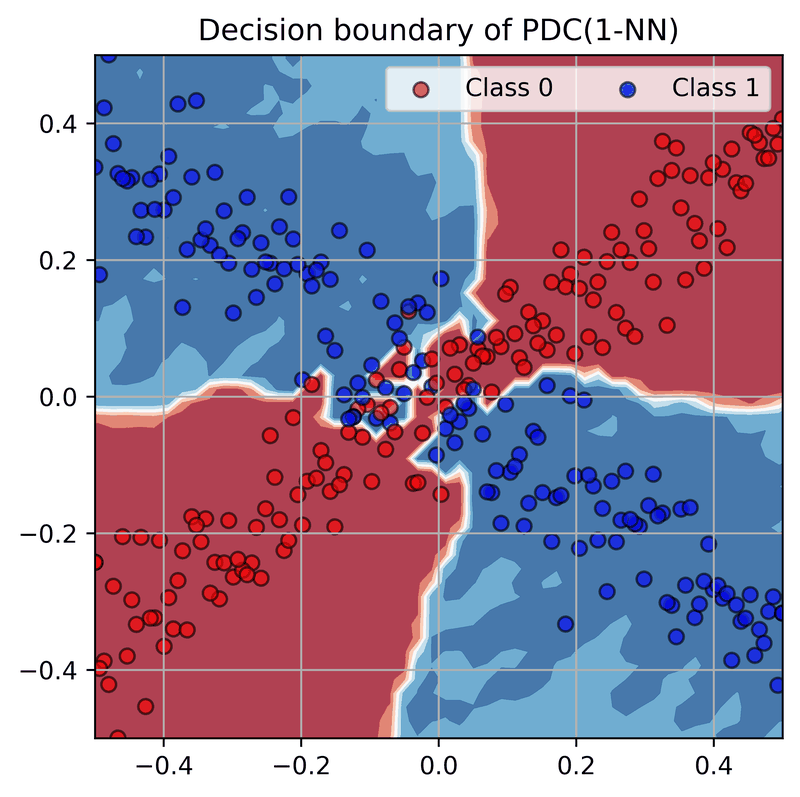}
    \includegraphics[width=0.24\linewidth]{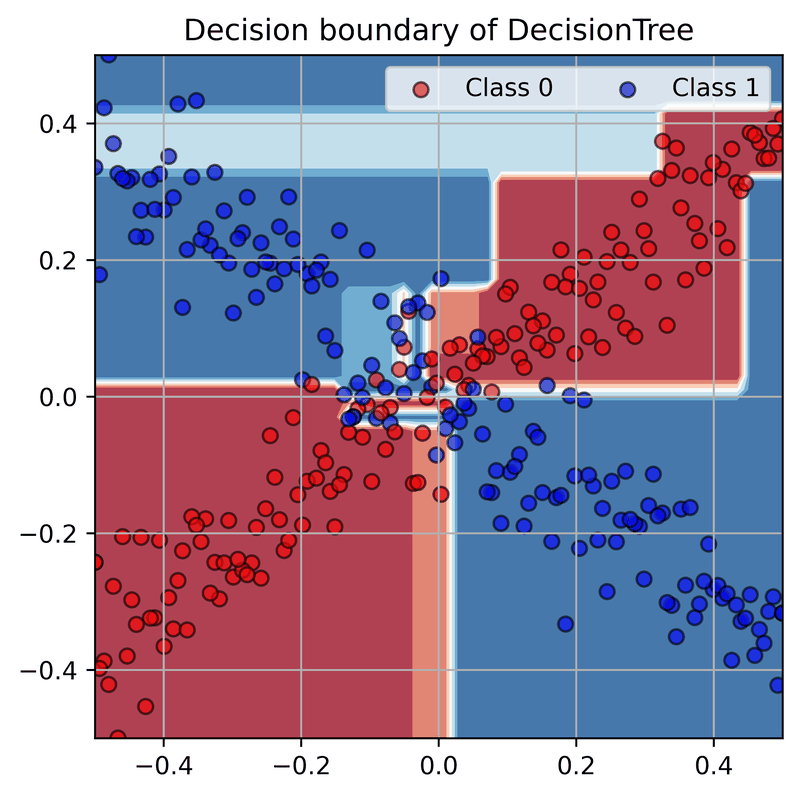}
    \includegraphics[width=0.24\linewidth]{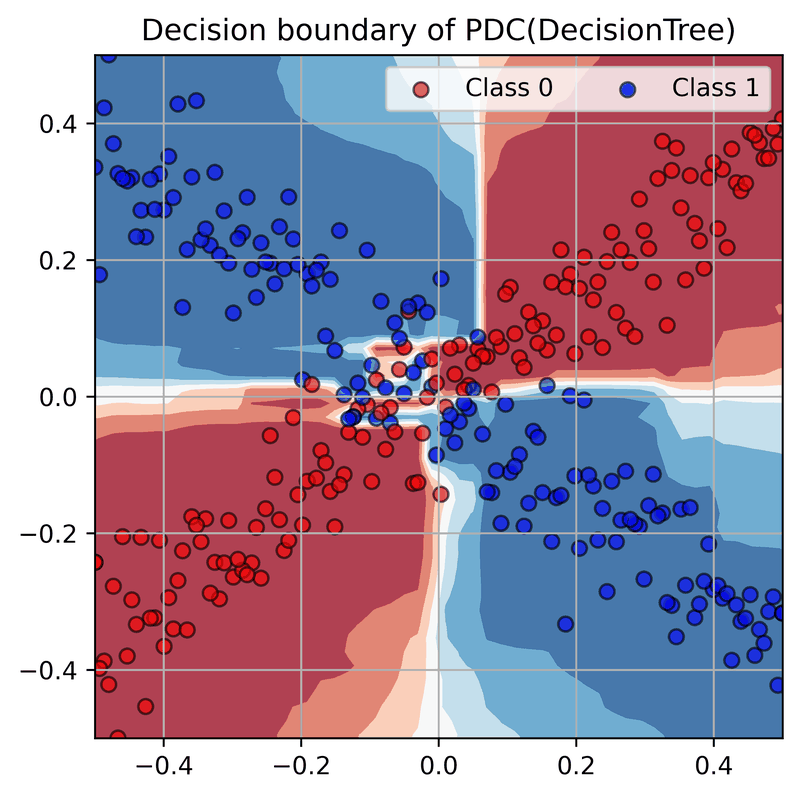}
    \caption{Comparison of learned patterns: (a) Deterministic predictor (1-NN) and its dual \(\text{pdc}_{\text{1-NN}}\); (b) Probabilistic predictor (DT) and its dual \(\PDC_{\text{DT}}\).}
    \label{fig:2d_deterministic_probabilistic}
\end{figure}

\cref{fig:2d_deterministic_probabilistic} (a) illustrates this transformation by comparing the decision boundaries of a 1-Nearest-Neighbor (1-NN) classifier, which outputs only class labels, with those of its corresponding PDL variant, denoted \(\text{pdc}_{\text{1-NN}}\). While 1-NN provides no uncertainty information, the \(\text{pdc}_{\text{1-NN}}\) model yields a full probability distribution, thereby reflecting the inherent uncertainty in the prediction.
\subsection{Probabilistic Predictors}
When the base learner is probabilistic, it produces a finite set of first-order pair predictions for each input \(x\). Averaging these predictions yields a single probability vector \(\theta_x \in \Delta_K\), thereby capturing only the aleatoric uncertainty inherent in the data. Alternatively, instead of simply averaging, the first-order predictions can be integrated into a second-order predictor \(Q_x \in \mathbb{P}(\Delta_K)\). This second-order representation, which can be viewed as a distribution over probability distributions, encapsulates epistemic uncertainty by reflecting the variability among the individual pair predictions.

\cref{fig:2d_deterministic_probabilistic} (b) illustrates the difference between a conventional probabilistic predictor, a Decision Tree (DT) with a minimum of 4 training samples per leaf, and its PDL-enhanced variant. Although DT outputs a first-order prediction \(\theta\) that reflects uncertainty in the predicted class, its decision boundaries partition the input space \(\mathcal{X}\) into large homogeneous rectangles, each with a constant uncertainty estimate. In contrast, both the averaged (\(\text{pdc}\)) and integrated (\(\text{PDC}\)) predictors yield a smoother and more precise approximation of the underlying first-order uncertainty. Moreover, the integrated PDC model is capable of discerning differences in the second-order distribution \(Q_x(\theta)\) for input points with similar first-order predictions \(\theta\), thereby providing a richer representation of uncertainty.

PDL models built with probabilistic base learners achieve robust second-order uncertainty representations, widely regarded as the optimal level for UQ. In practice, second-order learners such as Random Forests or Extra Trees are more commonly employed due to their superior performance \cite{belaid2024pairwise}. This practical preference underscores the critical need to investigate effective methods for representing and leveraging uncertainty within the PDL framework when the base learner already delivers second-order predictions.

\subsection{Second-order Predictors}\label{sub:predictor_bayesian}

A second-order base learner $\gamma$ outputs for each pair \((x, x_i)\) a second-order distribution \(Q_x \in \mathbb{P}(\Delta_K)\), where $Q_x(\theta)$ is the probability (density) of the first-order distribution $\theta$, i.e., the probability (density) that $p(\cdot \given x) = \theta$.
Practically, we assume such a learner to be realized in the form of an ensemble of first-order predictors \(\{\gamma_1, \dots, \gamma_M\}\), which is the most common way of representing second-order uncertainty information in machine learning \cite{lakshminarayanan2017simple}. The second order predictive distribution $Q_x$ is then given by a finite set of first-order predictions $\{ \theta_1, \ldots , \theta_M \}$, which can be thought of as an approximation of the underlying second-order distribution. We will use the same notation $Q_x$ for both, a second-order distribution and its set-based approximation.

For a fixed input \(x\) and any anchor \(x_i\) (with \(i=1,\dots,N\)), an ensemble \(\{\gamma_1, \dots, \gamma_M\}\) produces a set of predictive distributions
\[
Q_{x,i} = \big\{ \theta_{x,i,j} = p_{\text{post},i}( \cdot \mid x, \gamma_j) : j = 1, \dots, M \big\} \, .
\]
Collecting these over all anchors yields
\[
\mathcal{Q}_x = \big\{ \{ \theta_{x,i,j} : j = 1, \dots, M \} , \, i = 1, \dots, N \big\} \, ,
\]
i.e., a nested structure forming a higher-order uncertainty representation that can be represented as follows:

\medskip

    \begin{tabular}{l|ccccl}
    & $\gamma_1$ & $\gamma_2$ & $\ldots$ & $\gamma_M$ & \\
    \hline
$x_1$ & $\theta_{x,1,1}$ & $\theta_{x,1,2}$ & $\ldots$ & $\theta_{x,1,M}$ & $\theta_{x,1,\bullet}$\\
$x_2$ & $\theta_{x,2,1}$ & $\theta_{x,2,2}$ & $\ldots$ & $\theta_{x,2,M}$ & $\theta_{x,2,\bullet}$ \\
$\vdots$ & $\vdots$ & $\vdots$ & $\vdots$ & $\ddots$ & \vdots \\
$x_N$ & $\theta_{x,N,1}$ & $\theta_{x,N,2}$ & $\ldots$ & $\theta_{x,N,M}$ & $\theta_{x,N,\bullet}$\\
\hline
&  $\theta_{x,\bullet,1}$ & $\theta_{x,\bullet,2}$ & \ldots & $\theta_{x,\bullet,M}$  &
    \end{tabular}

\medskip

We consider three strategies for turning this rich information into a manageable second-order distribution through averaging:
\begin{itemize}
    \item \textbf{Uncertainty over ensemble learner predictions (UL).} One strategy is to condense the information from each ensemble member $\gamma_j$ from second-order to first-order by averaging over anchors:
    $$
    \theta_{x,\bullet,j} = \frac{1}{N}  \sum_{i=1}^N  \theta_{x,i,j}
    $$
    Collecting these first-order distributions yields the second-order distribution
    $$
    Q_x^{\text{UL}} = \big\{ \theta_{x,\bullet,j}  : j = 1, \dots, M \big\}.
    $$

\item \textbf{Uncertainty over Pair predictions (UP).} Likewise, one can condense the information from each anchor $x_i$ from second-order to first-order by averaging over the learners in the ensemble:
$$
\theta_{x,i, \bullet} = \frac{1}{M} \sum_{j=1}^M  \theta_{x,i,j}
$$
Collecting these first-order distributions yields the second-order distribution
    $$
    Q_x^{\text{UP}} = \big\{ \theta_{x,i, \bullet}  : i = 1, \dots, N \big\}.
$$
\item
\textbf{Uncertainty over both learner member and Pair predictions (ULP).} Yet another strategy merges all distributions into a single second-order representation:
    \[
    Q_x^{\text{ULP}} = \big\{ \theta_{x,i, j}  : i = 1, \dots, N , \, j=1, \ldots , M \big\}.
    \]
\end{itemize}

\begin{figure}
    \centering
    \includegraphics[width=1\linewidth]{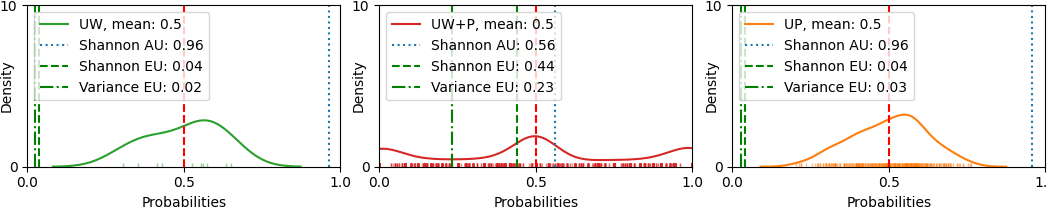}
    \caption{Second-order probability distributions predicted by $\PDC_{RF}$ using, from left to right, the UL, ULP, and UP approaches. Each representation impacts the uncertainty measurements differently.}
    \label{fig:2d_3u_pdc_rf}
\end{figure}

A concrete example of a second-order base learner in a PDC model is depicted in \cref{fig:2d_3u_pdc_rf}. The base learner used is a Random Forest (RF) with ten decision trees, supplemented by 300 anchor points for pairwise comparisons, applied to the previously introduced 2D synthetic dataset. The test instance \( x = (0, -0.5) \) lies in the center of a sparsely populated region of the input space \(\mathcal{X}\). Consequently, the prediction \( \PDC_{RF}(x) \) takes the form of a third-order distribution: 300 pairwise predictions, each comprising 10 individual predictions generated by the decision trees. As \cref{fig:2d_3u_pdc_rf} illustrates, the manner in which ensemble member predictions are grouped and merged significantly influences the shape of the resulting second-order distribution, ranging from nearly Gaussian to an almost uniform spread. This observation underscores the importance of selecting the appropriate method (UL, UP, or ULP) to capture the most informative representation of uncertainty.

\subsection{Uncertainty Measures}

When a model's predictive output is a second-order distribution \(Q_x\), measuring the overall uncertainty it represents in terms of a single scalar value is often useful. Moreover, this total uncertainty (TU) can be decomposed into aleatoric uncertainty (AU) and epistemic uncertainty (EU). AU reflects the inherent randomness in the data-generating process that cannot be reduced by additional data. In contrast, EU arises from the learner's lack of knowledge (e.g., a self-driving car facing an unfamiliar road) and typically diminishes as more or better data become available \cite{hullermeier2021aleatoric}. This decomposition is crucial for decision-making: if uncertainty is mainly aleatoric, further data will not help, whereas high epistemic uncertainty indicates that improved data or models could enhance predictions. Such distinctions are especially valuable, e.g., in active learning \cite{shelmanov2021active,nguyen2022how} and Bayesian optimization \cite{hoffer2023robust,stanton2023bayesian}, where reducing EU is key to efficiency \cite{sale2023secondorder}.

\subsubsection{Entropy-based Information Measures}

A popular family of techniques for capturing predictive uncertainty relies on information-theoretic measures, notably Shannon entropy \cite{wimmer2023quantifying}.
A classical result in information theory shows that Shannon entropy $H$ decomposes (additively) into conditional entropy and mutual information \cite{ash2012information}: Let $Y \sim \theta_x$ be the outcome produced by the first-order distribution $\theta_x$, which is unknown and hence treated as a random variable $\Theta \sim Q_x$. Therefore, $Y \sim \bar{\theta}$ with expected first-order distribution $\bar{\theta} = \mathbb{E}_{\Theta \sim Q_x}(\Theta)$. The entropy $H(Y) = H(\bar{\theta})$, which qualifies as a measure of total uncertainty (about the outcome $Y$) can then be written as $H(Y \given \Theta) + I(Y,\Theta)$, where $H(Y \given \Theta)$ is the conditional entropy and $I(Y,\Theta)$ the mutual information between $Y$ and $\Theta$. In the literature, it has been suggested that the former qualifies as a measure of aleatoric and the latter as a measure of epistemic uncertainty \cite{depeweg2018decomposition}.

In the case where $Q_x$ is represented as a finite set of first-order distributions, expectations are replaced by arithmetic averages, so that
the total uncertainty (\(\TU\)), aleatoric uncertainty (\(\AU\)), and epistemic uncertainty (\(\EU\))
associated with $Q_x$ are given as follows:
\begin{align}
    \TU(Q_x) \; &= \; H(\bar{\theta}) = H \left(\frac{1}{|Q_x|}\sum_{\theta \in Q_x} \theta \right),\label{eq:measure_entropy_pdc_tu}\\
    \AU(Q_x) \; &= \; \frac{1}{|Q_x|} \sum_{\theta \in Q_x} H(\theta),\label{eq:measure_entropy_pdc_au}\\
    \EU(Q_x) \; &= \; \TU(Q_x) \;-\; \AU(Q_x).\label{eq:measure_entropy_pdc_eu}
\end{align}
Note that total uncertainty coincides for all three approaches (UL, UP, ULP)\,---\,they average over the $\theta_{x,i,j}$ in different ways but eventually produce the same $\bar{\theta}$.

\subsubsection{Variance-Based Measures}

As an alternative to entropy, (total) predictive uncertainty can also be captured in terms of variance, and a decomposition into aleatoric and epistemic parts can be obtained from the law of total variance
\cite{jimenez-luna2019deltadelta,tynes2021pairwise,sale2023secondorder}.
Unlike Shannon entropy, however, variance is not directly suited to categorical outcomes. Therefore, \citeauthor{sale2023secondorder} proposed to represent a categorical outcome $Y \in \{1, \ldots , K\}$ in terms of a binary vector (one-hot encoding) $(Y_1, \ldots , Y_K)$, where $Y_k = 1$ if $Y=k$ and $=0$ otherwise. If $Y \sim \theta = (\theta_1, \ldots , \theta_K)$, then each entry $Y_k$ can be seen as a Bernoulli-distributed random variable with expectation $\mathbb{E}(Y_k)= \theta_k$ and variance $\mathbb{V}(Y_k) = \theta_k (1- \theta_k)$. Moreover, assuming $\theta_k$ to be unknown and treating it as a random variable $\Theta_k$ distributed according to second-order distribution $Q_x$, we obtain the following decomposition:
\begin{align}\label{eq:measure_var_k}
    \TU_k(Q_x) &= \Var(Y_k) \;=\; \mathbb{E}\!\bigl[\Theta_k(1-\Theta_k)\bigr] \;+\; \Var(\Theta_k),\\
    \AU_k(Q_x) &= \mathbb{E}\!\bigl[\Theta_k\,(1 - \Theta_k)\bigr], \\
    \EU_k(Q_x) &= \Var(\Theta_k).
\end{align}
The overall uncertainty can then be computed by summing the uncertainties associated with the $K$ Bernoulli random variables:
\begin{align}\label{eq:measure_var}
    \TU(Q_x) &= \sum_{k=1}^{K} \,\Var(Y_{k}), \\
    \AU(Q_x) &= \sum_{k=1}^{K} \,\mathbb{E}\!\bigl[\Theta_{k}(1-\Theta_{k})\bigr], \\
    \EU(Q_x) &= \sum_{k=1}^{K} \,\Var(\Theta_{k}).
\end{align}

\section{Evaluation}
\label{sec:evaluation}
The architecture of our proposed PDL classifiers enables precise estimation of predictive uncertainty, which we decompose into AU and EU. These uncertainties can be quantified using either Shannon entropy–based or variance–based metrics. Although previous literature provides insights into which measures might perform best, our empirical evaluation aims to uncover potential interactions among the PDL architecture (i.e., the pairing of data), the baseline model (probabilistic versus Bayesian), the uncertainty representation (TU, AU, EU), and the UQ metric (entropy versus variance). In this section, we evaluate PDC's UQ performance with five base learners implemented in Scikit-learn, encompassing both probabilistic (DecisionTree, ExtraTree) and Bayesian learners (RandomForest, ExtraTrees, Bagging of MLP).\footnote{Source code available at: \url{https://github.com/Karim-53/pdll}}

\subsection{Data}
\begin{figure}[htb]
\centerline{\includegraphics[width=\columnwidth]{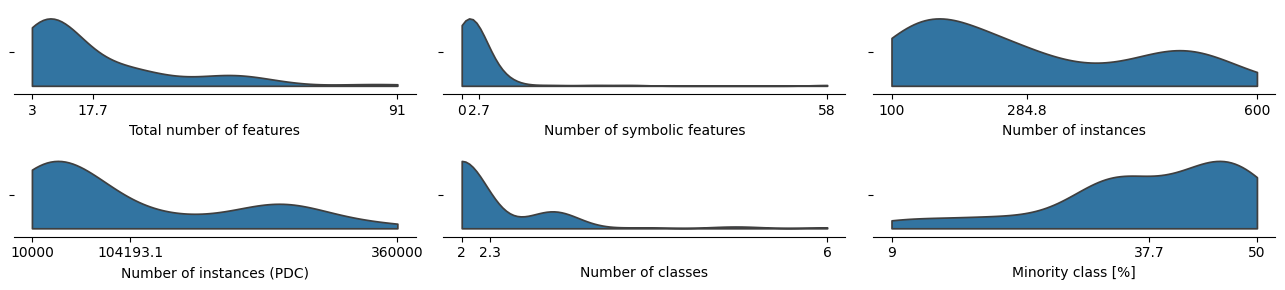}}
\caption{Distribution of key characteristics (minimum, mean, maximum) for the 99 OpenML classification datasets.}\label{fig:describe_data}
\end{figure}

OpenML offers a diverse collection of datasets, with a significant portion (37\%) comprising fewer than 600 data points. Our study focuses on these smaller datasets, as the pairwise learning approach is expected to be particularly effective in such scenarios.
Parameter tuning was performed via grid search CV using a subsampled TPOT search space \cite{olson2016evaluation}.
Adhering to dataset selection criteria similar to those employed in the OpenML-CC18 benchmark and previous PDL research \cite{bischl2017openml,belaid2024pairwise,belaid2024weighted}, we randomly selected 99 datasets (see summary statistics in \cref{fig:describe_data}). Although the original datasets are modest in size, the effective sample size for PDC is quadrupled through pairwise comparisons, yielding approximately $360,000$ data points. We further assessed class imbalance by examining the "minority class" metadata, which indicates the proportion of samples belonging to the minority class in each dataset. For each of the five baseline models, we conducted 5 repetitions of 5-fold CV to ensure robust performance evaluation.

We implemented a uniform data processing pipeline for all experimental runs using Scikit-learn \cite{pedregosa2011scikitlearn}. This pipeline standardized numeric features, applied one-hot encoding to nominal features, and utilized ordinal encoding for ordinal features. Since the PDL framework employs a joint feature representation—defined as \(\phi(x_i, x_j) = (x_i, x_j, x_i - x_j)\) that incorporates the pairwise difference, all processed features are treated as numeric when computing these differences.

\subsection{Metrics}
Performance is evaluated using the macro F1 score, which is particularly informative in imbalanced settings. The macro F1 score is the unweighted average of the F1 scores for each class:
\[
\text{Macro F1} = \frac{1}{K} \sum_{i=1}^{K} F1_i \, ,
\]
Results are reported as mean \(\pm\) standard error.
For each dataset, performance is aggregated over 25 runs (5 repetitions of 5-fold CV). Wins and losses are determined by comparing average scores per dataset, with statistical significance assessed via a Student's t-test (\(\alpha = 0.05\)). Note that ties may occur, so wins and losses may not sum to the total number of datasets. While overall averages provide a preliminary impression, further metrics and detailed results are available in the PDL library repository for a more nuanced evaluation.

A well-established tool for evaluating the quality of UQ is the accuracy-rejection curve (ARC), which plots model accuracy as a function of the fraction of test instances rejected due to high uncertainty \cite{hullermeier2008learning}. More concretely, consider a set of (test) instances $\{ x_1, \ldots , x_T \}$ and denote by $Q_{x_i}$ the (second-order) distribution predicted for $x_i$. Moreover, let $U$ be a measure that quantifies predictive uncertainty. We can then sort the instances in decreasing order of uncertainty: $x_{(1)}, x_{(2)}, \ldots, x_{(T)}$, where the permutation $(\cdot)$ is such that $U(Q_{x_{(1)}}) \geq U(Q_{x_{(2)}}) \geq \ldots \geq U(Q_{x_{(T)}})$. If the measure $U$ quantifies uncertainty in a meaningful way and indeed assigns higher uncertainty to more uncertain predictions, then we would expect that the accuracy on the ``less uncertain‘‘ cases $x_{(t+1)}, x_{(t+2)}, \ldots, x_{(T)}$, i.e., the accuracy of the learner when abstaining on the ``more uncertain'' cases $x_{(1)}, x_{(2)}, \ldots, x_{(t)}$, will be higher. More specifically, the accuracy should be an increasing function of the number $t$ of abstentions.

Given the extensive number of experiments, analyzing individual ARCs is impractical for assessing overall performance. To overcome this, we aggregate ARC performance into a scalar metric by computing the rank correlation between $U$ and a binary misclassification indicator ($z_i =1$ for misclassified $x_i$, $z_i = 0$ for correctly classified $x_i$) across all test instances. This idea is very much in line with the AUC metric, which ``integrates'' the ROC curve. For a perfect uncertainty measure, misclassified points $x_i$ would consistently have higher values $U(Q_{x_i})$, i.e., all misclassifications appear at the top of the sorted sequence $x_{(t+1)}, x_{(t+2)}, \ldots, x_{(T)}$, yielding a rank correlation of $+1$; negative values (between $-1$ and 0) would indicate an inverse relationship. We adopt Goodman and Kruskal's Gamma \cite{good_mo} as our rank correlation measure due to its effective handling of ties. Concretely, this means that we compute $C/(C+D)$, where $C$ is the number of concordant (correctly ordered) pairs $(x_i, x_j)$, i.e., $z_i=1$, $z_j=0$ and $U(x_{(i)}) > U(x_{(j)})$, and $D$ is the number of discordant (incorrectly ordered) pairs $(x_i, x_j)$, i.e., $z_i=1$, $z_j=0$ and $U(x_{(i)}) < U(x_{(j)})$.

While the measure effectively captures total ($U=$ TU) and aleatoric uncertainty ($U=$ AU), it is less apt for epistemic uncertainty (EU), which reflects uncertainty about the predicted distribution \(\theta_{x_i}\) rather than the outcome \(y_i\). Thus, EU quality is assessed by comparing it to the model's calibration error. Test instances with high EU should exhibit higher calibration errors. As the true calibration error is unobservable, it is approximated by the expected calibration error (ECE) \cite{naeini2015obtaining}. In our approach, test predictions are sorted by predicted probabilities and grouped into fixed bins; for each test point \(i\), \(ECE_i\) is defined as the absolute difference between its predicted probability and the average true label in its bin.

\subsection{Baseline}

To the best of our knowledge, this is the first work to comprehensively estimate TU, AU, and EU in PDC models. Although \citeauthor{tynes2021pairwise}'s method shares conceptual similarities with our variance-based estimation of TU using the UP strategy, it was originally developed for regression tasks and employs distinct computational formulations. Consequently, we do not include direct comparisons with that approach. Instead, we establish our baseline by comparing the performance of the PDC model against its underlying base learner (e.g., when decision trees are used as base learners in PDC, we compare with standard decision tree learning, when random forests are the base learner, we compare with standard random forests, etc.).

\subsection{Results}
In this section, we provide a comprehensive evaluation of the empirical quality of the PDC framework in terms of performance and UQ. Our findings are organized according to the type of base learner employed, with both probabilistic and Bayesian models exhibiting consistent behavior within their respective groups.

\subsubsection{Predictive Performance}
The extended PDL classifier, when built upon ensemble models such as Random Forests and Extra Trees, achieved the highest average Macro F1 scores across 99 datasets, surpassing all baseline methods. As shown in \cref{tab:exp2_f1_wins}, the proportion of significant wins favors the PDL classifier, indicating that in direct comparisons it is more likely to outperform its baseline counterpart.

\begin{table}
\caption{Comparing baseline classifiers to PDC using 99 datasets.}\label{tab:exp2_f1_wins}
\centering
\begin{tabular}{lrr|rr|rrrr}
 & \multicolumn{2}{c|}{\textbf{Significant wins}}  &\multicolumn{2}{c|}{\textbf{Wins}}  &\multicolumn{4}{c}{\textbf{Average Test Macro F1}} \\
Classifier & base & PDC & base & PDC & base & $\pm$ sem & PDC & $\pm$ sem \\
\hline
DecisionTree & 2 & \textbf{50} & 22 & \textbf{76} & 0.7694 & 0.0037 & \textbf{0.7982} & 0.0034 \\
ExtraTree & 1 & \textbf{61} & 9 & \textbf{90} & 0.7434 & 0.0037 & \textbf{0.7987} & 0.0035 \\
Bagging & 3 & \textbf{26} & 27 & \textbf{70} & 0.7906 & 0.0035 & \textbf{0.8062} & 0.0034 \\
RandomForest & 5 & \textbf{27} & 22 & \textbf{73} & 0.7933 & 0.0035 & \textbf{0.8073} & 0.0034 \\
ExtraTrees & 6 & \textbf{24} & 21 & \textbf{77} & 0.7951 & 0.0036 & \textbf{0.8113} & 0.0035 \\
\hline
\end{tabular}
\end{table}

Moreover, the PDL framework can simplify the final model. For instance, the test performance of \(\text{PDC}_{\text{DT}}\) is comparable to or better than that of the five benchmarked learners. This suggests that, by employing PDL, a single decision tree can match or exceed the performance of ensembles that typically consist of around 100 trees, thereby offering a more interpretable solution.
 more results can be found in the appendix.

\subsubsection{UQ of PDC Using a Probabilistic Base Classifier}
\begin{figure}[!ht]
    \centering
    \includegraphics[width=0.49\linewidth]{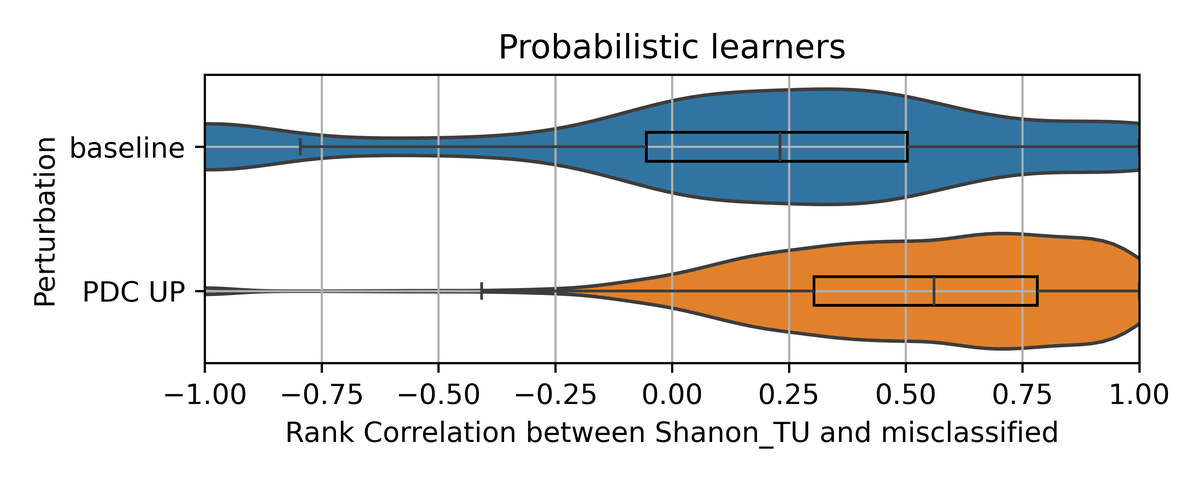}
    \includegraphics[width=0.49\linewidth]{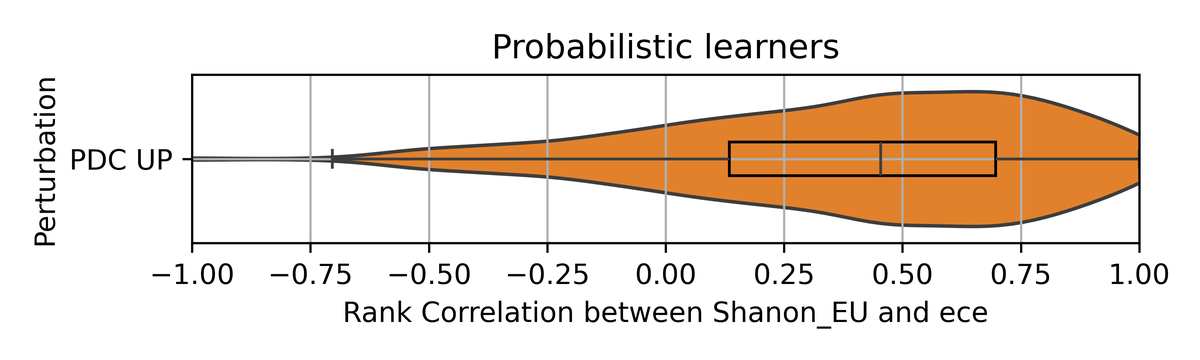}
    \includegraphics[width=0.49\linewidth]{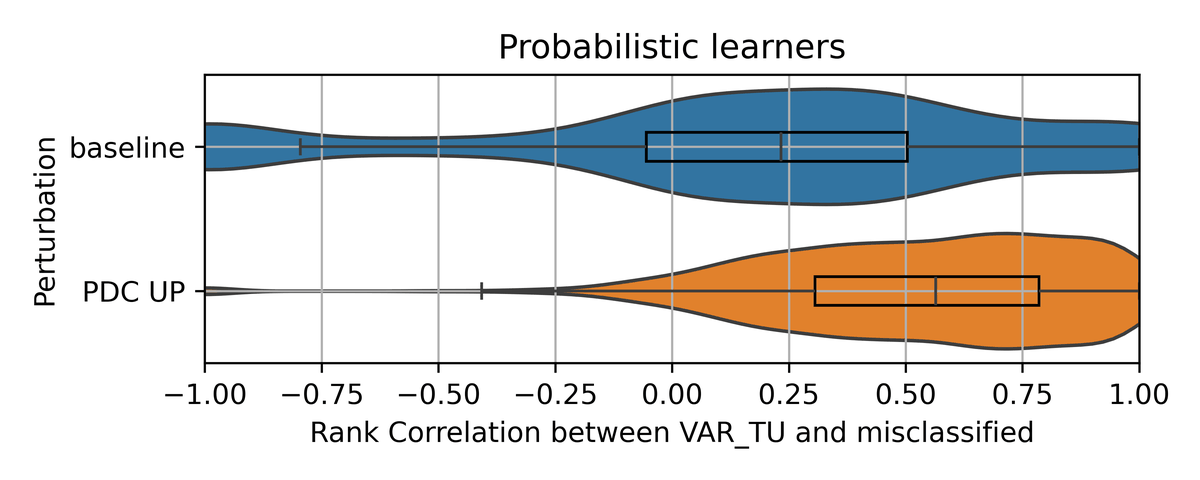}
    \includegraphics[width=0.49\linewidth]{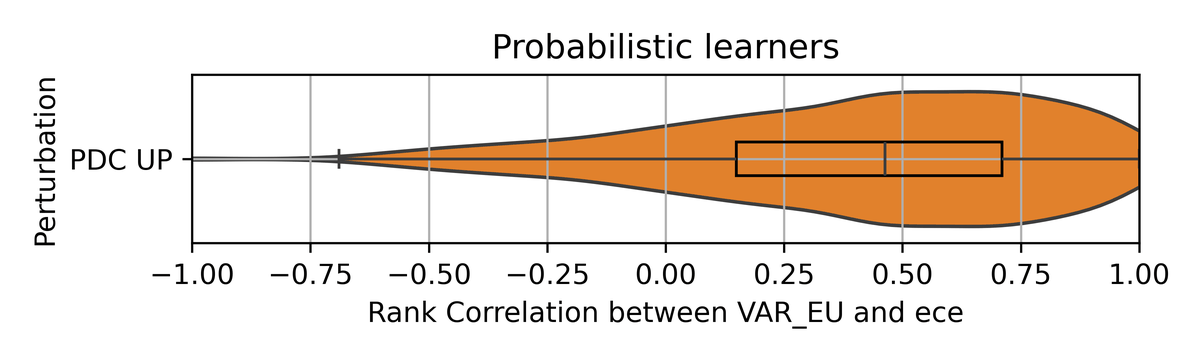}
    \caption{Distribution of Rank Correlation with probabilistic learners.}
    \label{fig:rank_probabilistic}
\end{figure}

When employing a probabilistic base learner, the direct estimation of total uncertainty (TU) yields a median rank correlation of 0.23 (see the blue distribution in \cref{fig:rank_probabilistic}). In contrast, the PDC model built atop the same probabilistic learner achieves a marked improvement, with a mean rank correlation of 0.56 and only 5\% negative correlations. Moreover, beyond enhancing TU estimation, the PDC framework also enables the assessment of EU, as demonstrated by the predominance of positive ECE-EU rank correlations.

Both entropy-based and variance-based uncertainty measures facilitate meaningful quantification of prediction uncertainty within the PDC framework, and these trends are consistent across all tested probabilistic classifiers. %
Additionally, we observe a high Pearson correlation between AU and TU (\(\rho = 0.98\)), indicating that in these scenarios, TU is predominantly driven by aleatoric noise. Similar distributions are obtained when evaluating AU on the misranking task.

\subsubsection{UQ of PDC Using a Second-order Base Classifier}
\begin{figure}[!ht]
    \centering
    \includegraphics[width=0.49\linewidth]{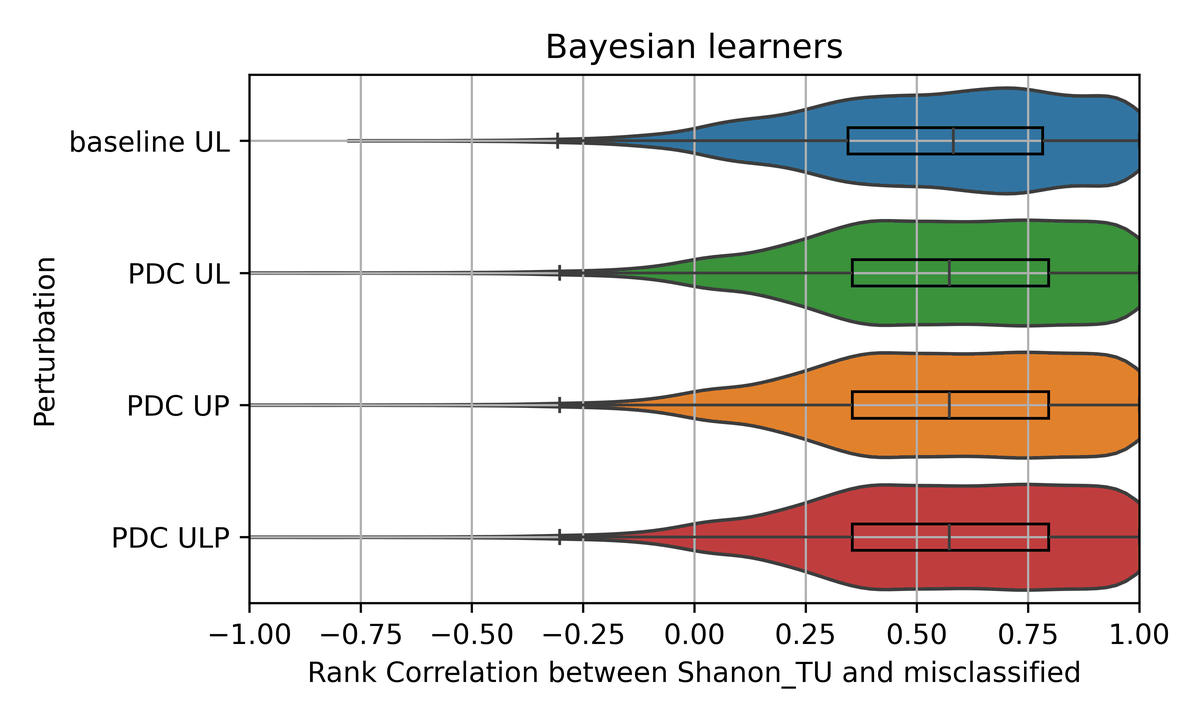}
    \includegraphics[width=0.49\linewidth]{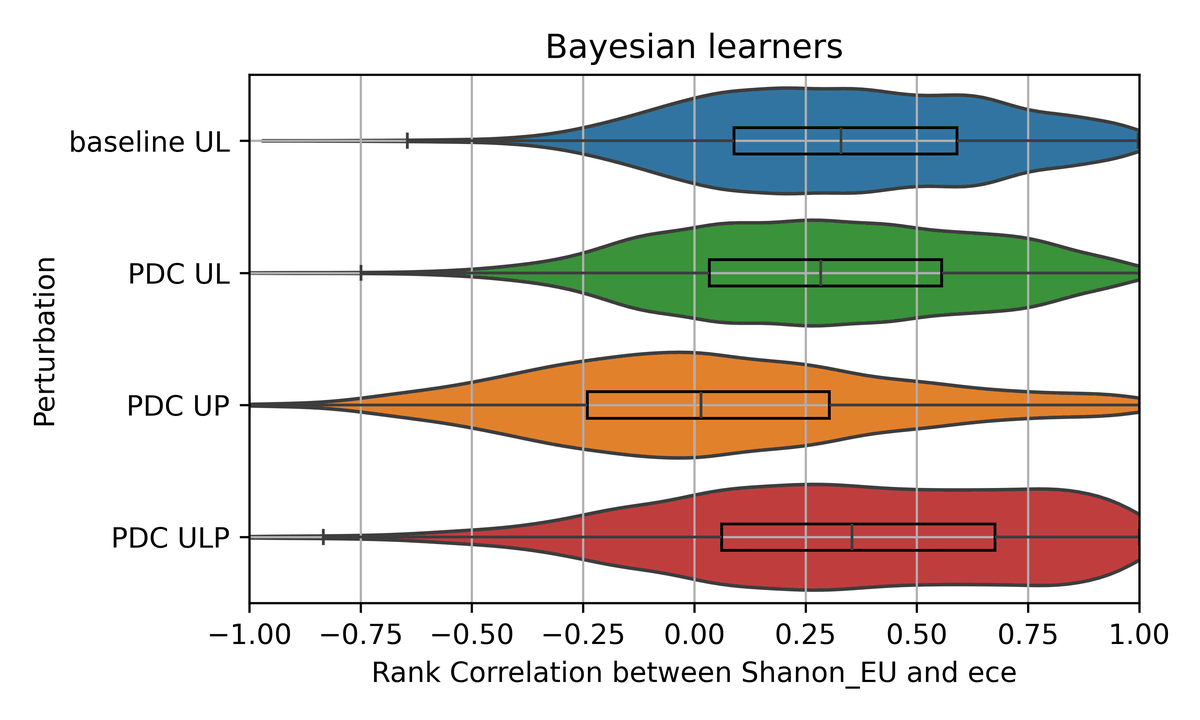}
    \includegraphics[width=0.49\linewidth]{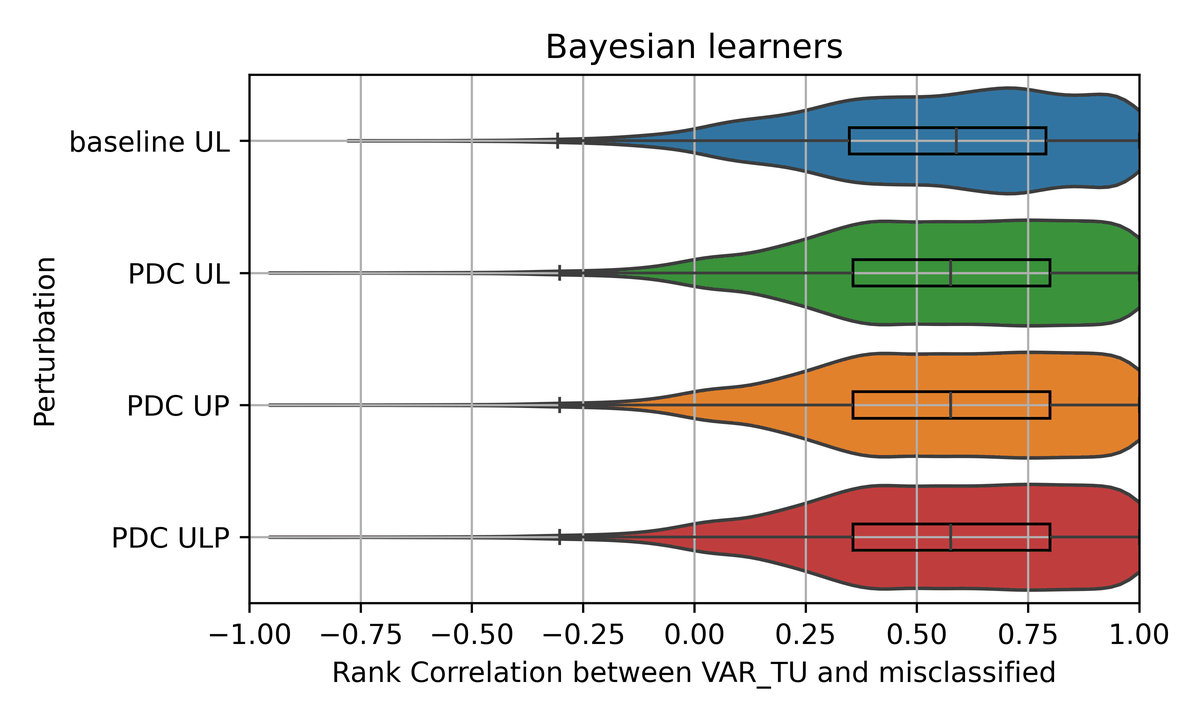}
    \includegraphics[width=0.49\linewidth]{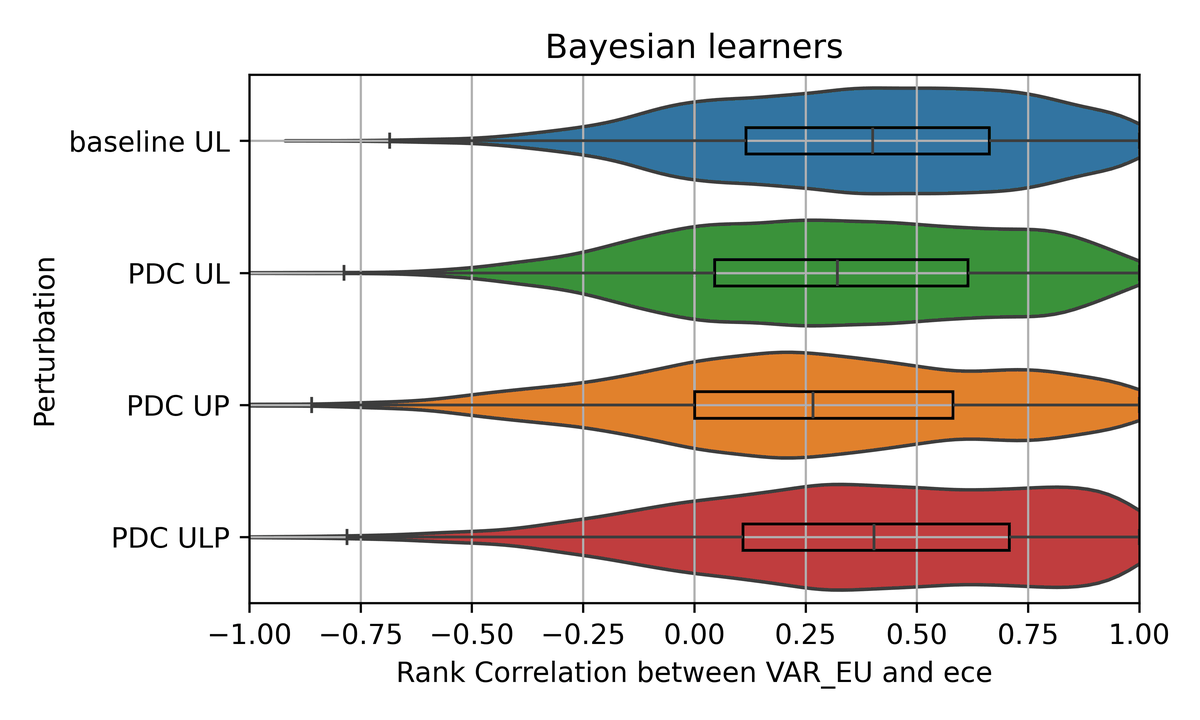}
    \caption{Distribution of Rank Correlation with Bayesian learners.}
    \label{fig:rank_beyesian}
\end{figure}

When the base learner is Bayesian, higher-order uncertainties become accessible, providing more nuanced estimates of uncertainty. Specifically, we consider three strategies for representing the second-order uncertainty: UL, UP, or ULP. %
As in the probabilistic setting, we find that the TU again exhibits a high correlation with AU, and both effectively guide selective prediction as evidenced by high rank correlations. %
However, the estimation of EU varies in quality depending on the chosen strategy. The EU values from ULP lead to the highest rank correlations with a slight non-significant improvement over a baseline Bayesian classifier, see \cref{fig:rank_beyesian}.

\section{Discussion}
\label{sec:discussion}

\subsection{Effect of PDC on the predictive performance}

The enhanced prediction performance of PDL can be attributed to three main factors. First, the framework integrates instance-based and model-based learning by leveraging anchor points to capture local information, while employing a global classifier to capture general patterns of the data distribution.
Second, PDL simplifies the original multiclass task by reducing it to a single binary problem\,---\,determining whether two instances share the same label or not. This binary reduction not only streamlines the learning problem but also contributes to improved predictive accuracy.
Lastly, PDL benefits from an ensemble effect through averaging predictions from multiple anchors, which reduces the variance of prediction errors. Collectively, these mechanisms explain the superior performance of PDL compared to conventional baseline methods.

\subsection{Effect of PDC on the Uncertainty Quantification}

\begin{figure}
    \centering
    \includegraphics[width=0.24\linewidth]{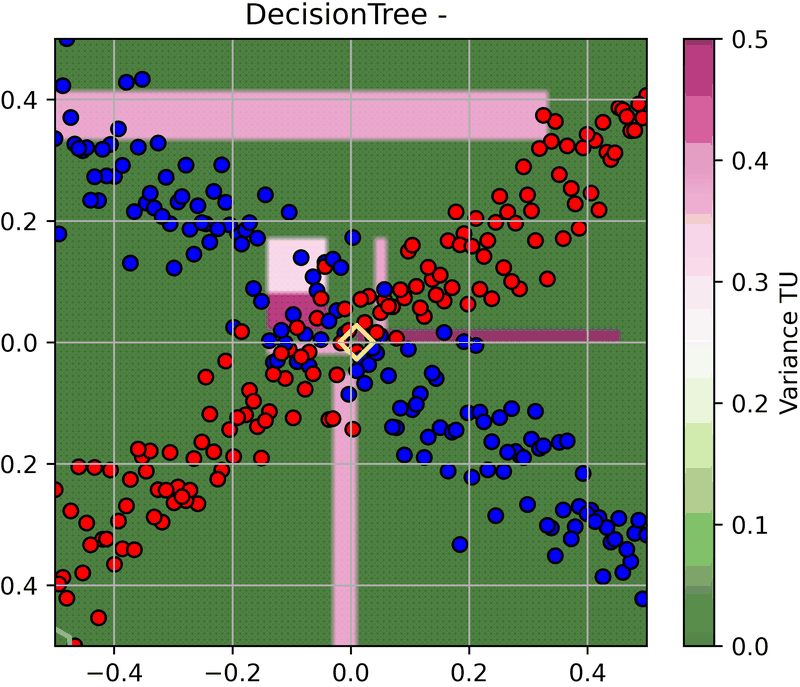}
    \includegraphics[width=0.24\linewidth]{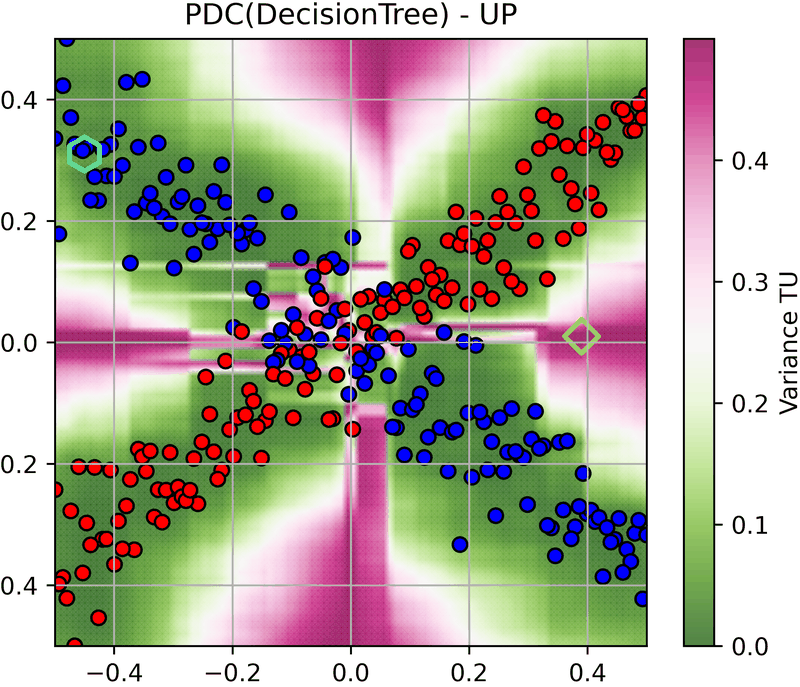}
    \includegraphics[width=0.24\linewidth]{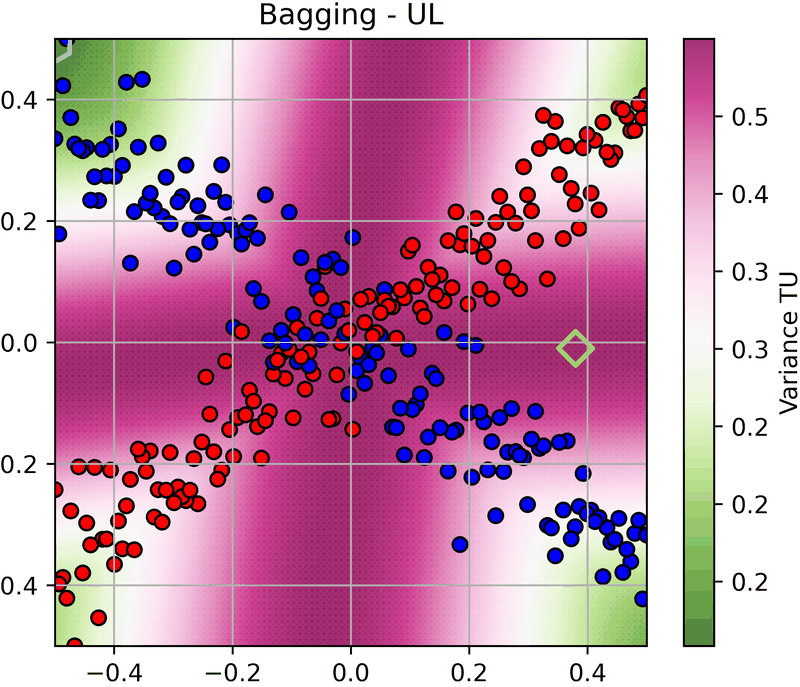}
    \includegraphics[width=0.24\linewidth]{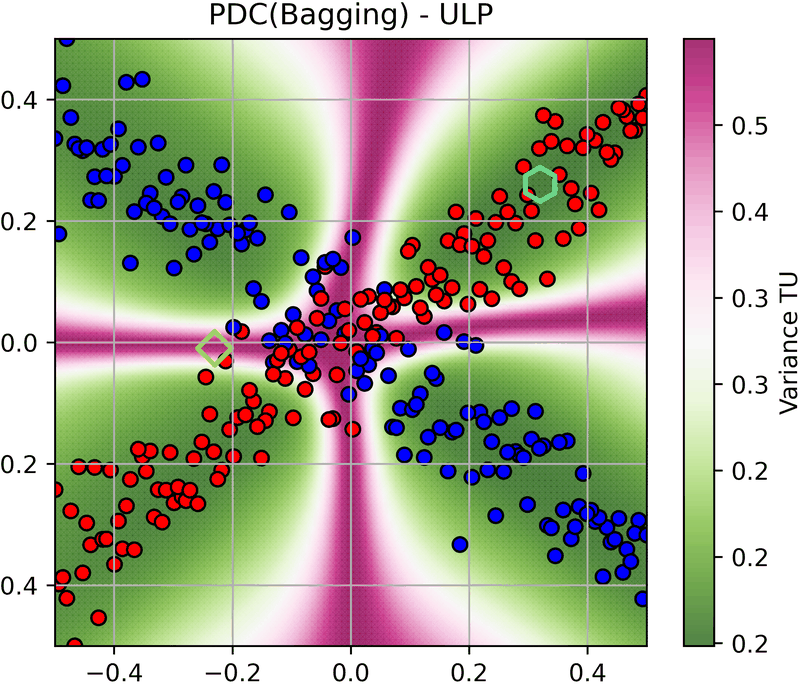}
    \caption{Estimation of TU: Baseline DT and Bagging(MLP) models fail to capture total uncertainty accurately, whereas their PDC counterparts yield robust TU estimates.}
    \label{fig:2d_different_tu}
\end{figure}

Empirical results show that variance-based measures match or outperform Shannon entropy-based ones, supporting both prior studies and theoretical arguments \cite{sale2023secondorder,wimmer2023quantifying}.
Integrating probabilistic learners in PDC yields a significant improvement in the average correlation between total uncertainty and misclassifications, while also substantially reducing the incidence of negative correlations across datasets—from 29\% to just 5\%.
This observation is further corroborated by the TU estimation on a 2D synthetic dataset (see \cref{fig:2d_different_tu}). A standard DecisionTree tends to assign uniformly high TU across both dense and sparse regions, whereas the PDC-enhanced DecisionTree produces a smoother uncertainty distribution that increases gradually with input sparsity. Similarly, a basic Bagging ensemble of MLPs assigns similar uncertainty levels to dense, sparse, and noisy regions, while its PDC counterpart leverages the neural network's feature extraction capabilities to delineate curvilinear boundaries that effectively separate uncertain regions from those that are noisy or sparsely populated.

Furthermore, regarding EU, a PDC model with a probabilistic base learner delivers EU estimates comparable to those from a baseline second-order ensemble model, despite relying on only a single trained tree. This efficiency highlights the potential of the PDC framework to achieve robust UQ with significantly reduced computational complexity.

When comparing the three strategies UL, UP, and ULP for second-order base learners, ULP leads to the best quantification of EU. Apparently, the estimation of EU benefits from combining both sources of variability, namely variability from the base learners and from the anchor points (pair predictions). UL and UP lose this advantage, because they both ``average out'' one of these sources.

\begin{figure}[h]
    \includegraphics[width=0.24\linewidth]{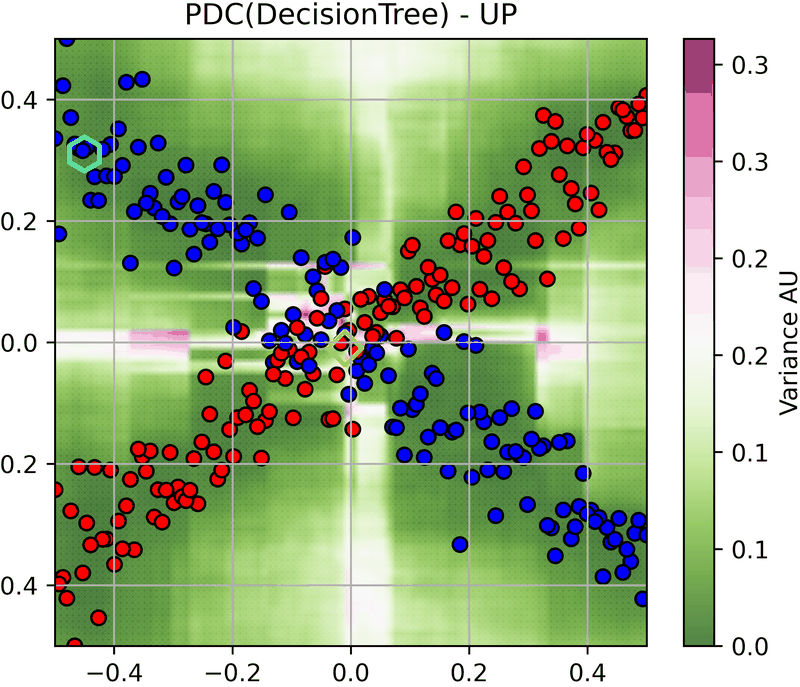}
    \includegraphics[width=0.24\linewidth]{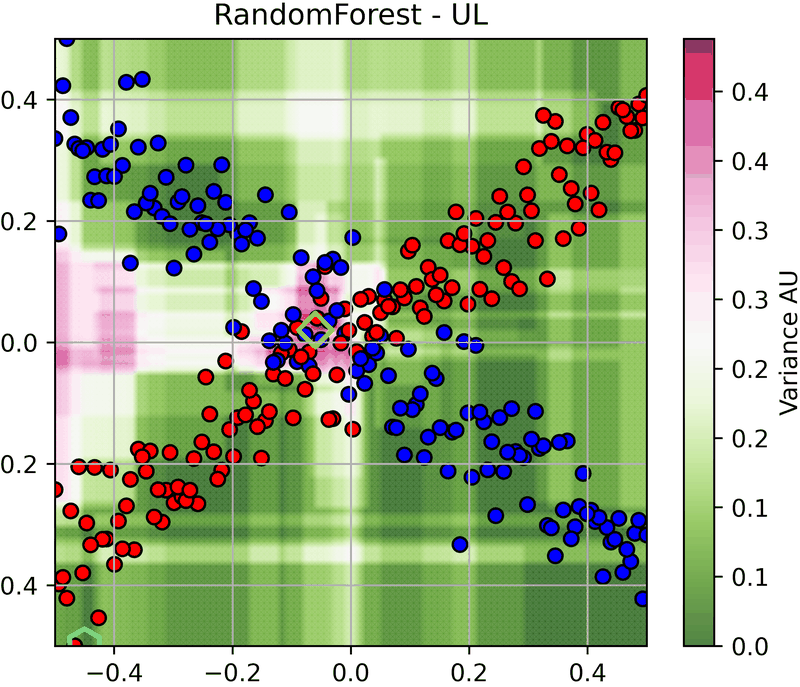}
    \includegraphics[width=0.24\linewidth]{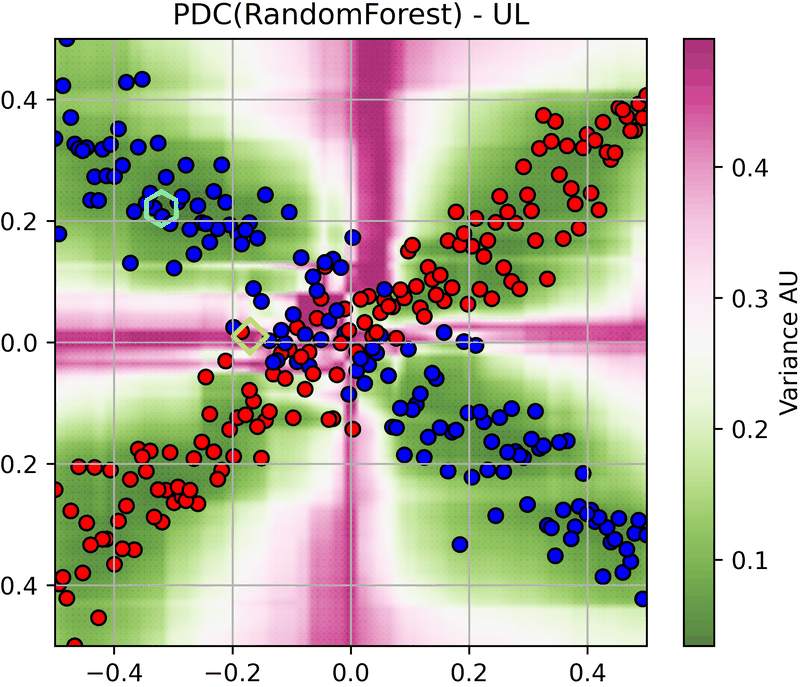}
    \includegraphics[width=0.24\linewidth]{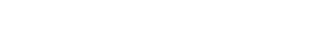}

    \includegraphics[width=0.24\linewidth]{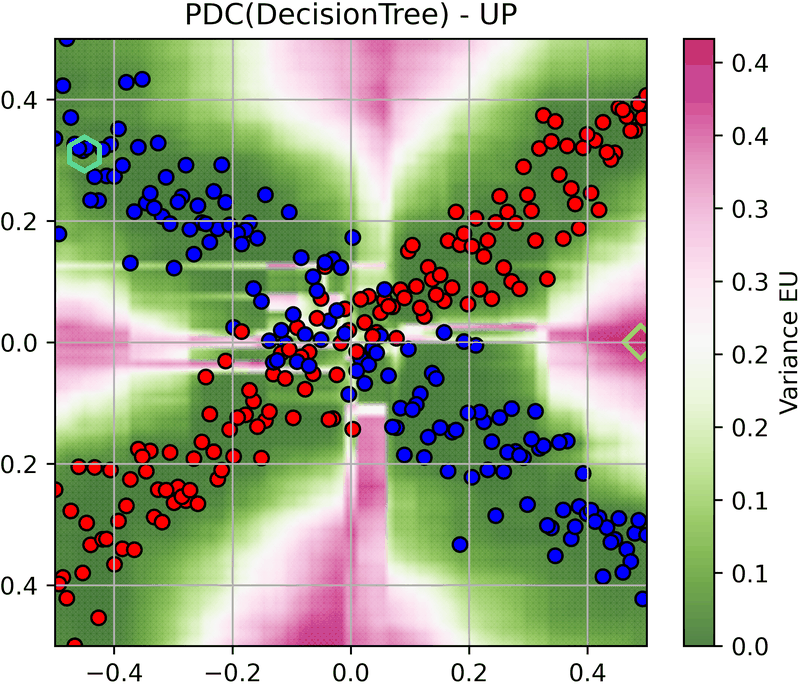}
    \includegraphics[width=0.24\linewidth]{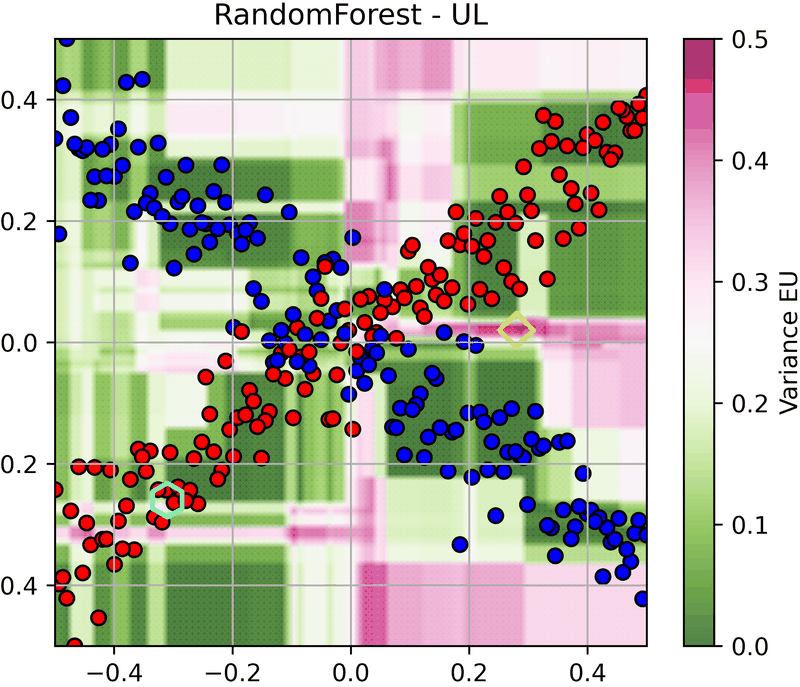}
    \includegraphics[width=0.24\linewidth]{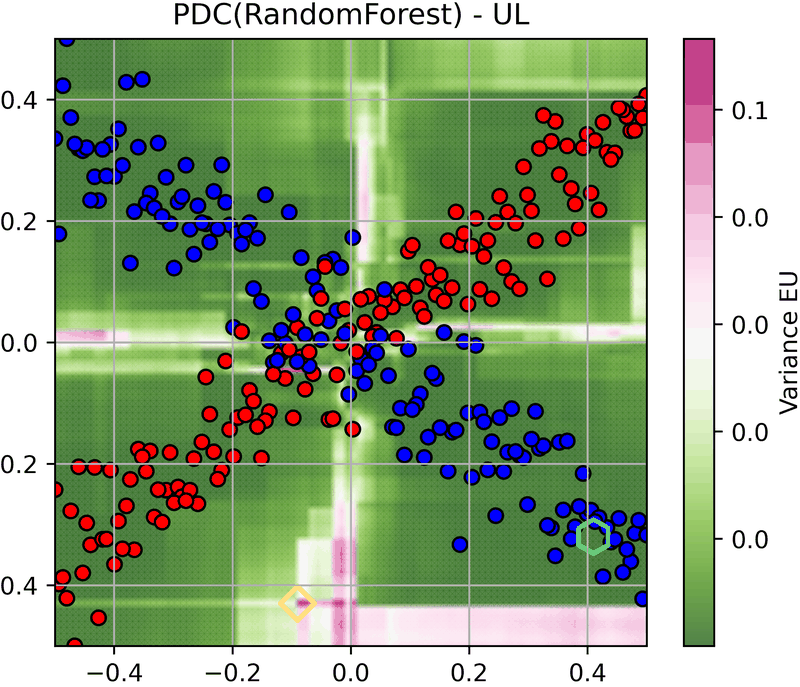}
    \includegraphics[width=0.24\linewidth]{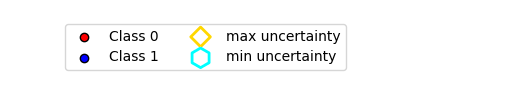}
    \caption{Splitting of AU and EU. RandomForest incorrectly allocates AU to an empty region, whereas both PDC(DT) and PDC(RandomForest) consistently assign the highest AU to the noisy center and the highest EU to the sparse region.}
    \label{fig:2d_different_eu}
\end{figure}
As shown in \cref{fig:2d_different_eu}, the RandomForest classifier tends to incorrectly split AU and EU by allocating excessive AU to an empty region. In contrast, both PDC(DT) and PDC(RandomForest) demonstrate a consistent pattern by assigning the highest AU values to the noisy center and the highest EU values to the sparse region. This observation highlights the superior performance of the PDC methods in distinguishing between AU and EU under challenging conditions.

\section{Conclusion}

In this paper, we we developed an extension of pairwise difference learning from regression to the case of multiclass classification. To this end, we proposed a meta-learning technique that specifies the ``difference'' between outcomes in terms of class (in)equality, thereby reducing the problem to a binary classification task.
Like in the case of regression, empirical studies have shown significantly improved accuracy compared to conventional classification.

Building on the concept of pairwise difference classification, we also established a comprehensive framework for uncertainty quantification that adapts to the nature of the base learner. Specifically, we distinguish three levels of uncertainty-awareness of the base learner: deterministic, probabilistic, and second-order (Bayesian). Thanks to its ensembling nature, PDC ``lifts'' the uncertainty-awareness of a deterministic base learner to the probabilistic level, which allows it to capture aleatoric uncertainty. Likewise, PDC lifts a probabilistic base learner from the first to the second level, thereby allowing for the representation of both AU and EU. For the case of a base learner that is already Bayesian, we developed strategies to keep it second-order, because a third-order learner would be overly complex to interpret.

Our methodology was rigorously validated through extensive empirical studies.
As a takeaway, we recommend using the uncertainty over ensemble learners and pair predictions (ULP) approach, which coincides with the uncertainty over the pair predictions (UP) approach when the base learner is probabilistic instead of Bayesian. Although other approaches could perform better on specific datasets, these approaches yield the best results most of the time. Furthermore, variance-based measures consistently outperform information-theoretic uncertainty measures. Following these recommendations, PDC models can represent predictive uncertainty in a faithful manner and, compared to the underlying base learners, deliver significantly enhanced uncertainty quantification, thereby fostering more reliable and interpretable decisions in high-stakes applications.

\bibliography{bibtex}

%% BioMed_Central_Bib_Style_v1.01

\begin{thebibliography}{39}
% BibTex style file: bmc-mathphys.bst (version 2.1), 2014-07-24
\ifx \bisbn   \undefined \def \bisbn  #1{ISBN #1}\fi
\ifx \binits  \undefined \def \binits#1{#1}\fi
\ifx \bauthor  \undefined \def \bauthor#1{#1}\fi
\ifx \batitle  \undefined \def \batitle#1{#1}\fi
\ifx \bjtitle  \undefined \def \bjtitle#1{#1}\fi
\ifx \bvolume  \undefined \def \bvolume#1{\textbf{#1}}\fi
\ifx \byear  \undefined \def \byear#1{#1}\fi
\ifx \bissue  \undefined \def \bissue#1{#1}\fi
\ifx \bfpage  \undefined \def \bfpage#1{#1}\fi
\ifx \blpage  \undefined \def \blpage #1{#1}\fi
\ifx \burl  \undefined \def \burl#1{\textsf{#1}}\fi
\ifx \doiurl  \undefined \def \doiurl#1{\url{https://doi.org/#1}}\fi
\ifx \betal  \undefined \def \betal{\textit{et al.}}\fi
\ifx \binstitute  \undefined \def \binstitute#1{#1}\fi
\ifx \binstitutionaled  \undefined \def \binstitutionaled#1{#1}\fi
\ifx \bctitle  \undefined \def \bctitle#1{#1}\fi
\ifx \beditor  \undefined \def \beditor#1{#1}\fi
\ifx \bpublisher  \undefined \def \bpublisher#1{#1}\fi
\ifx \bbtitle  \undefined \def \bbtitle#1{#1}\fi
\ifx \bedition  \undefined \def \bedition#1{#1}\fi
\ifx \bseriesno  \undefined \def \bseriesno#1{#1}\fi
\ifx \blocation  \undefined \def \blocation#1{#1}\fi
\ifx \bsertitle  \undefined \def \bsertitle#1{#1}\fi
\ifx \bsnm \undefined \def \bsnm#1{#1}\fi
\ifx \bsuffix \undefined \def \bsuffix#1{#1}\fi
\ifx \bparticle \undefined \def \bparticle#1{#1}\fi
\ifx \barticle \undefined \def \barticle#1{#1}\fi
\bibcommenthead
\ifx \bconfdate \undefined \def \bconfdate #1{#1}\fi
\ifx \botherref \undefined \def \botherref #1{#1}\fi
\ifx \url \undefined \def \url#1{\textsf{#1}}\fi
\ifx \bchapter \undefined \def \bchapter#1{#1}\fi
\ifx \bbook \undefined \def \bbook#1{#1}\fi
\ifx \bcomment \undefined \def \bcomment#1{#1}\fi
\ifx \oauthor \undefined \def \oauthor#1{#1}\fi
\ifx \citeauthoryear \undefined \def \citeauthoryear#1{#1}\fi
\ifx \endbibitem  \undefined \def \endbibitem {}\fi
\ifx \bconflocation  \undefined \def \bconflocation#1{#1}\fi
\ifx \arxivurl  \undefined \def \arxivurl#1{\textsf{#1}}\fi
\csname PreBibitemsHook\endcsname

%%% 1
\bibitem[\protect\citeauthoryear{Wetzel et~al.}{2020}]{wetzel2020twin}
\begin{barticle}
\bauthor{\bsnm{Wetzel}, \binits{S.J.}},
\bauthor{\bsnm{Ryczko}, \binits{K.}},
\bauthor{\bsnm{Melko}, \binits{R.G.}},
\bauthor{\bsnm{Tamblyn}, \binits{I.}}:
\batitle{Twin neural network regression}.
\bjtitle{Applied AI Letters}
\bvolume{3}(\bissue{4}),
\bfpage{78}
(\byear{2020})
\end{barticle}
\endbibitem

%%% 2
\bibitem[\protect\citeauthoryear{Tynes et~al.}{2021}]{tynes2021pairwise}
\begin{botherref}
\oauthor{\bsnm{Tynes}, \binits{M.}},
\oauthor{\bsnm{Gao}, \binits{W.}},
\oauthor{\bsnm{Burrill}, \binits{D.J.}},
\oauthor{\bsnm{Batista}, \binits{E.R.}},
\oauthor{\bsnm{Perez}, \binits{D.}},
\oauthor{\bsnm{Yang}, \binits{P.}},
\oauthor{\bsnm{Lubbers}, \binits{N.}}:
{Pairwise difference regression: A ML meta-algorithm for improved prediction
  and uncertainty quantification in chemical search}.
J. Chem. Inf. Model.
\textbf{61}
(2021)
\end{botherref}
\endbibitem

%%% 3
\bibitem[\protect\citeauthoryear{Xie et~al.}{2022}]{xie2022automated}
\begin{bchapter}
\bauthor{\bsnm{Xie}, \binits{J.}},
\bauthor{\bsnm{Kaiwei}, \binits{C.}},
\bauthor{\bsnm{Kong}, \binits{L.}},
\bauthor{\bsnm{Zhou}, \binits{J.}},
\bauthor{\bsnm{Qu}, \binits{W.}}:
\bctitle{Automated essay scoring via pairwise contrastive regression}.
In: \bbtitle{Proc.\ of the 29th COLING}
(\byear{2022})
\end{bchapter}
\endbibitem

%%% 4
\bibitem[\protect\citeauthoryear{Corbara et~al.}{2023}]{corbara2023same}
\begin{barticle}
\bauthor{\bsnm{Corbara}, \binits{S.}},
\bauthor{\bsnm{Moreo}, \binits{A.}},
\bauthor{\bsnm{Sebastiani}, \binits{F.}}:
\batitle{Same or different? diff-vectors for authorship analysis}.
\bjtitle{ACM Transactions on KDD}
\bvolume{18}(\bissue{1}),
\bfpage{1}--\blpage{36}
(\byear{2023})
\end{barticle}
\endbibitem

%%% 5
\bibitem[\protect\citeauthoryear{Belaid et~al.}{2024a}]{belaid2024pairwise}
\begin{bchapter}
\bauthor{\bsnm{Belaid}, \binits{M.K.}},
\bauthor{\bsnm{Rabus}, \binits{M.}},
\bauthor{\bsnm{H{\"u}llermeier}, \binits{E.}}:
\bctitle{Pairwise difference learning for classification}.
In: \bbtitle{DS},
pp. \bfpage{284}--\blpage{299}
(\byear{2024}).
\bcomment{Springer}
\end{bchapter}
\endbibitem

%%% 6
\bibitem[\protect\citeauthoryear{Belaid et~al.}{2024b}]{belaid2024weighted}
\begin{bchapter}
\bauthor{\bsnm{Belaid}, \binits{M.K.}},
\bauthor{\bsnm{Wibiral}, \binits{T.}},
\bauthor{\bsnm{Rabus}, \binits{M.}},
\bauthor{\bsnm{H{\"u}llermeier}, \binits{E.}}:
\bctitle{Weighted {Pairwise Difference Learning}}.
In: \bbtitle{Proc.\ of LWDA'24}.
\bpublisher{CEUR-WS.org},
\blocation{Würzburg}
(\byear{2024})
\end{bchapter}
\endbibitem

%%% 7
\bibitem[\protect\citeauthoryear{Wetzel}{2023}]{wetzel2023how}
\begin{botherref}
\oauthor{\bsnm{Wetzel}, \binits{S.J.}}:
How to Get the Most out of {Twinned Regression Methods}.
arXiv
(2023)
\end{botherref}
\endbibitem

%%% 8
\bibitem[\protect\citeauthoryear{Fralish et~al.}{}]{fralish2023deepdelta}
\begin{botherref}
\oauthor{\bsnm{Fralish}, \binits{Z.}},
\oauthor{\bsnm{Chen}, \binits{A.}},
\oauthor{\bsnm{Skaluba}, \binits{P.}},
\oauthor{\bsnm{Reker}, \binits{D.}}:
{DeepDelta: predicting ADMET improvements of molecular derivatives with deep
  learning}.
J. Cheminform.
\textbf{15}(1)
\end{botherref}
\endbibitem

%%% 9
\bibitem[\protect\citeauthoryear{Spiers
  et~al.}{2023}]{spiers2023physicochemical}
\begin{botherref}
\oauthor{\bsnm{Spiers}, \binits{R.C.}},
\oauthor{\bsnm{Norby}, \binits{C.}},
\oauthor{\bsnm{Kalivas}, \binits{J.H.}}:
Physicochemical {Responsive Integrated Similarity Measure} ({PRISM}) for a
  {Comprehensive Quantitative Perspective} of {Sample Similarity Dynamically
  Assessed} with {NIR Spectra}.
Anal. Chem.
(2023)
\end{botherref}
\endbibitem

%%% 10
\bibitem[\protect\citeauthoryear{Hu et~al.}{2023}]{hu2023exploring}
\begin{barticle}
\bauthor{\bsnm{Hu}, \binits{J.}},
\bauthor{\bsnm{Yang}, \binits{S.}},
\bauthor{\bsnm{Mao}, \binits{J.}},
\bauthor{\bsnm{Shi}, \binits{C.}},
\bauthor{\bsnm{Wang}, \binits{G.}},
\bauthor{\bsnm{Liu}, \binits{Y.}},
\bauthor{\bsnm{Pu}, \binits{X.}}:
\batitle{Exploring a general convolutional neural network-based prediction
  model for critical casting diameter of metallic glasses}.
\bjtitle{J. Alloys Compd.}
\bvolume{947},
\bfpage{169479}
(\byear{2023})
\end{barticle}
\endbibitem

%%% 11
\bibitem[\protect\citeauthoryear{Chen et~al.}{2023}]{chen2023benchmark}
\begin{botherref}
\oauthor{\bsnm{Chen}, \binits{Y.}},
\oauthor{\bsnm{Ou}, \binits{Y.}},
\oauthor{\bsnm{Zheng}, \binits{P.}},
\oauthor{\bsnm{Huang}, \binits{Y.}},
\oauthor{\bsnm{Ge}, \binits{F.}},
\oauthor{\bsnm{Dral}, \binits{P.O.}}:
Benchmark of general-purpose {ML-based} quantum mechanical method {AIQM1} on
  reaction barrier heights.
J. Chem. Phys.
\textbf{158}(7)
(2023)
\end{botherref}
\endbibitem

%%% 12
\bibitem[\protect\citeauthoryear{Wang and King}{2023}]{wang2023extrapolation}
\begin{bchapter}
\bauthor{\bsnm{Wang}, \binits{Y.}},
\bauthor{\bsnm{King}, \binits{R.D.}}:
\bctitle{Extrapolation is {Not} the {Same} as {Interpolation}}.
In: \bbtitle{26th Int.\ Conf.\, DS2023, Porto, Portugal, Oct 9--11, 2023,
  Proc.}
\bsertitle{Lecture Notes in Computer Science},
vol. \bseriesno{14276},
pp. \bfpage{277}--\blpage{292}.
\bpublisher{Springer},
\blocation{Cham}
(\byear{2023})
\end{bchapter}
\endbibitem

%%% 13
\bibitem[\protect\citeauthoryear{Radford et~al.}{2021}]{radford2021learning}
\begin{bchapter}
\bauthor{\bsnm{Radford}, \binits{A.}},
\bauthor{\bsnm{Kim}, \binits{J.W.}},
\bauthor{\bsnm{Hallacy}, \binits{C.}},
\bauthor{\bsnm{Ramesh}, \binits{A.}},
\bauthor{\bsnm{Goh}, \binits{G.}},
\bauthor{\bsnm{Agarwal}, \binits{S.}},
\bauthor{\bsnm{Sastry}, \binits{G.}},
\bauthor{\bsnm{Askell}, \binits{A.}},
\bauthor{\bsnm{Mishkin}, \binits{P.}},
\bauthor{\bsnm{Clark}, \binits{J.}}, \betal:
\bctitle{Learning transferable visual models from natural language
  supervision}.
In: \bbtitle{ICML},
pp. \bfpage{8748}--\blpage{8763}
(\byear{2021}).
\bcomment{PmLR}
\end{bchapter}
\endbibitem

%%% 14
\bibitem[\protect\citeauthoryear{Sensoy et~al.}{2018}]{sensoy2018evidential}
\begin{botherref}
\oauthor{\bsnm{Sensoy}, \binits{M.}},
\oauthor{\bsnm{Kaplan}, \binits{L.}},
\oauthor{\bsnm{Kandemir}, \binits{M.}}:
Evidential deep learning to quantify classification uncertainty.
NeurIPS
\textbf{31}
(2018)
\end{botherref}
\endbibitem

%%% 15
\bibitem[\protect\citeauthoryear{Sale et~al.}{2023}]{sale2023secondorder}
\begin{botherref}
\oauthor{\bsnm{Sale}, \binits{Y.}},
\oauthor{\bsnm{Hofman}, \binits{P.}},
\oauthor{\bsnm{Wimmer}, \binits{L.}},
\oauthor{\bsnm{H{\"u}llermeier}, \binits{E.}},
\oauthor{\bsnm{Nagler}, \binits{T.}}:
Second-{Order Uncertainty Quantification}: {Variance-Based Measures}.
arXiv
(2023)
\end{botherref}
\endbibitem

%%% 16
\bibitem[\protect\citeauthoryear{Yu et~al.}{2022}]{yu2022uncertainty}
\begin{barticle}
\bauthor{\bsnm{Yu}, \binits{J.}},
\bauthor{\bsnm{Wang}, \binits{D.}},
\bauthor{\bsnm{Zheng}, \binits{M.}}:
\batitle{Uncertainty quantification: {Can} we trust artificial intelligence in
  drug discovery?}
\bjtitle{iScience}
\bvolume{25}(\bissue{8}),
\bfpage{104814}
(\byear{2022})
\end{barticle}
\endbibitem

%%% 17
\bibitem[\protect\citeauthoryear{Valdenegro-Toro
  et~al.}{2024}]{valdenegro2024unified}
\begin{bchapter}
\bauthor{\bsnm{Valdenegro-Toro}, \binits{M.}},
\bauthor{\bsnm{Jong}, \binits{I.P.}},
\bauthor{\bsnm{Zullich}, \binits{M.}}:
\bctitle{Unified uncertainties: Combining input, data and model uncertainty
  into a single formulation}.
In: \bbtitle{Proceedings of the LatinX in AI Research Workshop at ICML 2024}
(\byear{2024})
\end{bchapter}
\endbibitem

%%% 18
\bibitem[\protect\citeauthoryear{H{\"u}llermeier and
  Waegeman}{2021}]{hullermeier2021aleatoric}
\begin{botherref}
\oauthor{\bsnm{H{\"u}llermeier}, \binits{E.}},
\oauthor{\bsnm{Waegeman}, \binits{W.}}:
Aleatoric and epistemic uncertainty in machine learning: An introduction to
  concepts and methods.
Mach Learn
\textbf{110}(3)
(2021)
\end{botherref}
\endbibitem

%%% 19
\bibitem[\protect\citeauthoryear{L{\"o}hr et~al.}{2024}]{lohr2024aleatoric}
\begin{bchapter}
\bauthor{\bsnm{L{\"o}hr}, \binits{T.}},
\bauthor{\bsnm{Ingrisch}, \binits{M.}},
\bauthor{\bsnm{H{\"u}llermeier}, \binits{E.}}:
\bctitle{Towards {Aleatoric} and {Epistemic Uncertainty} in {Medical Image
  Classification}}.
In: \beditor{\bsnm{Finkelstein}, \binits{J.}},
\beditor{\bsnm{Moskovitch}, \binits{R.}},
\beditor{\bsnm{Parimbelli}, \binits{E.}} (eds.)
\bbtitle{Artificial {Intelligence} in {Medicine}}.
\bpublisher{Springer},
\blocation{Cham}
(\byear{2024})
\end{bchapter}
\endbibitem

%%% 20
\bibitem[\protect\citeauthoryear{Wimmer et~al.}{2023}]{wimmer2023quantifying}
\begin{bchapter}
\bauthor{\bsnm{Wimmer}, \binits{L.}},
\bauthor{\bsnm{Sale}, \binits{Y.}},
\bauthor{\bsnm{Hofman}, \binits{P.}},
\bauthor{\bsnm{Bischl}, \binits{B.}},
\bauthor{\bsnm{H{\"u}llermeier}, \binits{E.}}:
\bctitle{Quantifying {Aleatoric} and {Epistemic} {Uncertainty} in {Machine}
  {Learning}: {Are} {Conditional} {Entropy} and {Mutual} {Information}
  {Appropriate} {Measures}?}
In: \bbtitle{Proc.\ of the 39th {Conf.} on {Uncertainty} in AI},
pp. \bfpage{2282}--\blpage{2292}.
\bpublisher{PMLR},
\blocation{Budapest, Hungary}
(\byear{2023})
\end{bchapter}
\endbibitem

%%% 21
\bibitem[\protect\citeauthoryear{Hou et~al.}{2024}]{hou2024decomposing}
\begin{bchapter}
\bauthor{\bsnm{Hou}, \binits{B.}},
\bauthor{\bsnm{Liu}, \binits{Y.}},
\bauthor{\bsnm{Qian}, \binits{K.}},
\bauthor{\bsnm{Andreas}, \binits{J.}},
\bauthor{\bsnm{Chang}, \binits{S.}},
\bauthor{\bsnm{Zhang}, \binits{Y.}}:
\bctitle{Decomposing {Uncertainty} for {Large Language Models} through {Input
  Clarification Ensembling}}.
In: \bbtitle{ICML},
pp. \bfpage{19023}--\blpage{19042}
(\byear{2024}).
\bcomment{PMLR}
\end{bchapter}
\endbibitem

%%% 22
\bibitem[\protect\citeauthoryear{Kramer et~al.}{2014}]{kramer2014matched}
\begin{barticle}
\bauthor{\bsnm{Kramer}, \binits{C.}},
\bauthor{\bsnm{Fuchs}, \binits{J.E.}},
\bauthor{\bsnm{Whitebread}, \binits{S.}},
\bauthor{\bsnm{Gedeck}, \binits{P.}},
\bauthor{\bsnm{Liedl}, \binits{K.R.}}:
\batitle{Matched {Molecular Pair Analysis}: {Significance} and the {Impact} of
  {Experimental Uncertainty}}.
\bjtitle{J. Med. Chem.}
\bvolume{57}(\bissue{9}),
\bfpage{3786}--\blpage{3802}
(\byear{2014})
\end{barticle}
\endbibitem

%%% 23
\bibitem[\protect\citeauthoryear{Jim{\'e}nez
  et~al.}{2019}]{jimenez-luna2019deltadelta}
\begin{barticle}
\bauthor{\bsnm{Jim{\'e}nez}, \binits{J.}},
\bauthor{\bsnm{{P{\'e}rez-Benito}}, \binits{L.}},
\bauthor{\bsnm{{Martinez-Rosell}}, \binits{G.}},
\bauthor{\bsnm{Sciabola}, \binits{S.}},
\bauthor{\bsnm{Torella}, \binits{R.}},
\bauthor{\bsnm{Tresadern}, \binits{G.}},
\bauthor{\bsnm{De~Fabritiis}, \binits{G.}}:
\batitle{{DeltaDelta} neural networks for lead optimization of small molecule
  potency}.
\bjtitle{Chemical science}
\bvolume{10}(\bissue{47}),
\bfpage{10911}--\blpage{10918}
(\byear{2019})
\end{barticle}
\endbibitem

%%% 24
\bibitem[\protect\citeauthoryear{Aha}{1997}]{aha_ll}
\begin{bchapter}
\bauthor{\bsnm{Aha}, \binits{D.W.}}:
\bctitle{Lazy learning}.
In: \bbtitle{Lazy Learning}.
\bpublisher{Springer},
\blocation{Dordrecht}
(\byear{1997})
\end{bchapter}
\endbibitem

%%% 25
\bibitem[\protect\citeauthoryear{Lakshminarayanan et~al.}{2017a}]{laks_sa17}
\begin{bchapter}
\bauthor{\bsnm{Lakshminarayanan}, \binits{B.}},
\bauthor{\bsnm{Pritzel}, \binits{A.}},
\bauthor{\bsnm{Blundell}, \binits{C.}}:
\bctitle{Simple and scalable predictive uncertainty estimation using deep
  ensembles}.
In: \bbtitle{Proc.\ 31st Conf.\ on NeurIPS},
\bconflocation{Long Beach, California, USA}
(\byear{2017})
\end{bchapter}
\endbibitem

%%% 26
\bibitem[\protect\citeauthoryear{Lakshminarayanan
  et~al.}{2017b}]{lakshminarayanan2017simple}
\begin{botherref}
\oauthor{\bsnm{Lakshminarayanan}, \binits{B.}},
\oauthor{\bsnm{Pritzel}, \binits{A.}},
\oauthor{\bsnm{Blundell}, \binits{C.}}:
Simple and scalable predictive uncertainty estimation using deep ensembles.
NeurIPS
\textbf{30}
(2017)
\end{botherref}
\endbibitem

%%% 27
\bibitem[\protect\citeauthoryear{Shelmanov et~al.}{2021}]{shelmanov2021active}
\begin{bchapter}
\bauthor{\bsnm{Shelmanov}, \binits{A.}},
\bauthor{\bsnm{Puzyrev}, \binits{D.}},
\bauthor{\bsnm{Kupriyanova}, \binits{L.}},
\bauthor{\bsnm{Khromov}, \binits{N.}},
\bauthor{\bsnm{Dylov}, \binits{D.}},
\bauthor{\bsnm{Panchenko}, \binits{A.}},
\bauthor{\bsnm{Belyakov}, \binits{D.}},
\bauthor{\bsnm{Larionov}, \binits{D.}},
\bauthor{\bsnm{Artemova}, \binits{E.}},
\bauthor{\bsnm{Kozlova}, \binits{O.}}:
\bctitle{Active learning for sequence tagging with deep pre-trained models and
  bayesian uncertainty estimates}.
In: \bbtitle{16th EACL},
pp. \bfpage{1698}--\blpage{1712}
(\byear{2021})
\end{bchapter}
\endbibitem

%%% 28
\bibitem[\protect\citeauthoryear{Nguyen et~al.}{2022}]{nguyen2022how}
\begin{barticle}
\bauthor{\bsnm{Nguyen}, \binits{V.-L.}},
\bauthor{\bsnm{Shaker}, \binits{M.H.}},
\bauthor{\bsnm{H{\"u}llermeier}, \binits{E.}}:
\batitle{How to measure uncertainty in uncertainty sampling for active
  learning}.
\bjtitle{Mach Learn}
\bvolume{111}(\bissue{1}),
\bfpage{89}--\blpage{122}
(\byear{2022})
\end{barticle}
\endbibitem

%%% 29
\bibitem[\protect\citeauthoryear{Hoffer et~al.}{2023}]{hoffer2023robust}
\begin{barticle}
\bauthor{\bsnm{Hoffer}, \binits{J.G.}},
\bauthor{\bsnm{Ranftl}, \binits{S.}},
\bauthor{\bsnm{Geiger}, \binits{B.C.}}:
\batitle{Robust {Bayesian} target value optimization}.
\bjtitle{Computers \& Industrial Engineering}
\bvolume{180},
\bfpage{109279}
(\byear{2023})
\end{barticle}
\endbibitem

%%% 30
\bibitem[\protect\citeauthoryear{Stanton et~al.}{2023}]{stanton2023bayesian}
\begin{bchapter}
\bauthor{\bsnm{Stanton}, \binits{S.}},
\bauthor{\bsnm{Maddox}, \binits{W.}},
\bauthor{\bsnm{Wilson}, \binits{A.G.}}:
\bctitle{Bayesian optimization with conformal prediction sets}.
In: \bbtitle{AISTATS},
pp. \bfpage{959}--\blpage{986}.
\bpublisher{PMLR},
\blocation{Bordeaux, France}
(\byear{2023})
\end{bchapter}
\endbibitem

%%% 31
\bibitem[\protect\citeauthoryear{Ash}{2012}]{ash2012information}
\begin{bbook}
\bauthor{\bsnm{Ash}, \binits{R.B.}}:
\bbtitle{Information Theory}.
\bpublisher{Courier Corporation},
\blocation{Hoboken, NJ}
(\byear{2012})
\end{bbook}
\endbibitem

%%% 32
\bibitem[\protect\citeauthoryear{Depeweg
  et~al.}{2018}]{depeweg2018decomposition}
\begin{bchapter}
\bauthor{\bsnm{Depeweg}, \binits{S.}},
\bauthor{\bsnm{{Hernandez-Lobato}}, \binits{J.-M.}},
\bauthor{\bsnm{{Doshi-Velez}}, \binits{F.}},
\bauthor{\bsnm{Udluft}, \binits{S.}}:
\bctitle{Decomposition of uncertainty in {Bayesian} deep learning for efficient
  and risk-sensitive learning}.
In: \bbtitle{ICML},
pp. \bfpage{1184}--\blpage{1193}.
\bpublisher{PMLR},
\blocation{Stockholm, Sweden}
(\byear{2018})
\end{bchapter}
\endbibitem

%%% 33
\bibitem[\protect\citeauthoryear{Olson et~al.}{2016}]{olson2016evaluation}
\begin{bchapter}
\bauthor{\bsnm{Olson}, \binits{R.S.}},
\bauthor{\bsnm{Bartley}, \binits{N.}},
\bauthor{\bsnm{Urbanowicz}, \binits{R.J.}},
\bauthor{\bsnm{Moore}, \binits{J.H.}}:
\bctitle{Evaluation of a tree-based pipeline optimization tool for automating
  data science}.
In: \bbtitle{Proc.\ GECCO},
pp. \bfpage{485}--\blpage{492}
(\byear{2016})
\end{bchapter}
\endbibitem

%%% 34
\bibitem[\protect\citeauthoryear{Bischl et~al.}{2017}]{bischl2017openml}
\begin{bchapter}
\bauthor{\bsnm{Bischl}, \binits{B.}},
\bauthor{\bsnm{Casalicchio}, \binits{G.}},
\bauthor{\bsnm{Feurer}, \binits{M.}},
\bauthor{\bsnm{Gijsbers}, \binits{P.}},
\bauthor{\bsnm{Hutter}, \binits{F.}},
\bauthor{\bsnm{Lang}, \binits{M.}},
\bauthor{\bsnm{Mantovani}, \binits{R.G.}},
\bauthor{\bsnm{Rijn}, \binits{J.N.}},
\bauthor{\bsnm{Vanschoren}, \binits{J.}}:
\bctitle{Openml benchmarking suites}.
In: \bbtitle{35th NeurIPS Datasets and Benchmarks Track (Round 2)}
(\byear{2017})
\end{bchapter}
\endbibitem

%%% 35
\bibitem[\protect\citeauthoryear{Pedregosa
  et~al.}{2011}]{pedregosa2011scikitlearn}
\begin{barticle}
\bauthor{\bsnm{Pedregosa}, \binits{F.}},
\bauthor{\bsnm{Varoquaux}, \binits{G.}},
\bauthor{\bsnm{Gramfort}, \binits{A.}},
\bauthor{\bsnm{Michel}, \binits{V.}},
\bauthor{\bsnm{Thirion}, \binits{B.}},
\bauthor{\bsnm{Grisel}, \binits{O.}},
\bauthor{\bsnm{Blondel}, \binits{M.}},
\bauthor{\bsnm{Prettenhofer}, \binits{P.}},
\bauthor{\bsnm{Weiss}, \binits{R.}},
\bauthor{\bsnm{Dubourg}, \binits{V.}}, \betal:
\batitle{Scikit-learn: {ML} in {Python}}.
\bjtitle{the Journal of ML research}
\bvolume{12},
\bfpage{2825}--\blpage{2830}
(\byear{2011})
\end{barticle}
\endbibitem

%%% 36
\bibitem[\protect\citeauthoryear{H{\"u}llermeier and
  Brinker}{2008}]{hullermeier2008learning}
\begin{barticle}
\bauthor{\bsnm{H{\"u}llermeier}, \binits{E.}},
\bauthor{\bsnm{Brinker}, \binits{K.}}:
\batitle{Learning valued preference structures for solving classification
  problems}.
\bjtitle{Fuzzy Sets and Systems}
\bvolume{159}(\bissue{18}),
\bfpage{2337}--\blpage{2352}
(\byear{2008})
\end{barticle}
\endbibitem

%%% 37
\bibitem[\protect\citeauthoryear{Goodman and Kruskal}{1979}]{good_mo}
\begin{bbook}
\bauthor{\bsnm{Goodman}, \binits{L.A.}},
\bauthor{\bsnm{Kruskal}, \binits{W.H.}}:
\bbtitle{Measures of Association for Cross Classifications}.
\bpublisher{Springer},
\blocation{New York}
(\byear{1979})
\end{bbook}
\endbibitem

%%% 38
\bibitem[\protect\citeauthoryear{Naeini et~al.}{2015}]{naeini2015obtaining}
\begin{bchapter}
\bauthor{\bsnm{Naeini}, \binits{M.P.}},
\bauthor{\bsnm{Cooper}, \binits{G.F.}},
\bauthor{\bsnm{Hauskrecht}, \binits{M.}}:
\bctitle{Obtaining well calibrated probabilities using bayesian binning}.
In: \beditor{\bsnm{Bonet}, \binits{B.}},
\beditor{\bsnm{Koenig}, \binits{S.}} (eds.)
\bbtitle{Proc.\ of the 29th {AAAI}},
\bconflocation{Texas}
(\byear{2015})
\end{bchapter}
\endbibitem

%%% 39
\bibitem[\protect\citeauthoryear{H{\"u}hn and
  H{\"u}llermeier}{2008}]{huhn2008fr3}
\begin{barticle}
\bauthor{\bsnm{H{\"u}hn}, \binits{J.C.}},
\bauthor{\bsnm{H{\"u}llermeier}, \binits{E.}}:
\batitle{{FR3}: {A} fuzzy rule learner for inducing reliable classifiers}.
\bjtitle{IEEE Transactions on Fuzzy Systems}
\bvolume{17}(\bissue{1}),
\bfpage{138}--\blpage{149}
(\byear{2008})
\end{barticle}
\endbibitem

\end{thebibliography}

\begin{appendices}
\section{Choice of Base Learners}\label{sec:method_compatibility}

In principle, any (probabilistic) binary classifier can serve as a base learner for PDL, making it a flexible wrapper applicable to both binary and multiclass settings. In practice, however, some classifiers are more appropriate than others.

First, even if the original dataset \(\mathcal{D}\) is i.i.d., the paired dataset \(\mathcal{D}_{pair}\) is not, since an instance \(x_i\) can appear in multiple pairs—similar to scenarios in metric learning. While many algorithms are robust to such dependency violations, some methods may be more sensitive than others.

Second, the joint feature representation \(z = \phi(x, x')\) (e.g., the concatenation of \(x\), \(x'\), and \(x-x'\)) introduces perfect multicollinearity, which can hinder certain models, notably linear ones \cite[p.8]{tynes2021pairwise}, though others handle it more effectively.

Although a detailed analysis of base learner suitability is beyond our scope, our findings suggest that non-parametric methods are generally more robust and deliver superior performance. Accordingly, our experimental evaluation predominantly employs tree-based methods, which also benefit from fast training times.

\section{Complexity}
Assume that a base learner operates with complexity \(\mathcal{O}(p(N, M, F, K))\), where \(p(\cdot)\) is polynomial in the number of training examples (\(N\)), test examples (\(M\)), input features (\(F\)), and output classes (\(K\)). Within the PDL framework, this complexity becomes \(\mathcal{O}(p(N^2, 2MA, 3F, 2))\) due to several factors:
\begin{itemize}
    \item The training data are expanded into \(N^2\) pairs.
    \item The joint feature representation combines \(F\) features from each instance and their difference, resulting in \(3F\) features.
    \item Each test point is paired with \(A\) anchors and, to ensure symmetry, predictions are computed twice, yielding \(2MA\) predictions.
    \item The classification task is reduced to a binary decision (i.e., \(K=2\)).
\end{itemize}

\section{Extended performance analysis}
\subsection{Analyzing the Unique Contribution}

Although PDL classifiers often outperform baseline models in head-to-head comparisons, the primary goal is to achieve the best performance on each dataset. Prior to introducing PDC, the highest average Macro F1 score across 99 datasets was \(0.8112 \pm 0.0035\). With PDC, improvements were observed on 75 datasets, raising the record to \(0.8243 \pm 0.0031\). This improvement underscores PDC's distinct contribution to the field, as its incremental gain of \(0.0131\) significantly exceeds that of other estimators. In fact, PDC's contribution is 32 times greater than that of the best baseline.

\begin{table}[tbh]
\caption{Unique contribution of each estimator to the average Macro F1 score using the best optimized model on each dataset.} \label{tab:exp2_contirbution_f1}
\begin{tabular}{lrr}
Estimator & \textbf{Unique contribution}& \textbf{Wins}\\
\hline
ExtraTree & 0 & 0 \\
RandomForest & 0.00002 & 1 \\
Bagging & 0.00004 & 2 \\
DecisionTree & 0.00020 & 10 \\
ExtraTrees & 0.00041 & 9 \\
\textbf{PDC} & \textbf{0.01312} & \textbf{75} \\
\hline
\end{tabular}
\end{table}

\subsection{Analyzing Overfitting}

PDL classifiers exhibit a notable reduction in overfitting. Across 199 cross-validation runs where both the baseline and PDL classifiers showed non-significant differences in training Macro F1 scores, PDL consistently achieved a smaller train-test gap. This trend remains evident even without restricting the analysis to cases with similar training scores. Furthermore, an examination of the impact of dataset size on overfitting (\cref{tab:overfit}) shows that the PDL method consistently enhances test performance and mitigates overfitting, regardless of dataset size. These results highlight the robustness of the PDL approach, particularly in small datasets, and confirm its efficacy across diverse data scales.

\begin{table}[tbh]
\caption{Comparing test Macro F1 on the subset of runs where train scores are not significantly different.} \label{tab:overfit}
\begin{tabular}{lr|rr|rr|rr}
   &\textbf{\# CV} & \multicolumn{2}{c}{\textbf{Baseline Macro F1}}  & \multicolumn{2}{c}{\textbf{PDC Macro F1}} & \textbf{Test} & \textbf{Test}  \\
Estimator & \textbf{runs} & \textbf{Train} & \textbf{Test} & \textbf{Train} & \textbf{Test} &$\Delta F1$ & \textbf{p-value} \\
\hline
DecisionTree & 14 & 0.950 & 0.884 & 0.955 & \textbf{0.895} & 0.011 & $10^{-05}$ \\
ExtraTree & 11 & 0.915 & 0.844 & 0.924 & \textbf{0.861} & 0.017 & $10^{-04}$ \\
ExtraTrees & 26 & 0.985 & 0.828 & 0.991 & \textbf{0.853} & 0.025 & $10^{-16}$ \\
RandomForest & 18 & 0.992 & 0.855 & 0.997 & \textbf{0.881} & 0.026 & $10^{-11}$ \\
Bagging & 20 & 0.998 & 0.835 & 0.999 & \textbf{0.859} & 0.024 & $10^{-15}$ \\
\hline
\end{tabular}
\end{table} %

\section{Visualizing the Uncertainty}\label{annex:2d}
The "X" dataset is a synthetic dataset specifically designed to facilitate the visualization of both classifier decision boundaries and their associated uncertainty estimates. It is a two-class classification dataset characterized by non-linearly separable distributions. Each class is generated as a mixture of Gaussians, with one class following a diagonal alignment and the other along the anti-diagonal. This structured arrangement inherently creates a well-defined central region exhibiting high aleatoric uncertainty (AU), where class overlap is maximal. Additionally, the dataset contains four peripheral regions of low-density support, corresponding to areas of high EU, making it well-suited for benchmarking uncertainty-aware classification models.
The uncertainty patterns for the two probabilistic learners (DecisionTree and ExtraTree) are nearly indistinguishable, as are those for RandomForest and ExtraTrees. For clarity and conciseness, we therefore omit the redundant plots for ExtraTree and ExtraTrees.

\clearpage
\section{Extended Benchmark Results}\label{annex:extended_results}
\subsection{Accuracy-Rejection Curves}
To assess the informativeness of our uncertainty estimates, we construct Accuracy-Rejection Curves (ARCs). For each rejection percentage, we discard predictions corresponding to the most uncertain test instances and compute the accuracy on the remaining subset \citep{huhn2008fr3}. Under an effective UQ scheme, the accuracy should increase monotonically as the fraction of rejected instances rises, because misclassifications are more prevalent among highly uncertain predictions. In contrast, if the uncertainty estimates were non-informative (e.g., derived from random values), the ARC would remain nearly flat.
We show 5 example datasets.

\begin{figure}[h]
    \includegraphics[width=0.4\linewidth]{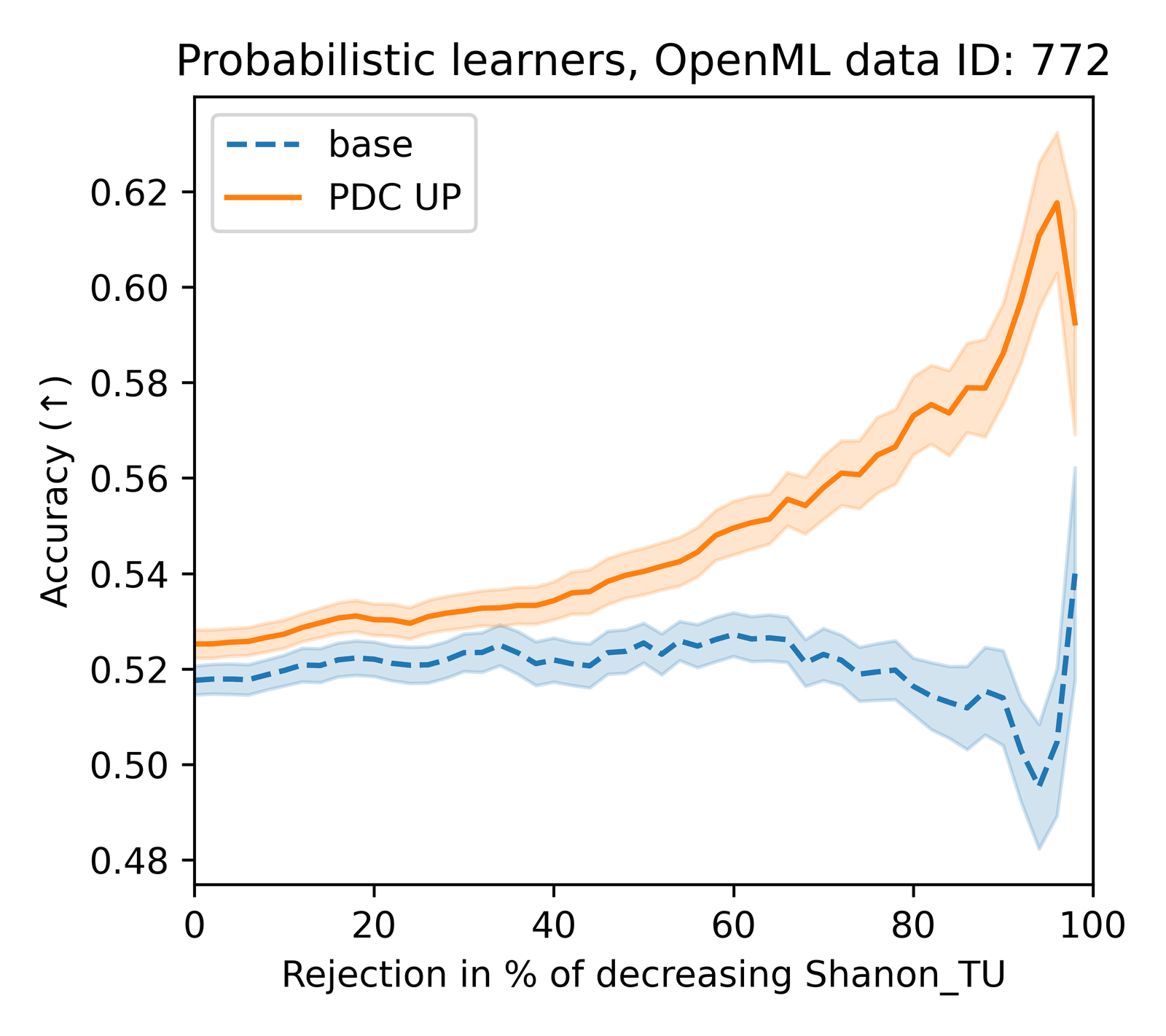}
    \includegraphics[width=0.4\linewidth]{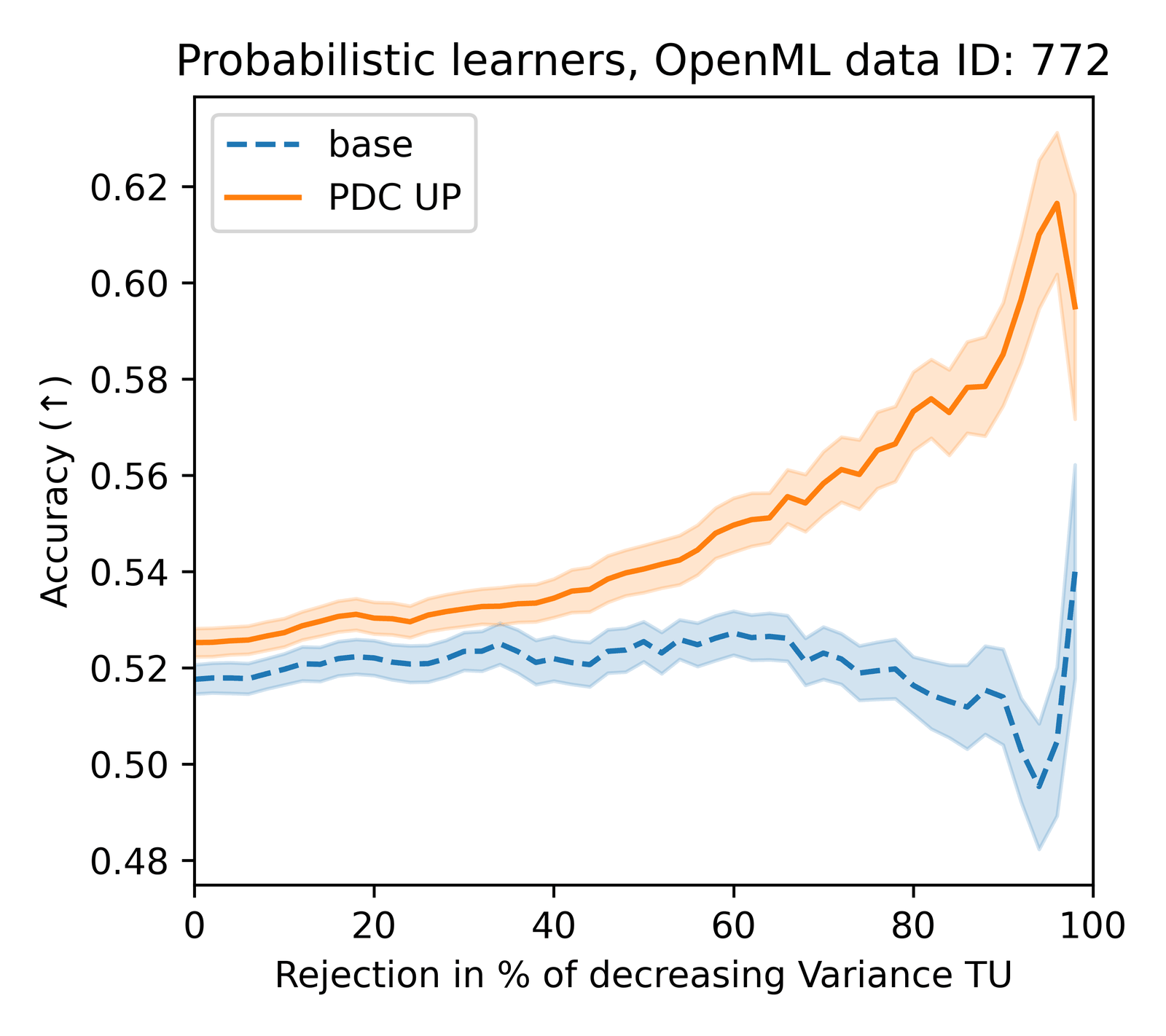}

    \includegraphics[width=0.4\linewidth]{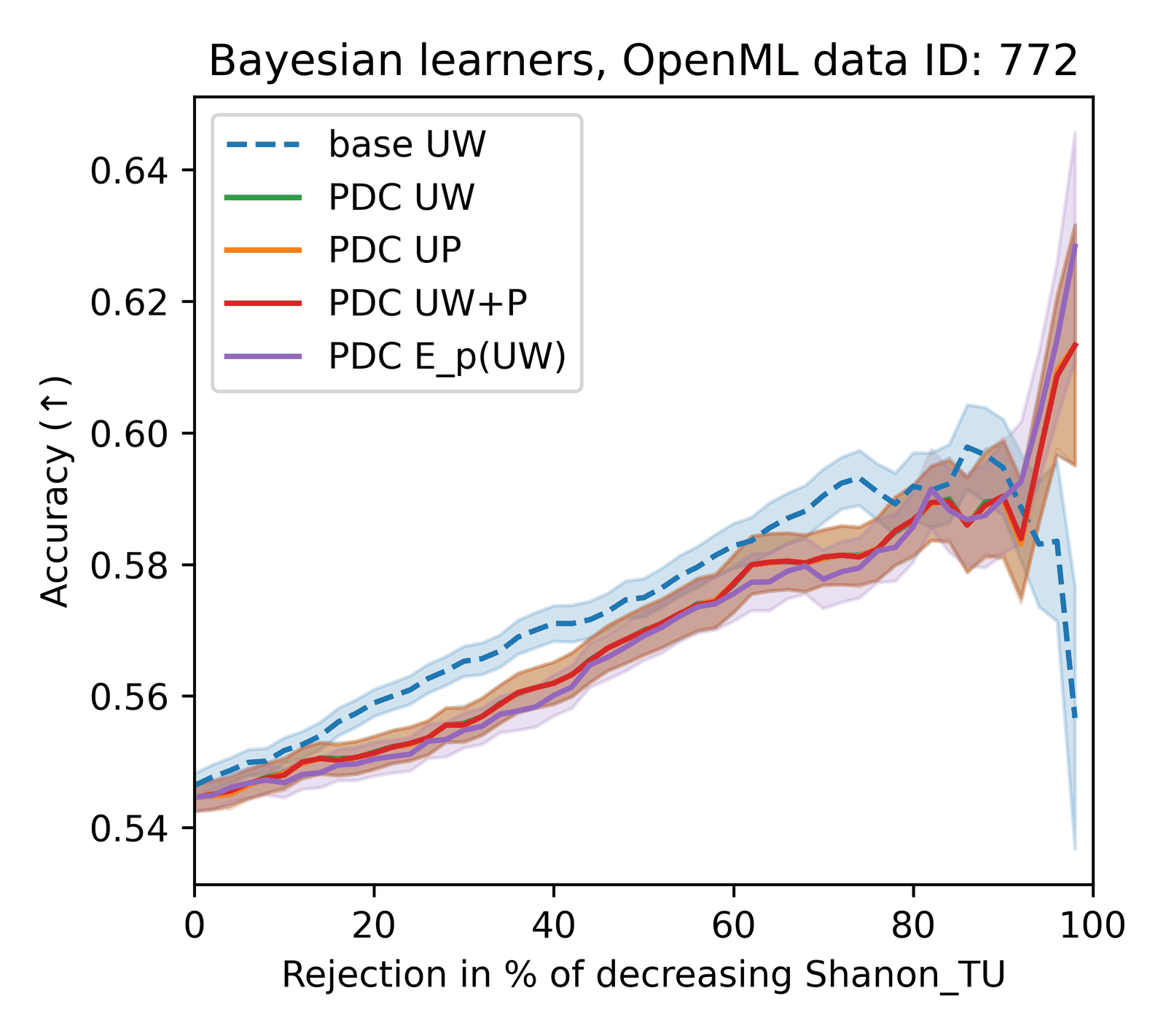}
    \includegraphics[width=0.4\linewidth]{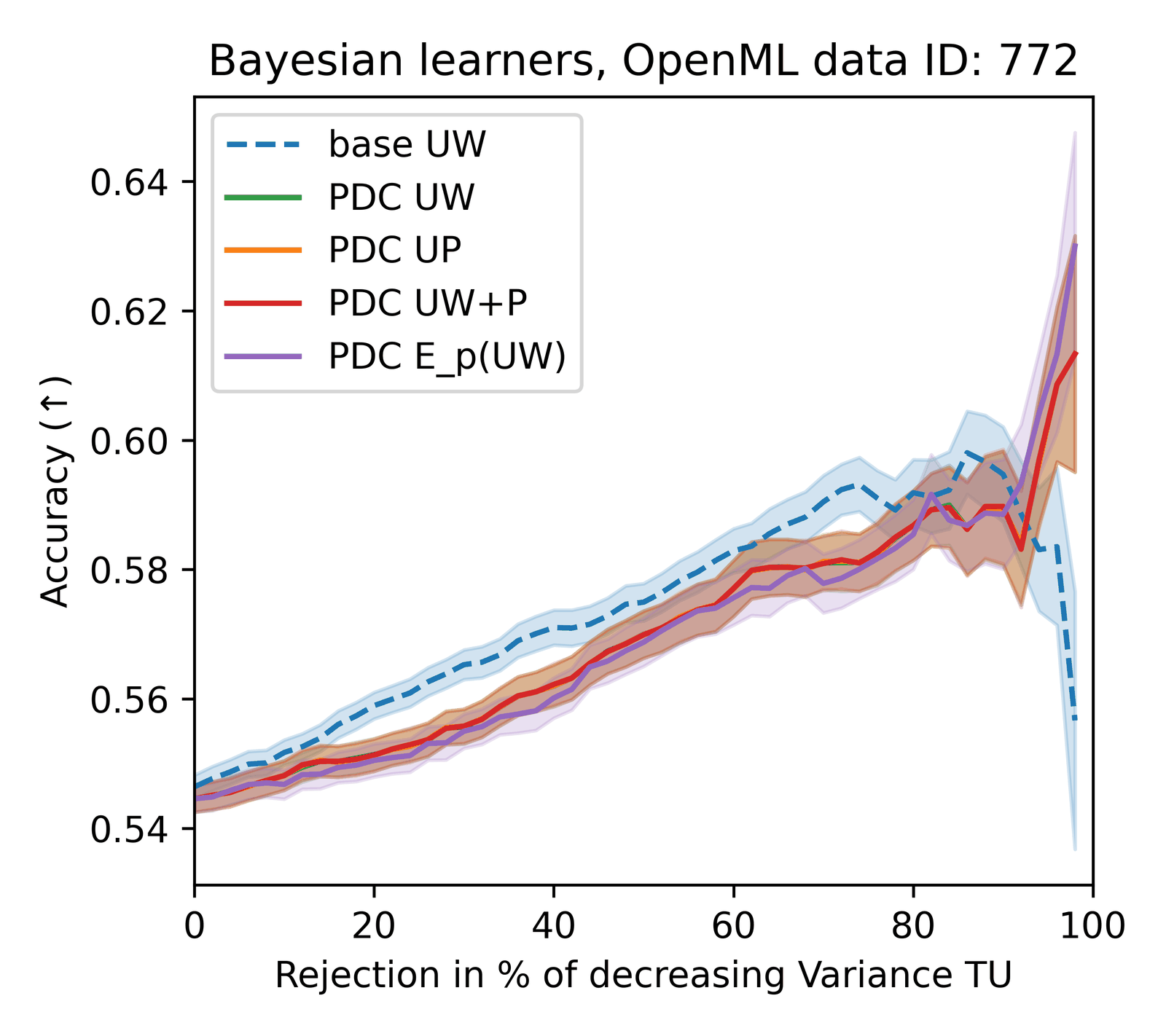}
    \caption{Accuracy rejection curve for the Dataset 772.}
\end{figure}
\begin{figure}[h]
    \includegraphics[width=0.4\linewidth]{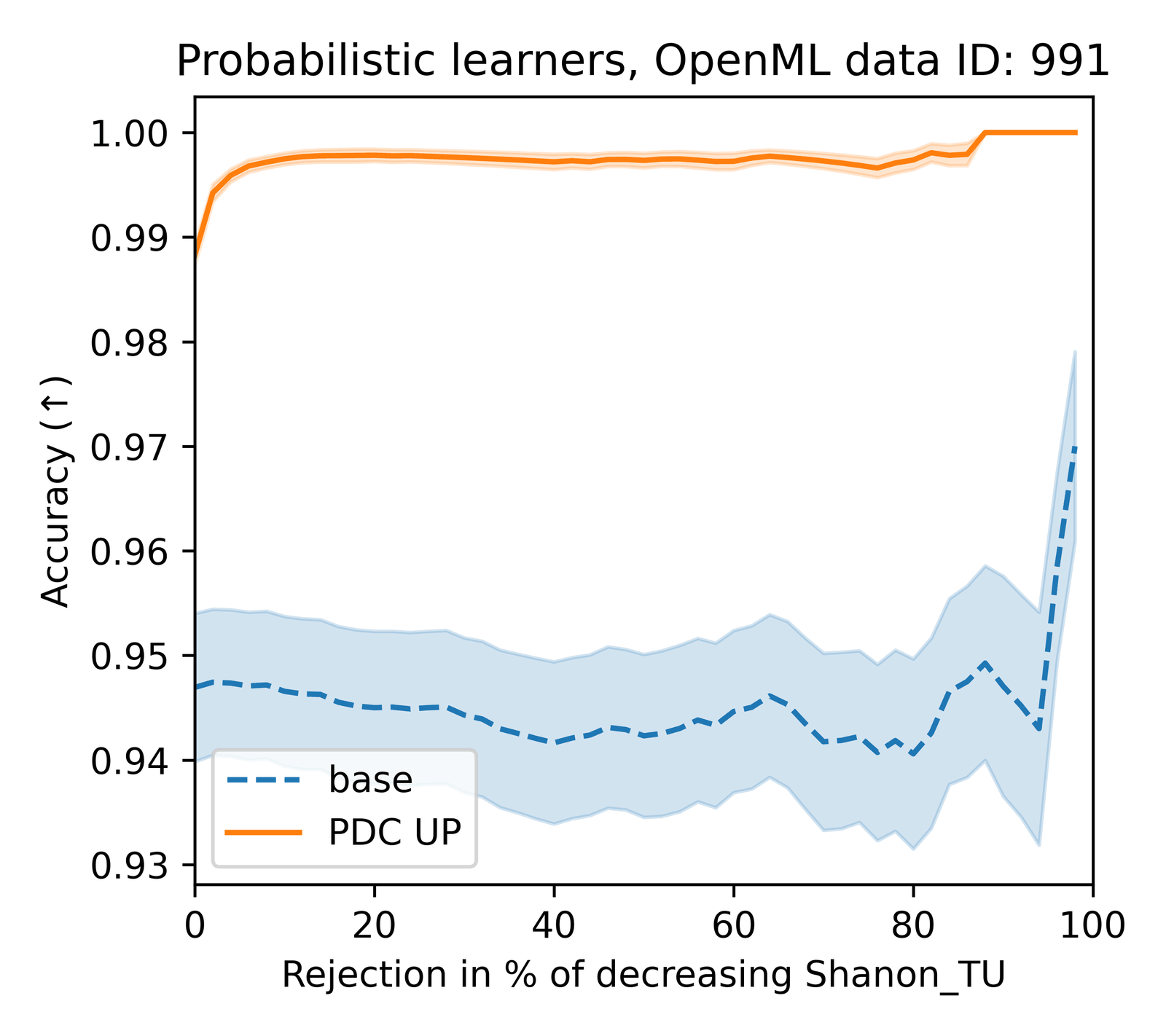}
    \includegraphics[width=0.4\linewidth]{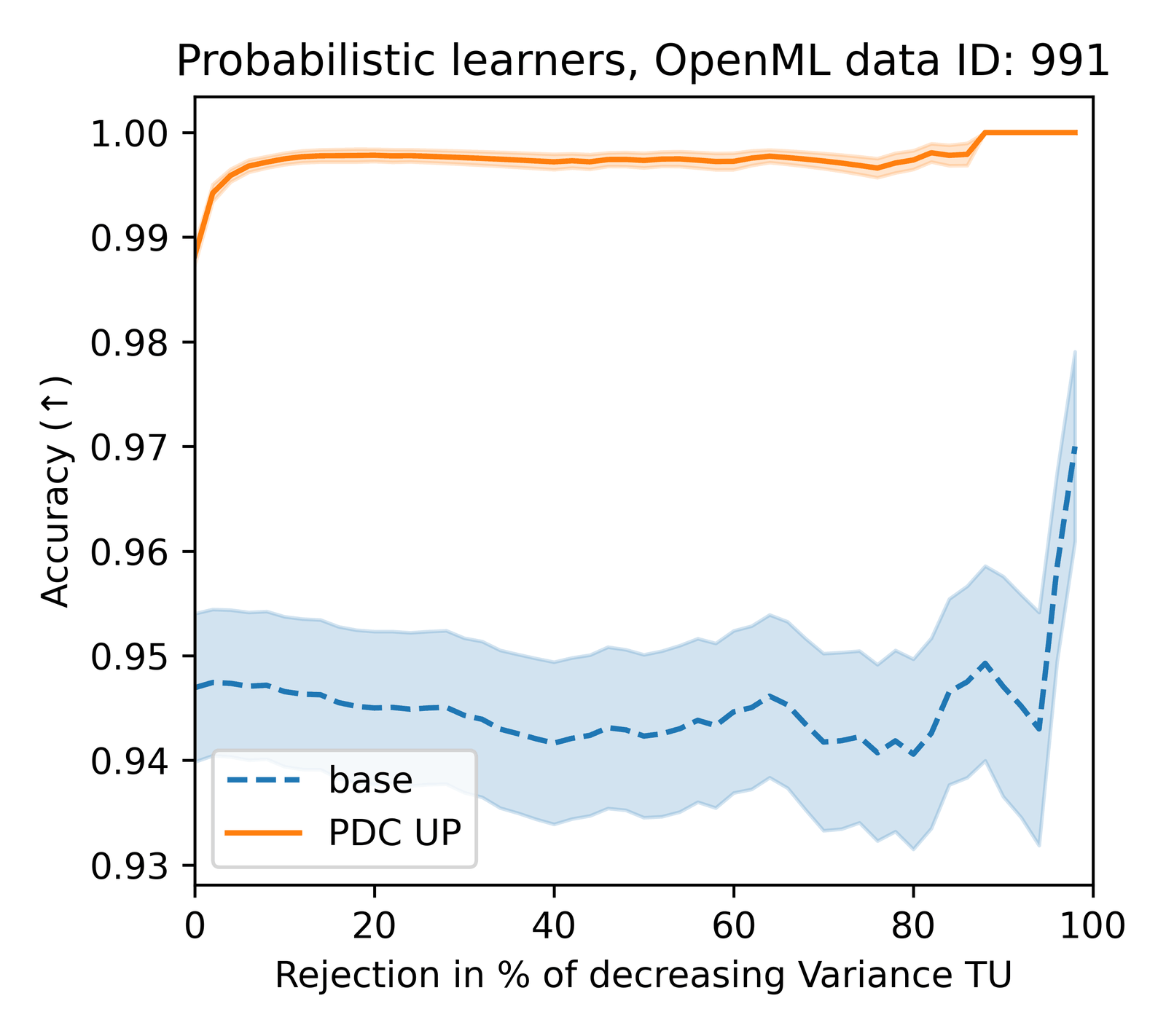}

    \includegraphics[width=0.4\linewidth]{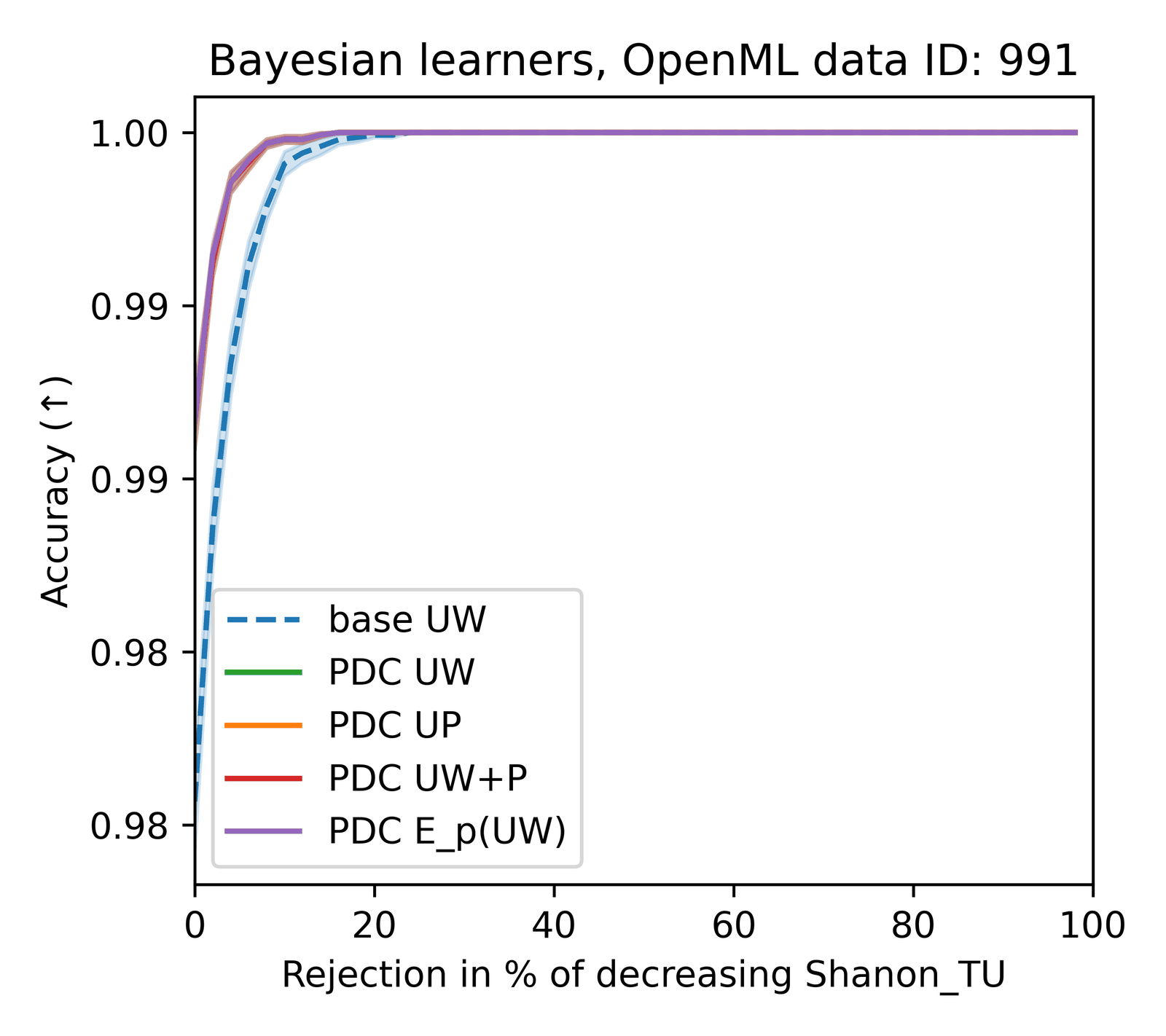}
    \includegraphics[width=0.4\linewidth]{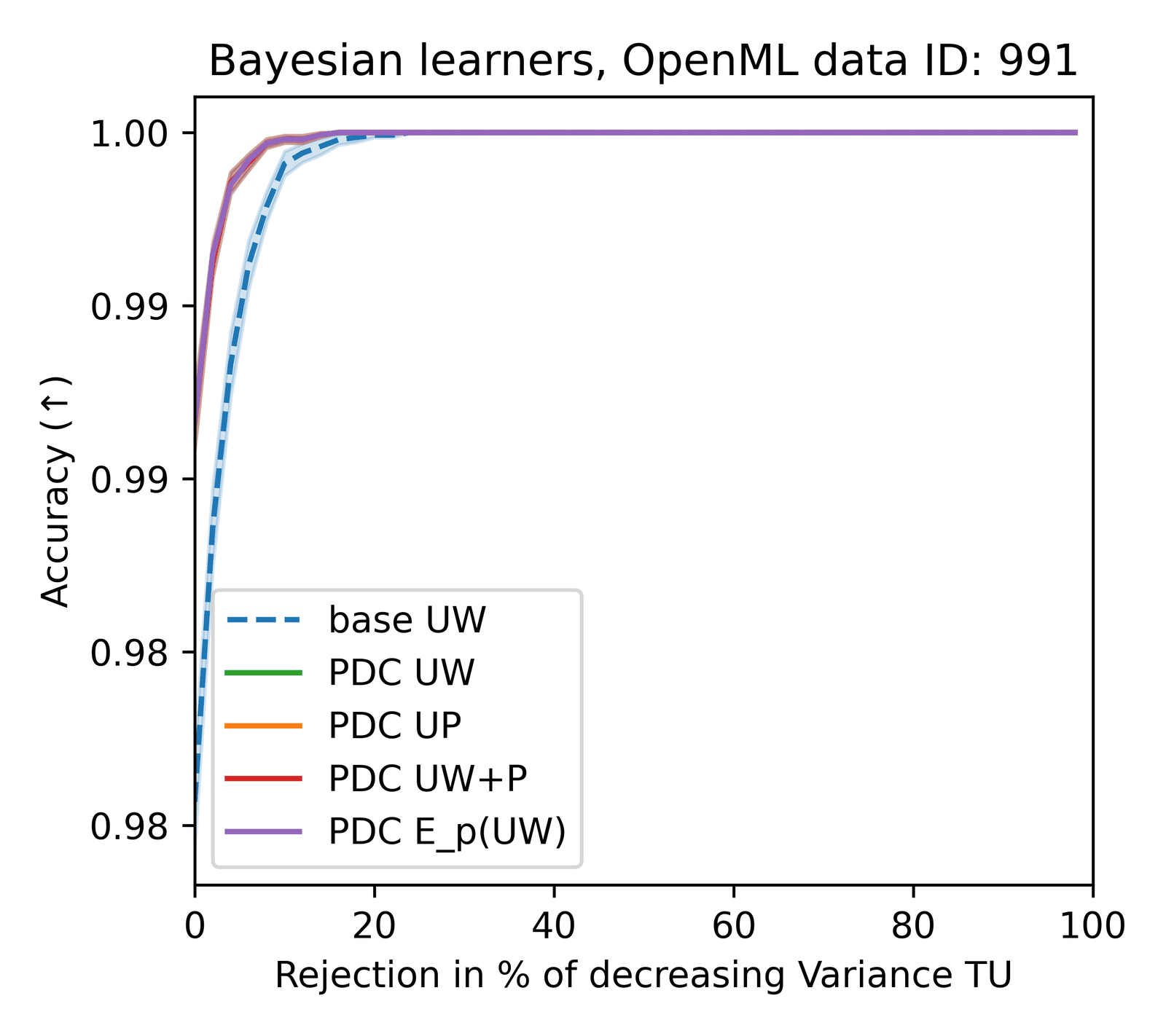}
    \caption{Accuracy rejection curve for the Dataset 991.}
\end{figure}
\begin{figure}[h]
    \includegraphics[width=0.4\linewidth]{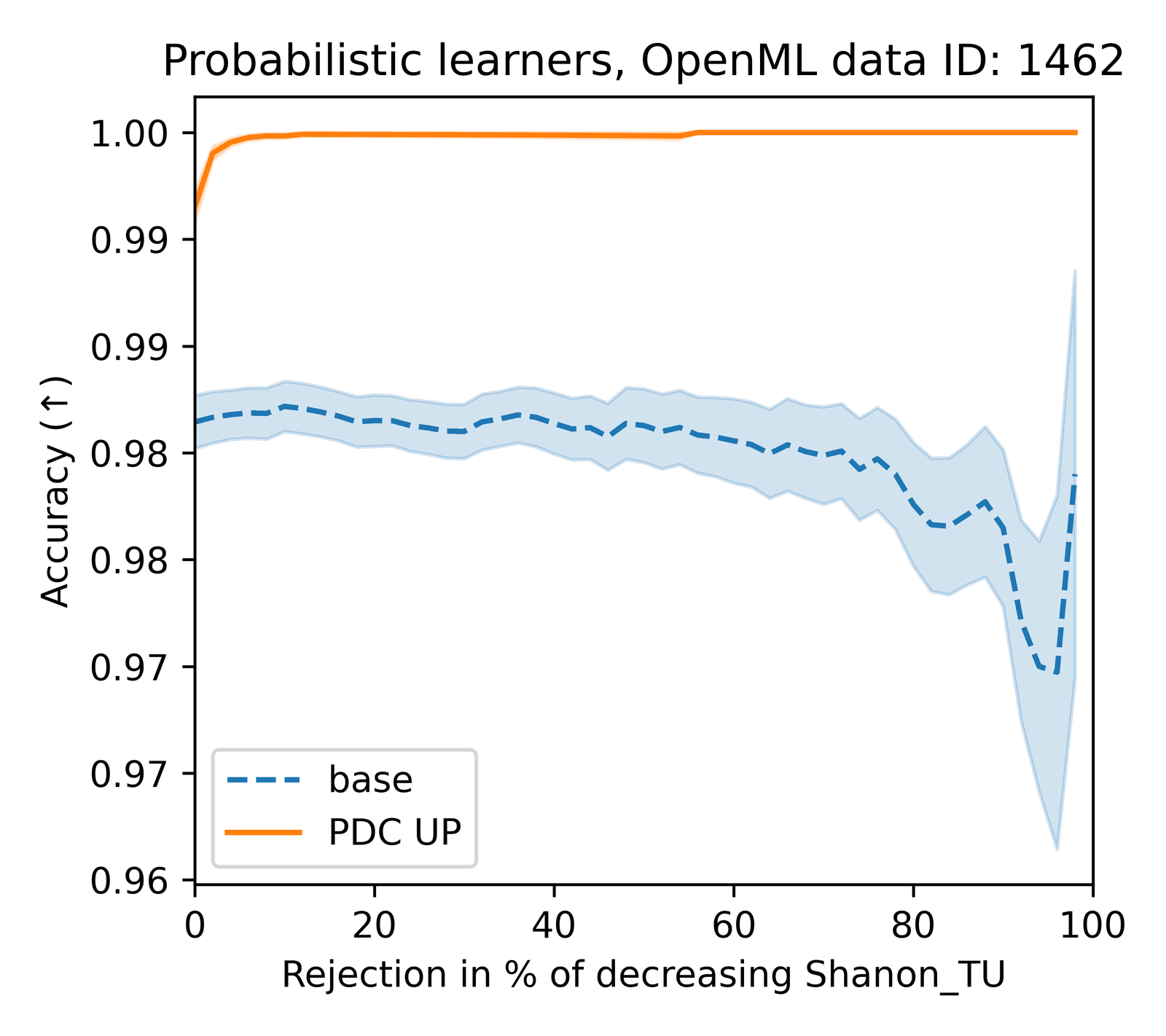}
    \includegraphics[width=0.4\linewidth]{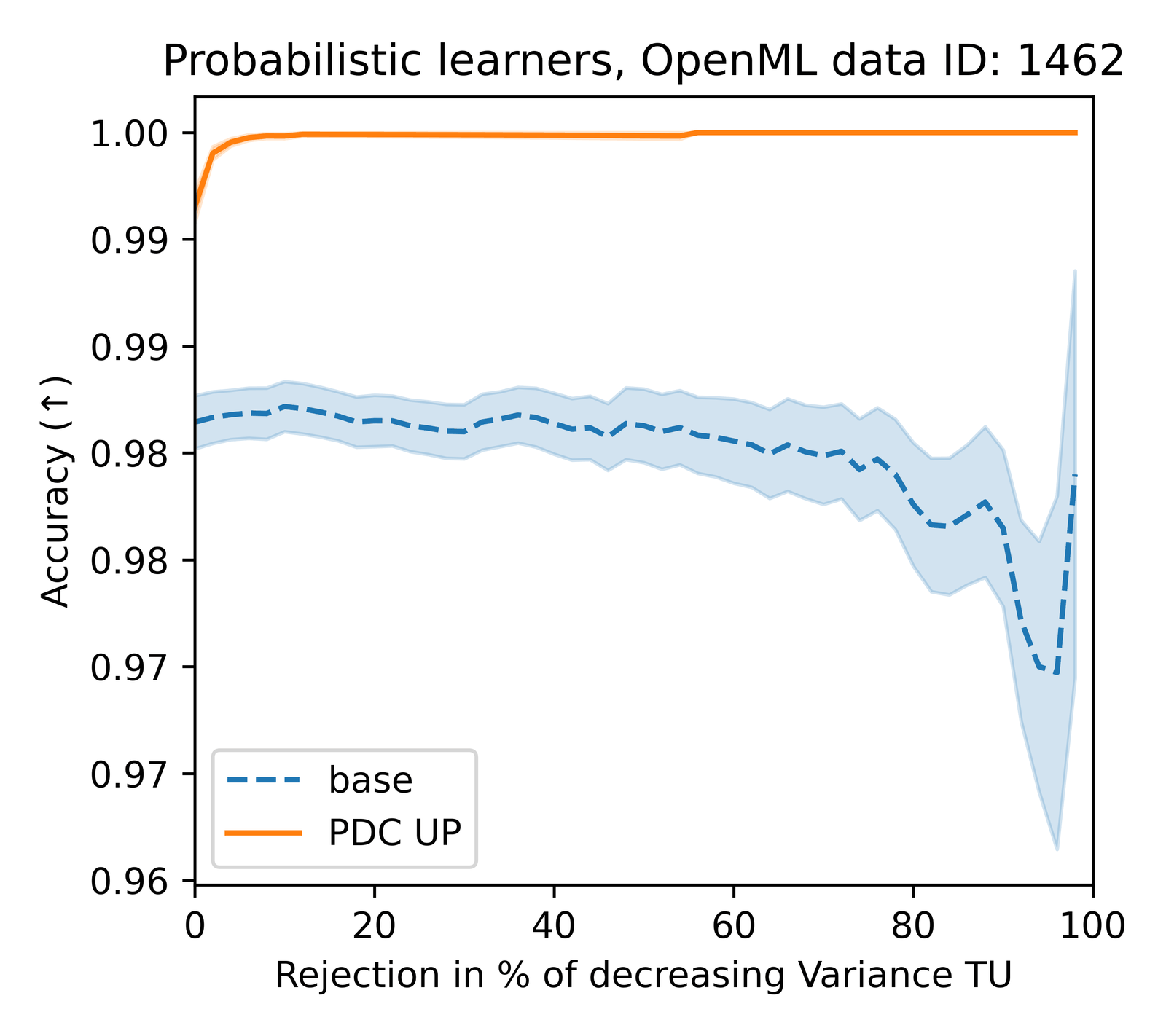}

    \includegraphics[width=0.4\linewidth]{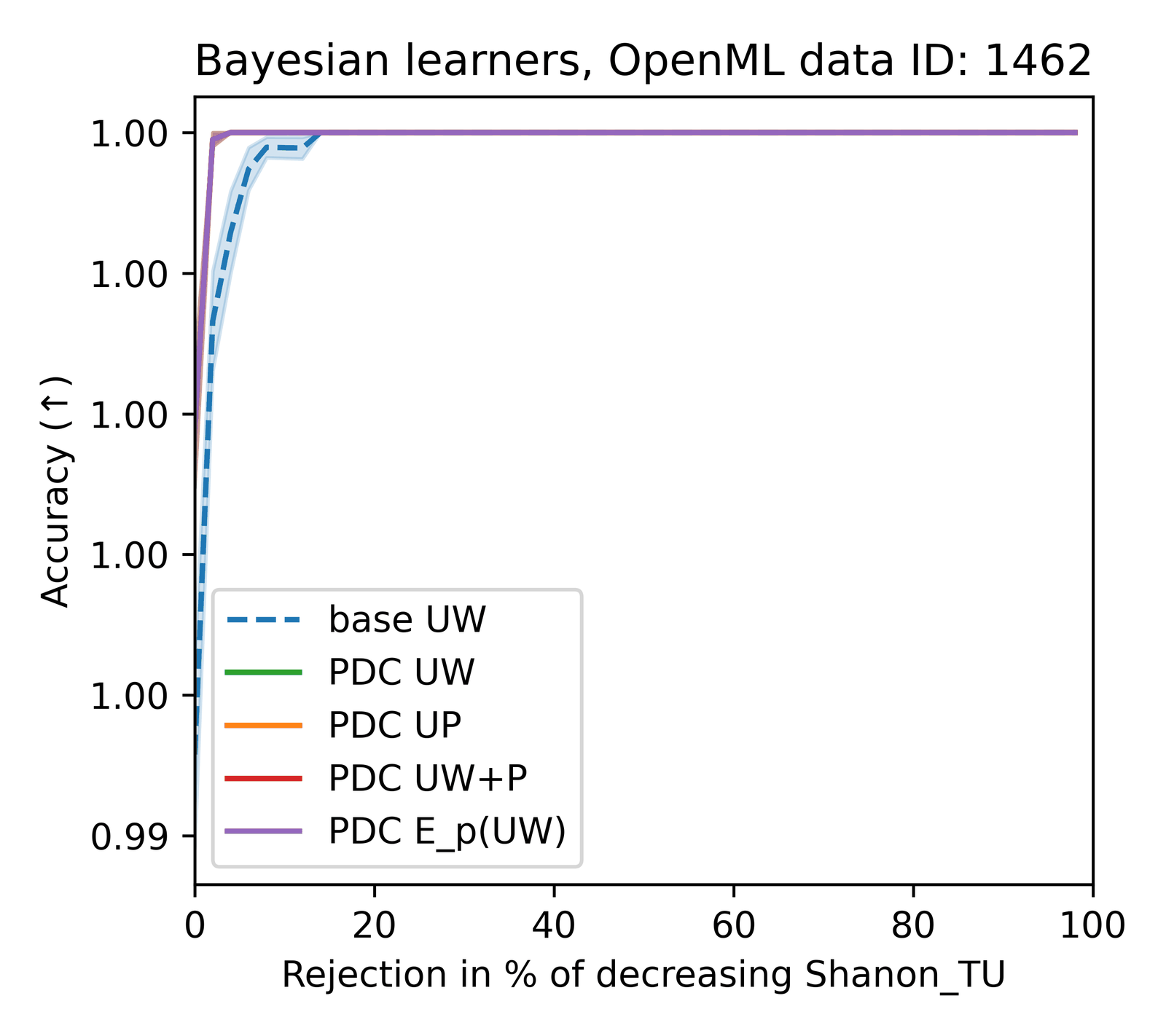}
    \includegraphics[width=0.4\linewidth]{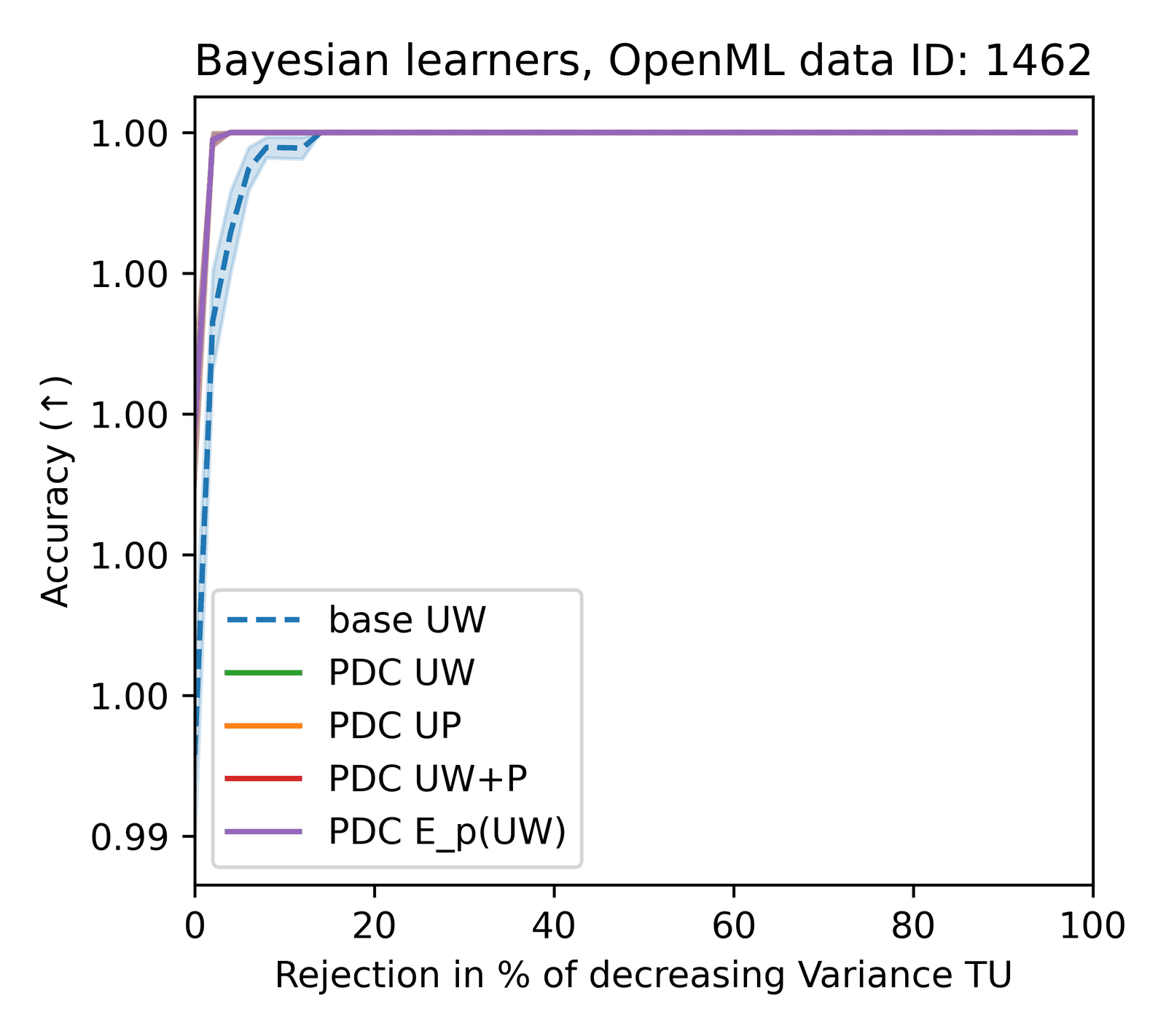}
    \caption{Accuracy rejection curve for the Dataset 1462.}
\end{figure}
\begin{figure}[h]
    \includegraphics[width=0.4\linewidth]{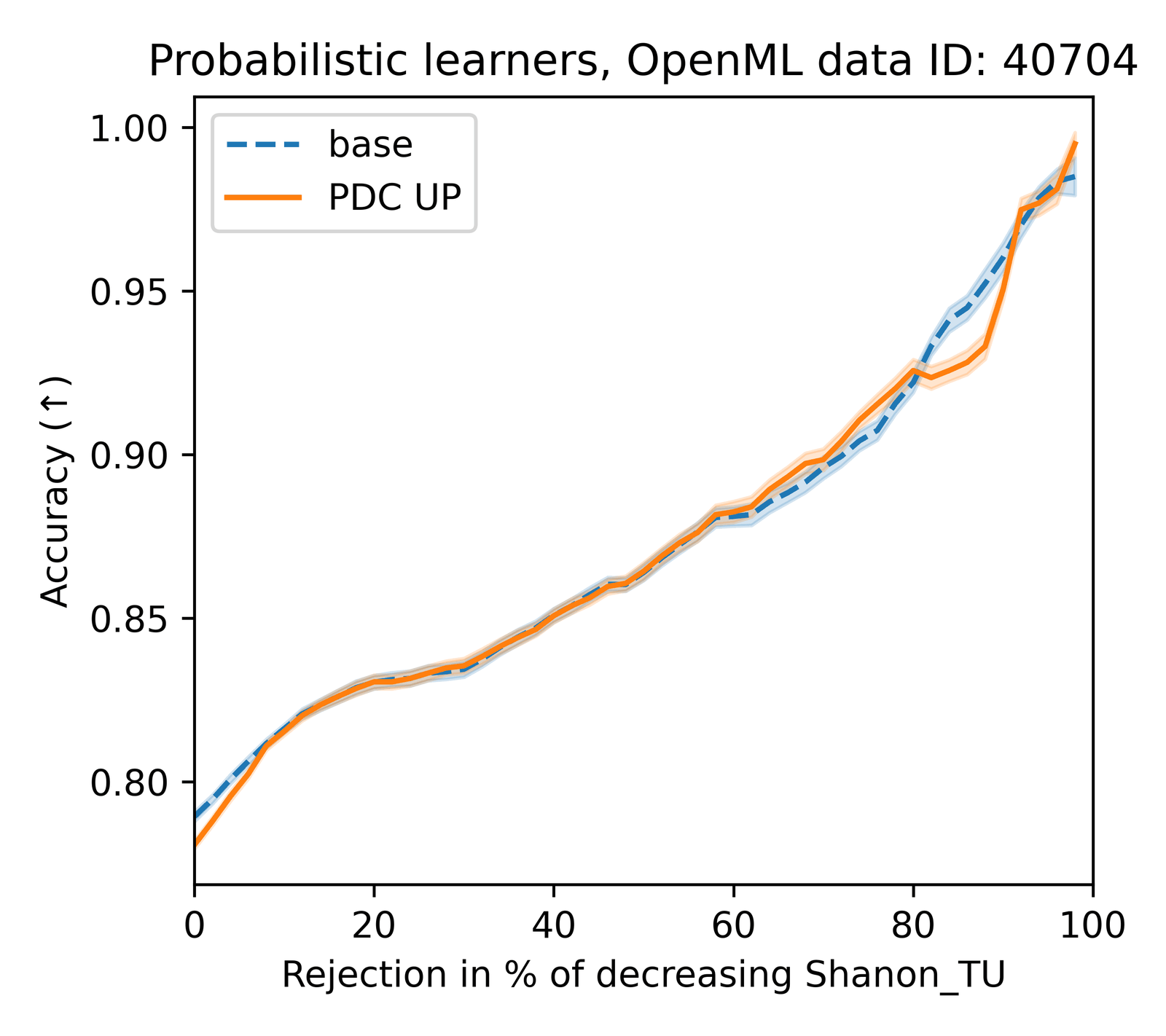}
    \includegraphics[width=0.4\linewidth]{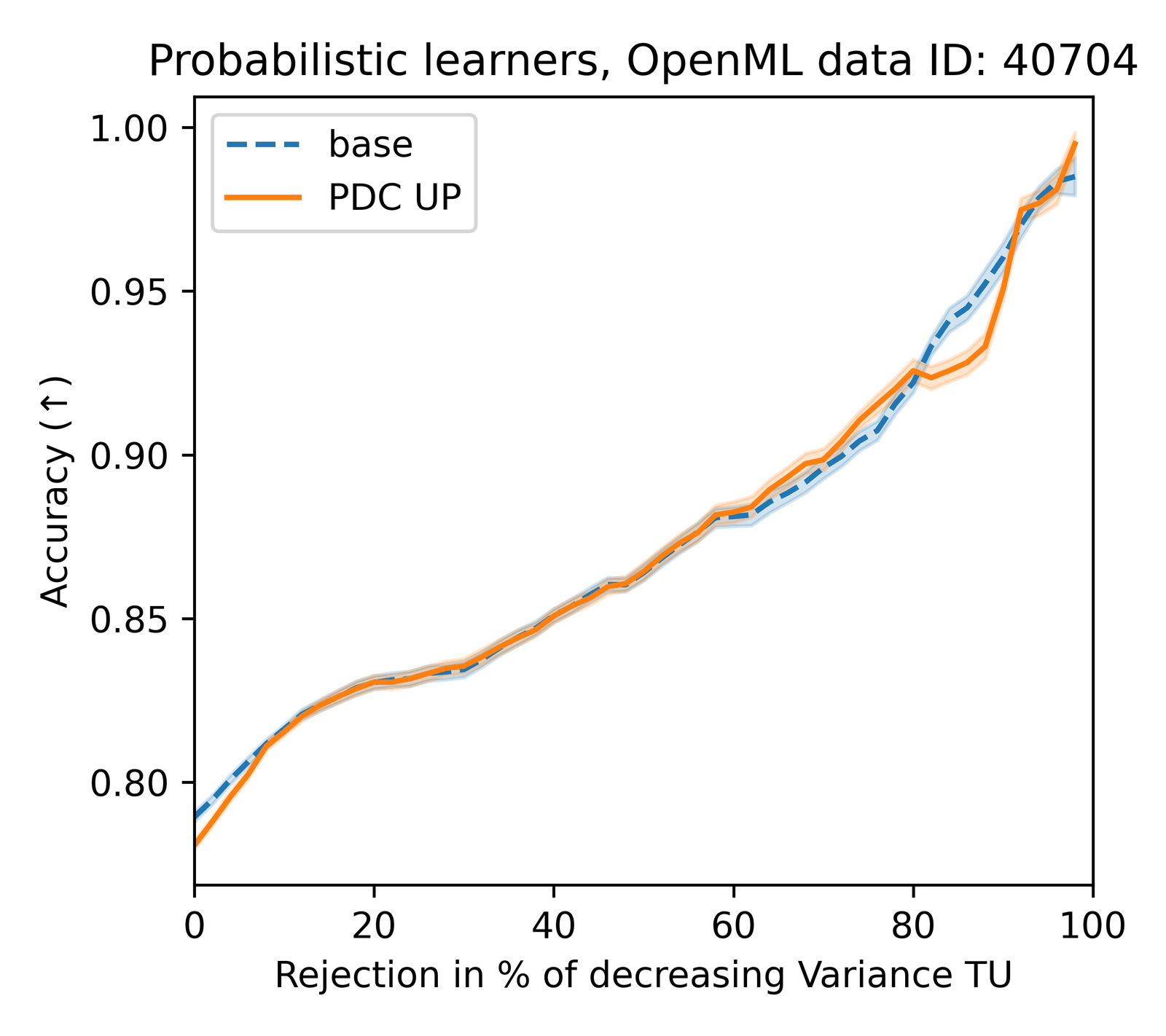}

    \includegraphics[width=0.4\linewidth]{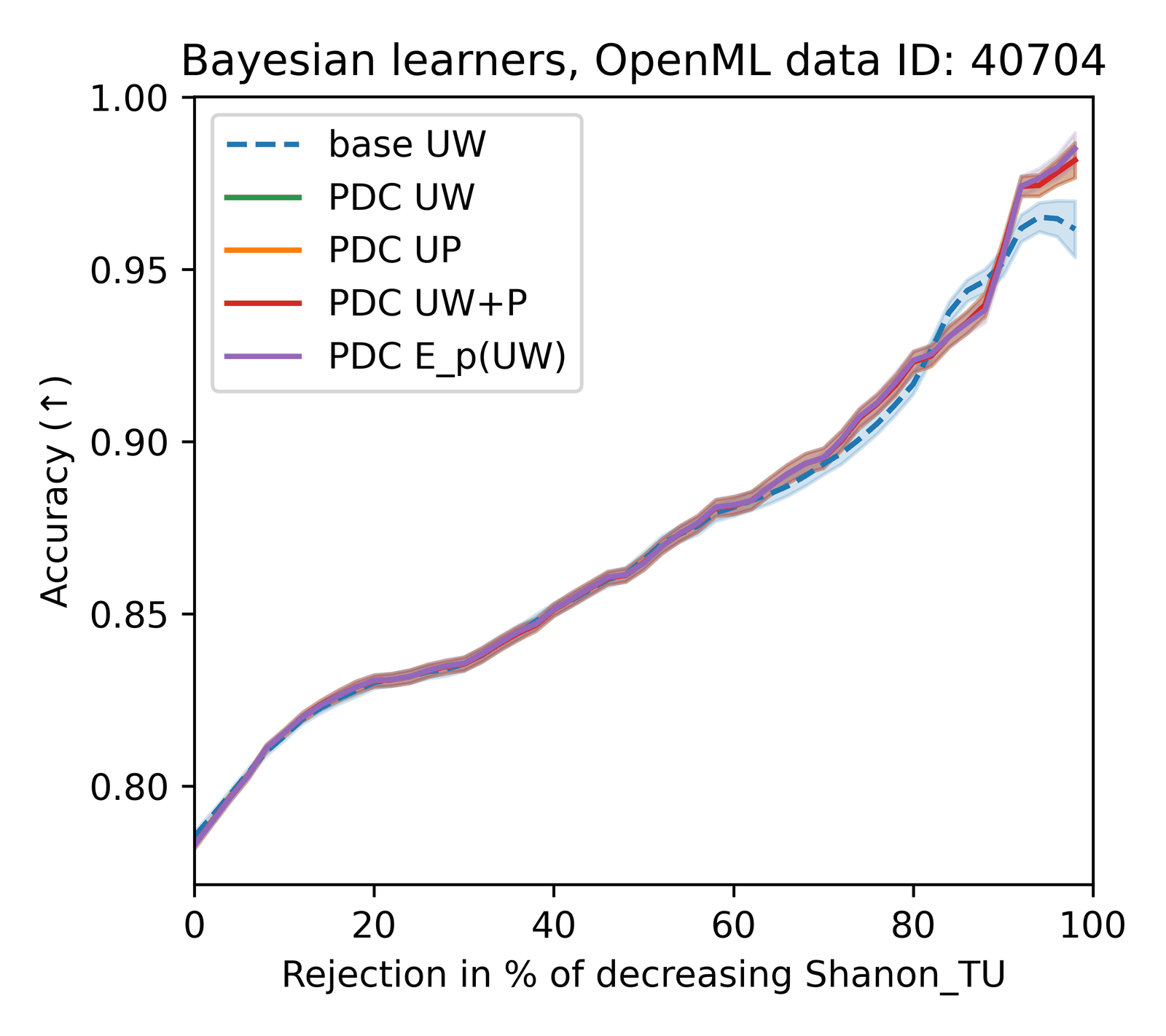}
    \includegraphics[width=0.4\linewidth]{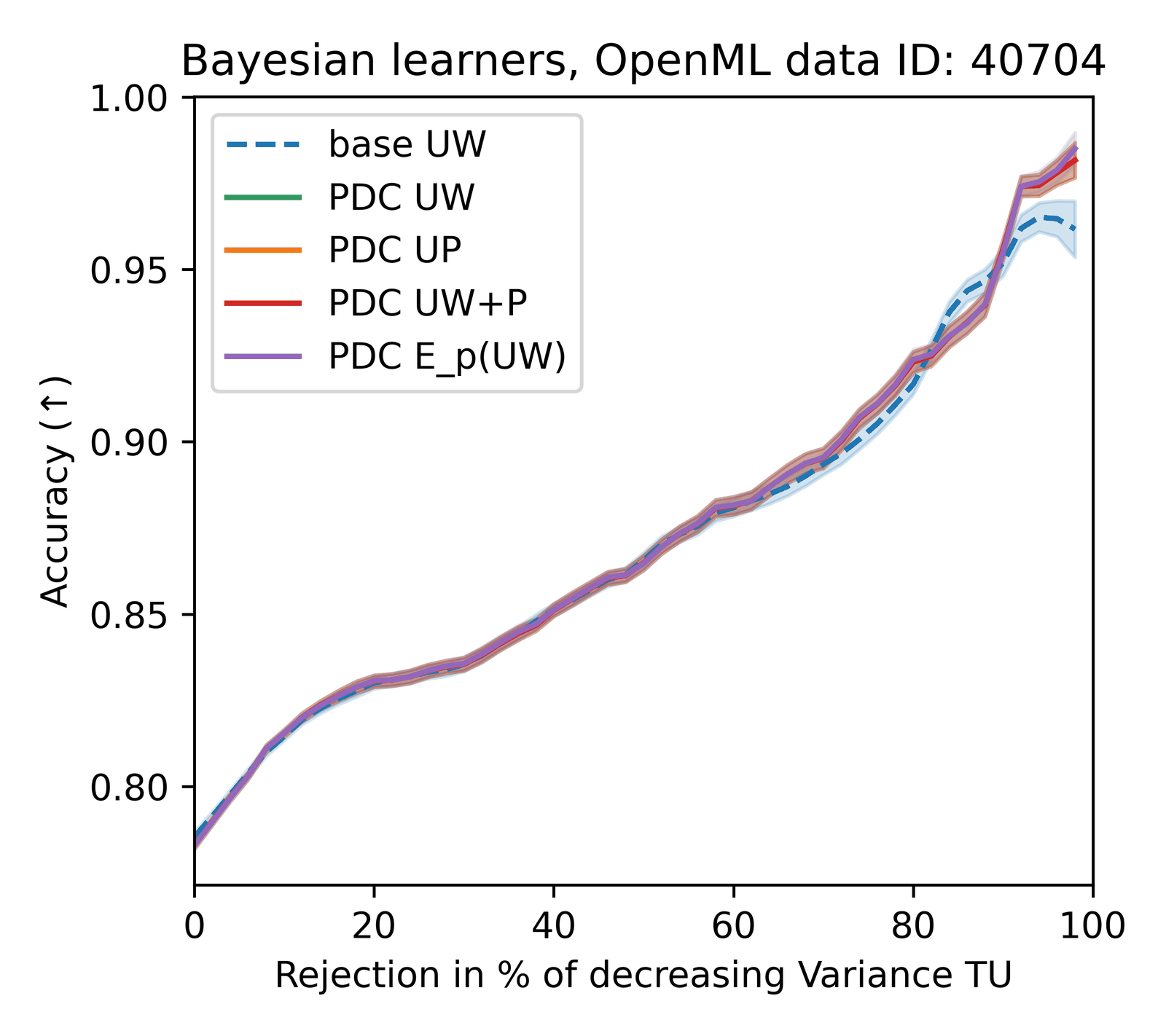}
    \caption{Accuracy rejection curve for the Dataset 40704.}
\end{figure}
\begin{figure}[h]
    \includegraphics[width=0.4\linewidth]{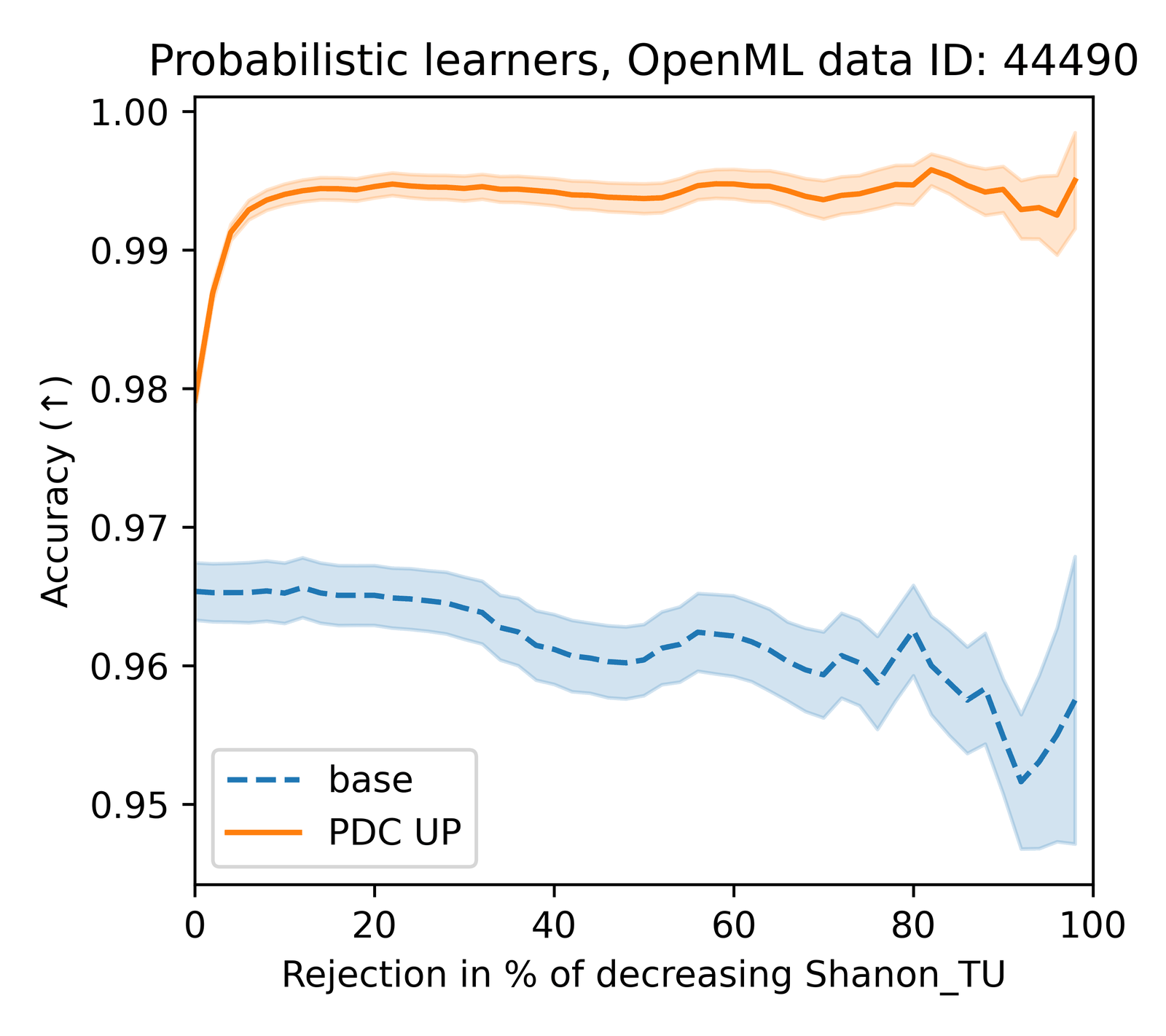}
    \includegraphics[width=0.4\linewidth]{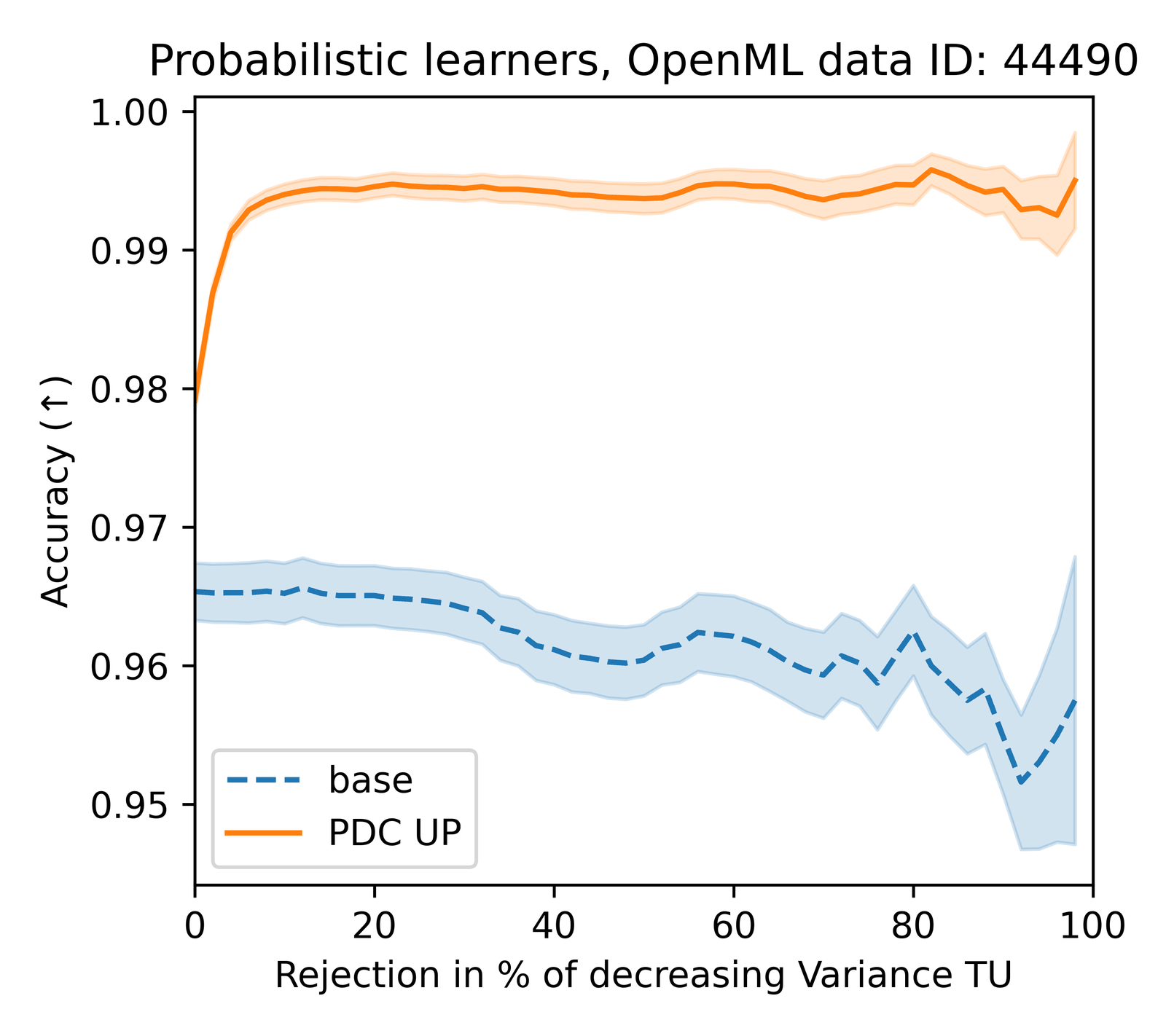}

    \includegraphics[width=0.4\linewidth]{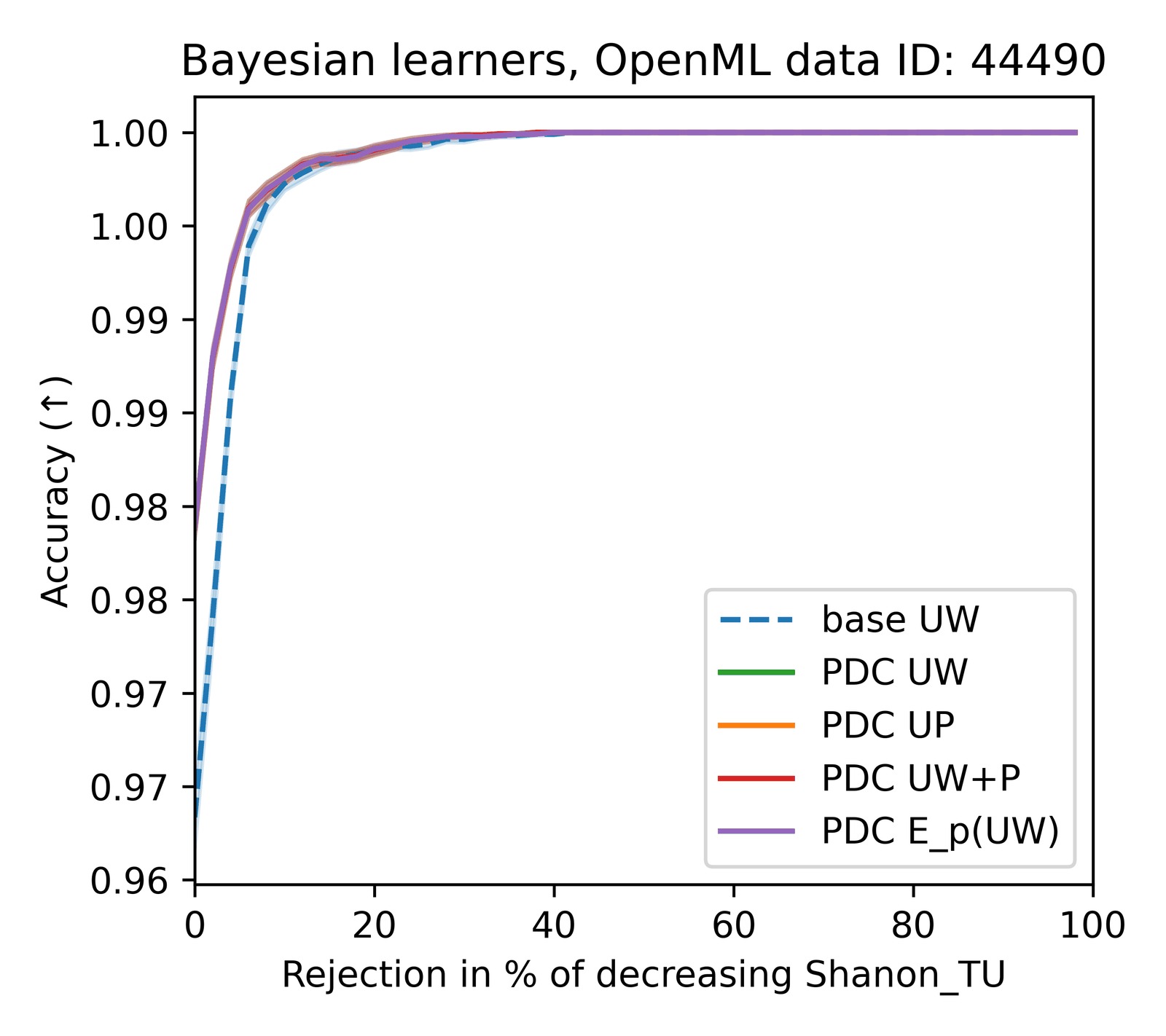}
    \includegraphics[width=0.4\linewidth]{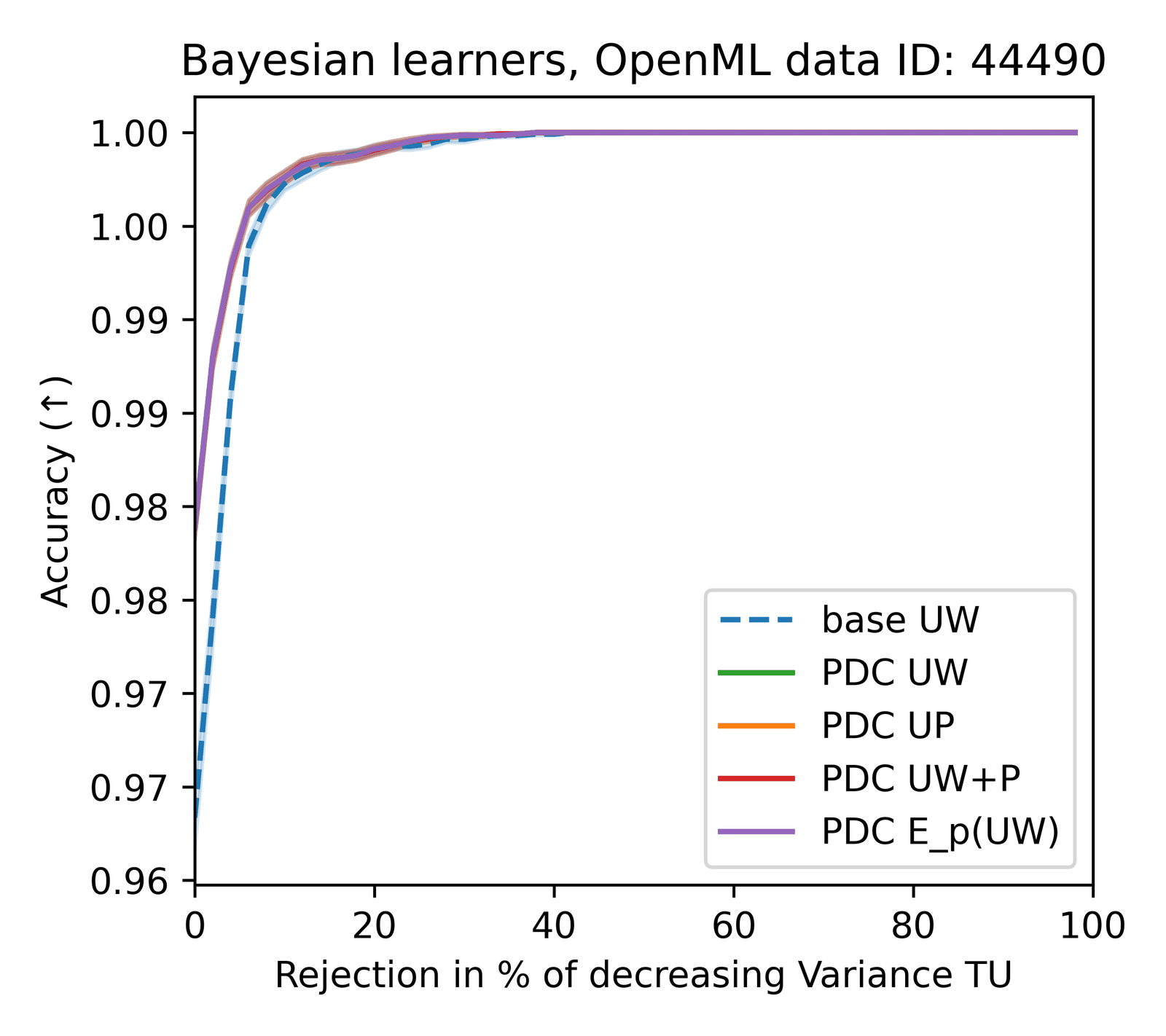}
    \caption{Accuracy rejection curve for the Dataset 44490.}
\end{figure}

\clearpage
\subsection{ECE-Rejection Curves}
A central aim of any EU measure is to capture the model's lack of knowledge. In an ideal scenario, test instances with higher EU should correspond to larger calibration errors. Since the true calibration error is unobservable in real-world settings, we approximate it by computing the Expected Calibration Error for each test point, denoted \(ECE_i\). Specifically, \(ECE_i\) is obtained by partitioning the test predictions into bins and measuring the absolute difference between the predicted probability and the average true label within each bin. By sorting test instances in descending order of EU, a well-calibrated EU measure should produce a quasi-monotonic decrease in \(ECE_i\), thereby confirming that higher uncertainty is indeed associated with poorer calibration.
We plot the average ECE for the subset of test points remaining after rejection, as a function of the rejection rate.

\begin{figure}[h]
    \includegraphics[width=0.4\linewidth]{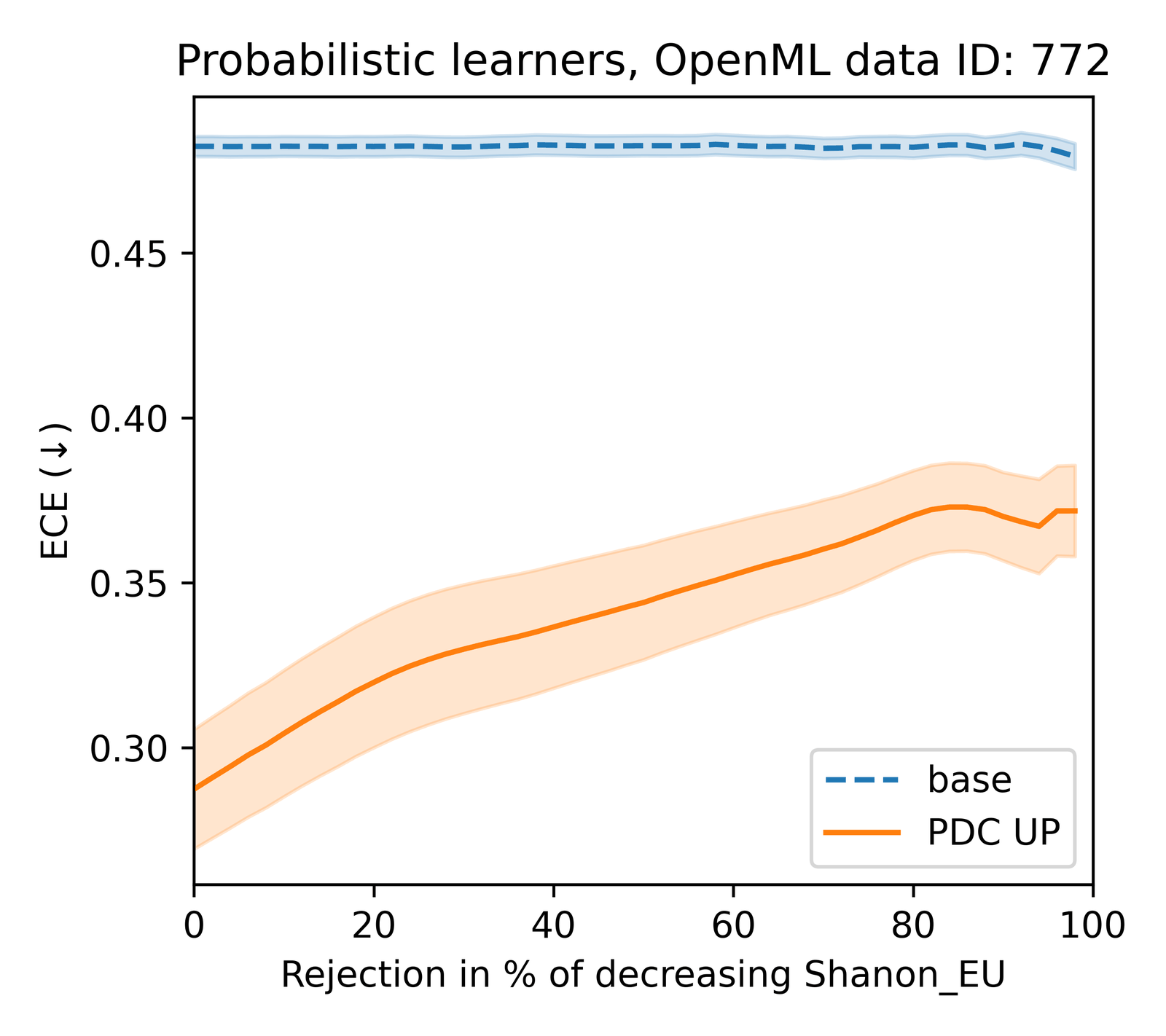}
    \includegraphics[width=0.4\linewidth]{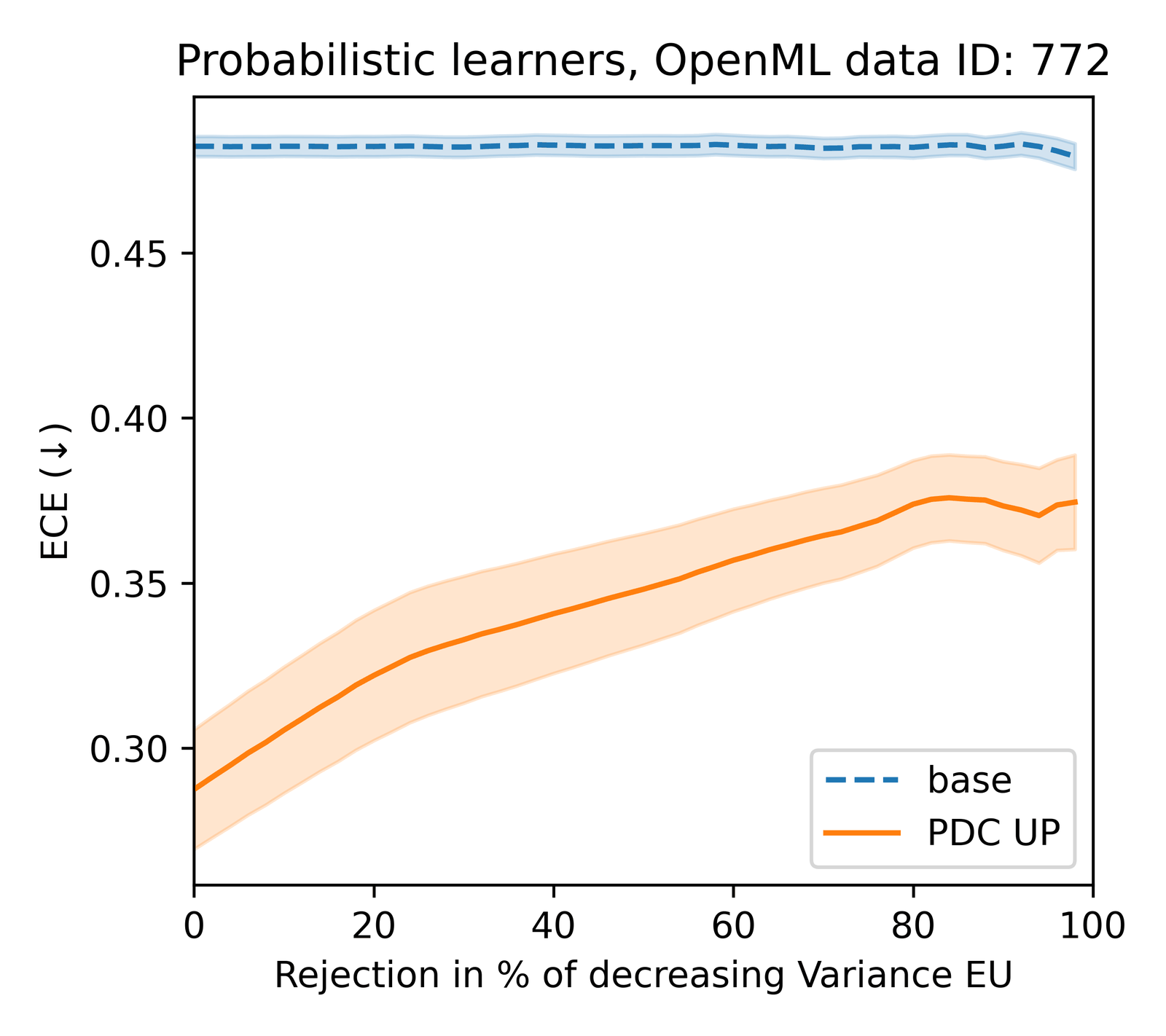}

    \includegraphics[width=0.4\linewidth]{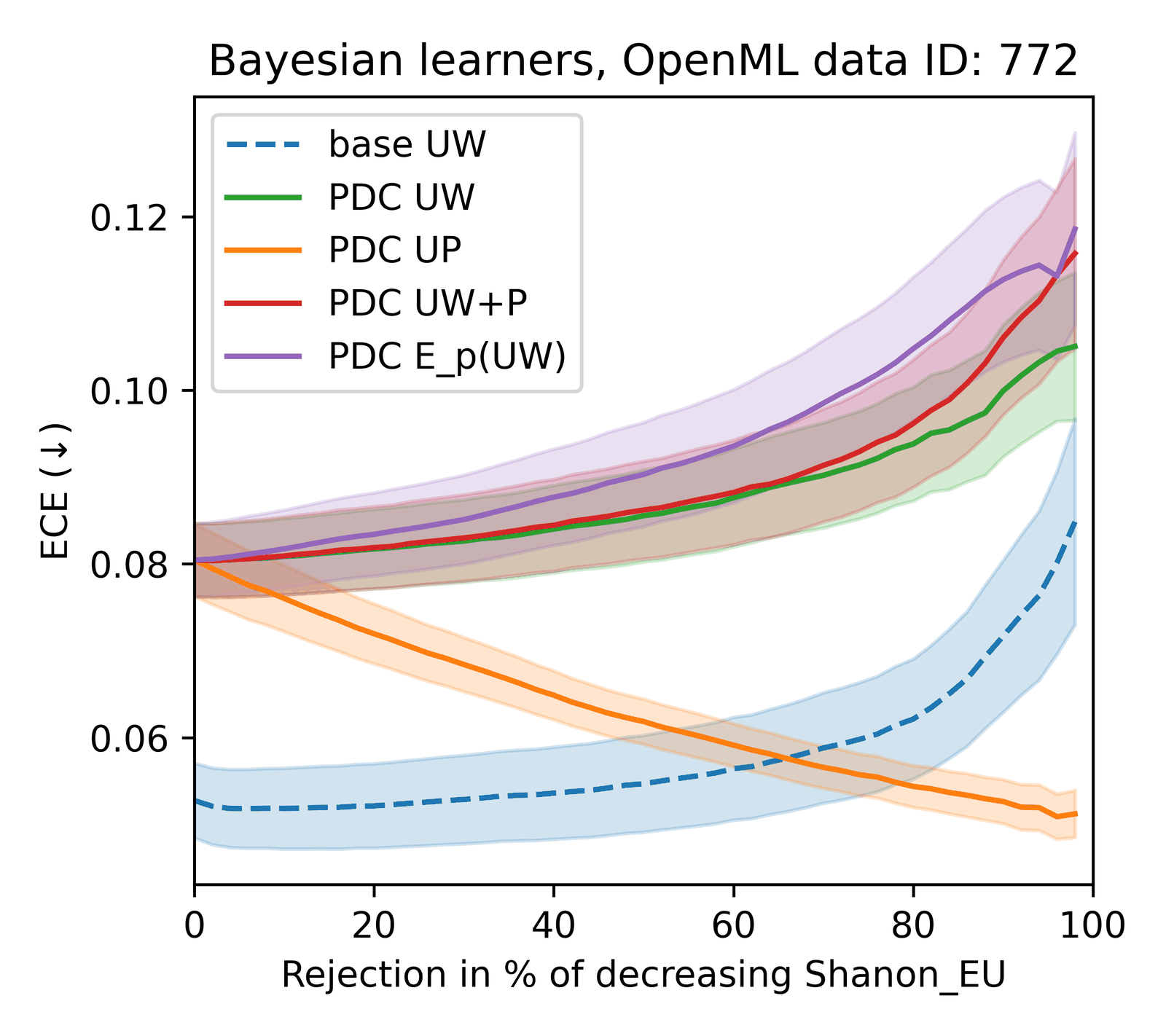}
    \includegraphics[width=0.4\linewidth]{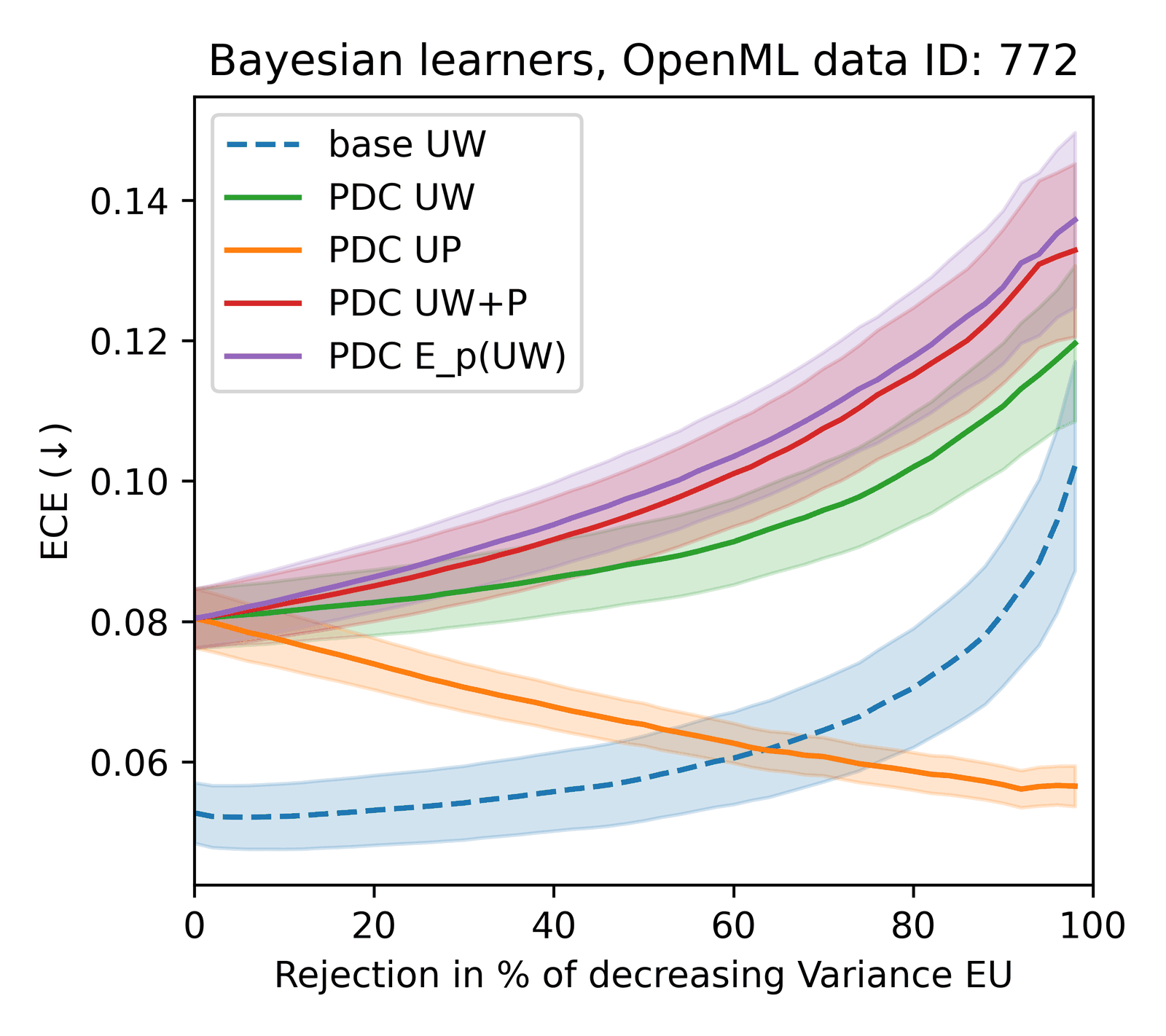}
    \caption{ECE rejection curve for the Dataset 772.}
\end{figure}
\begin{figure}[h]
    \includegraphics[width=0.4\linewidth]{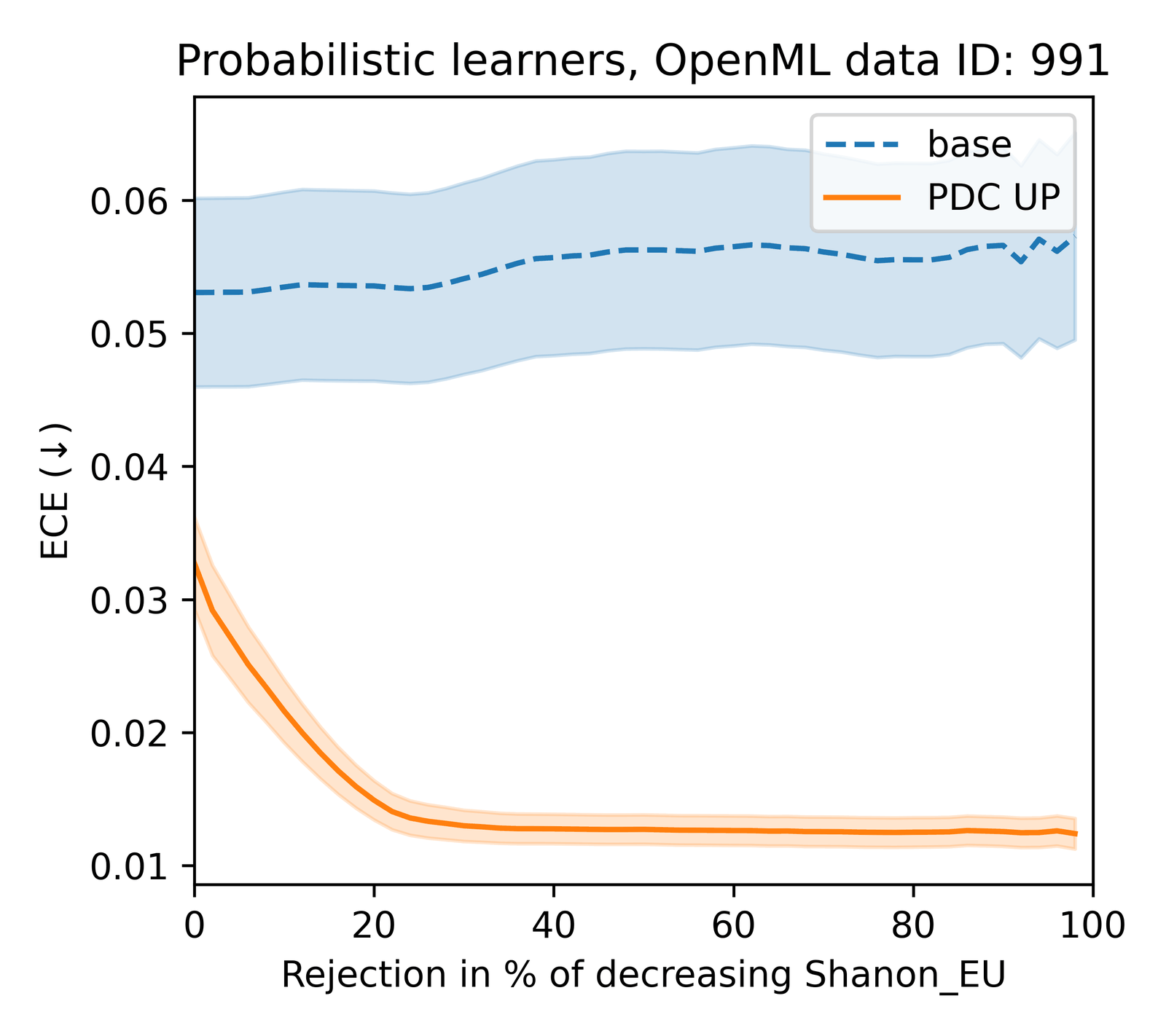}
    \includegraphics[width=0.4\linewidth]{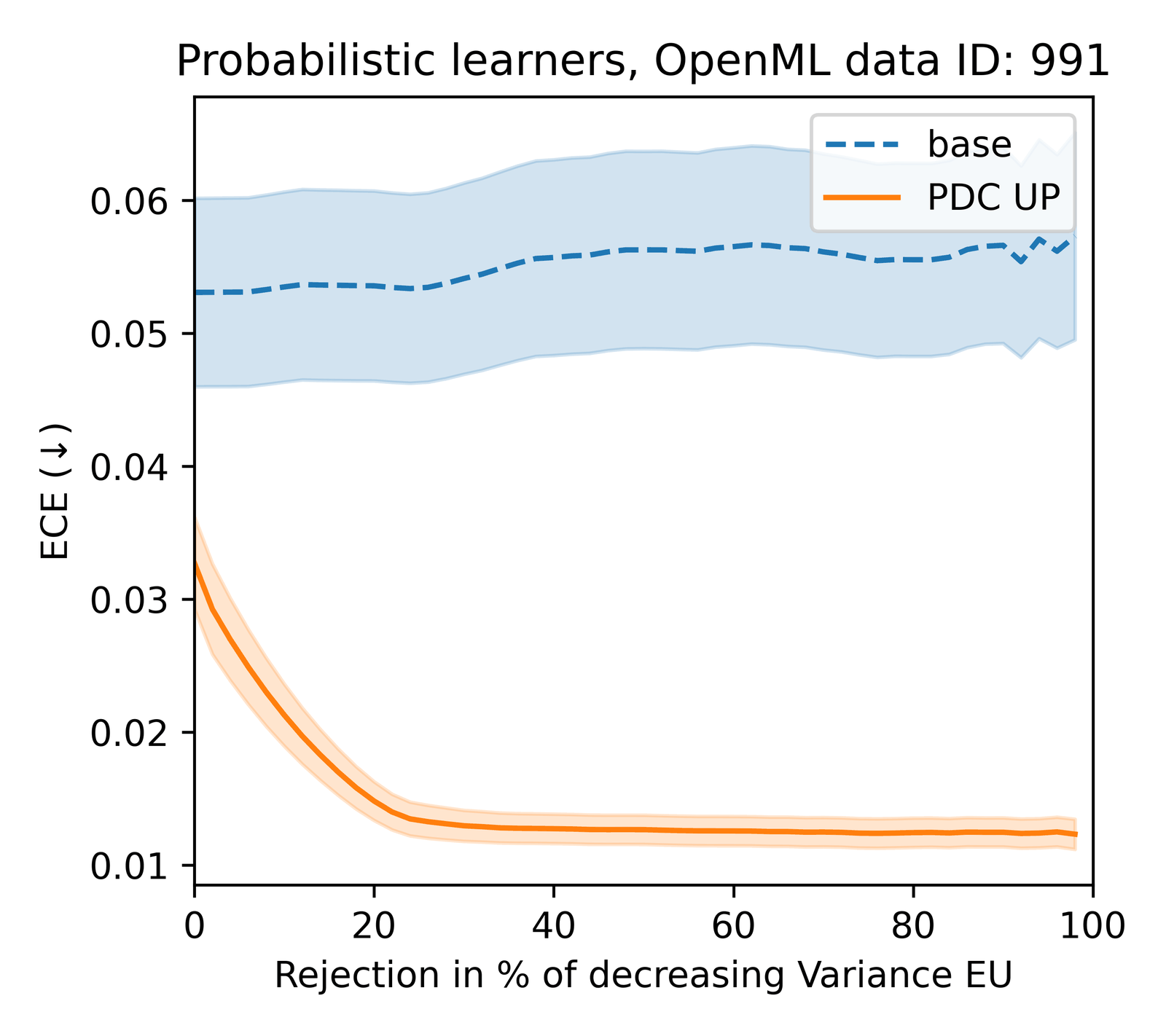}

    \includegraphics[width=0.4\linewidth]{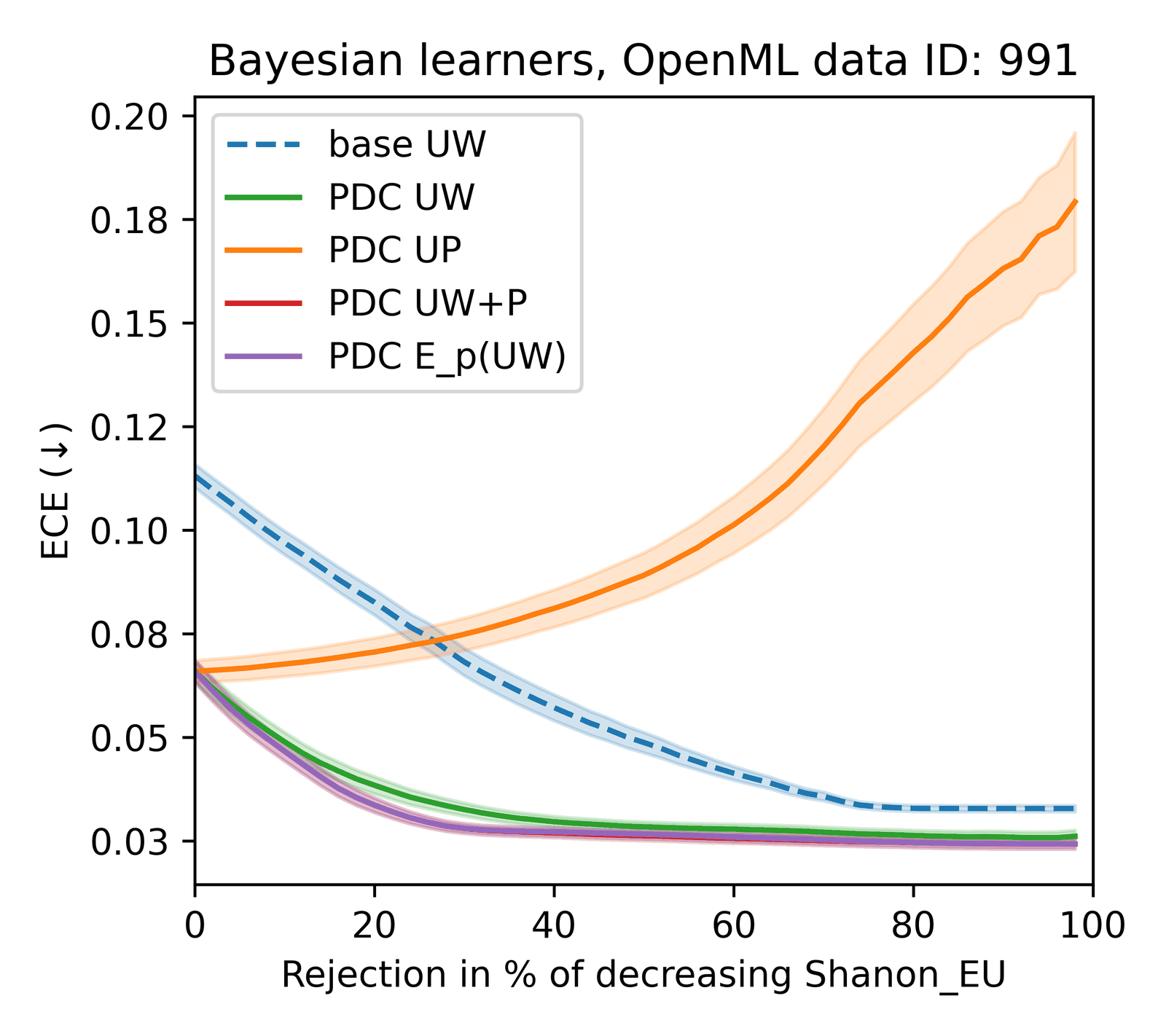}
    \includegraphics[width=0.4\linewidth]{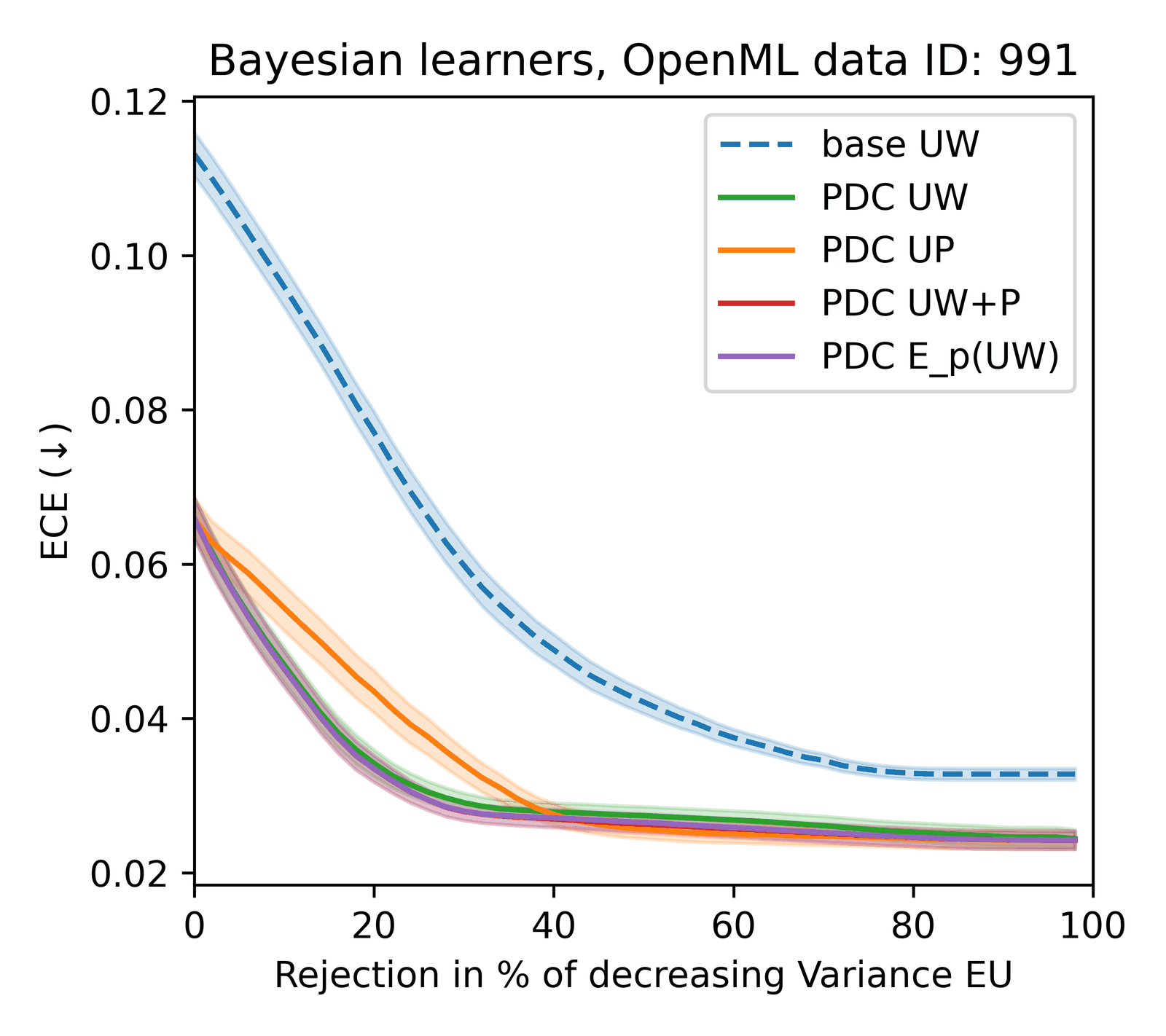}
    \caption{ECE rejection curve for the Dataset 991.}
\end{figure}
\begin{figure}[h]
    \includegraphics[width=0.4\linewidth]{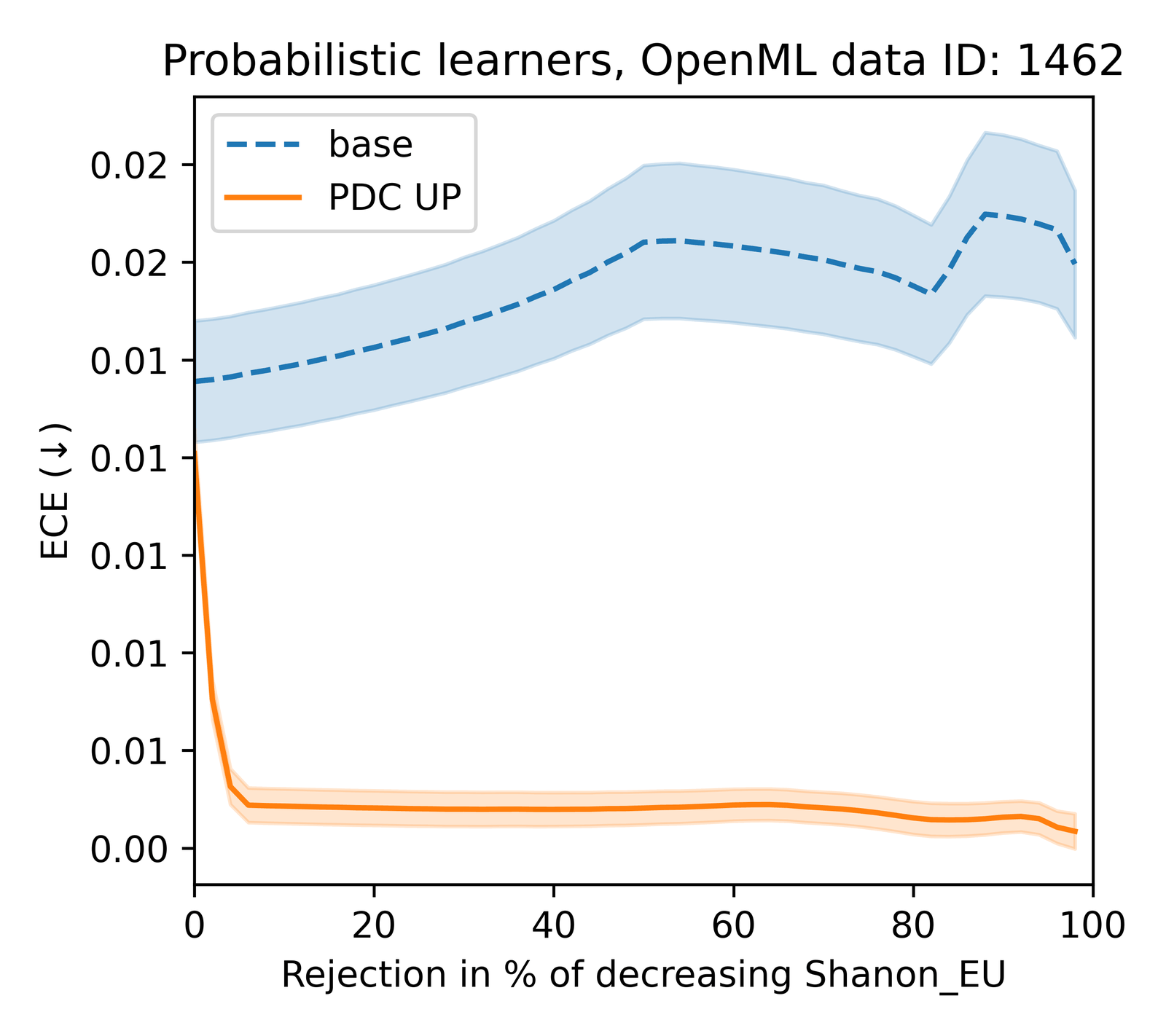}
    \includegraphics[width=0.4\linewidth]{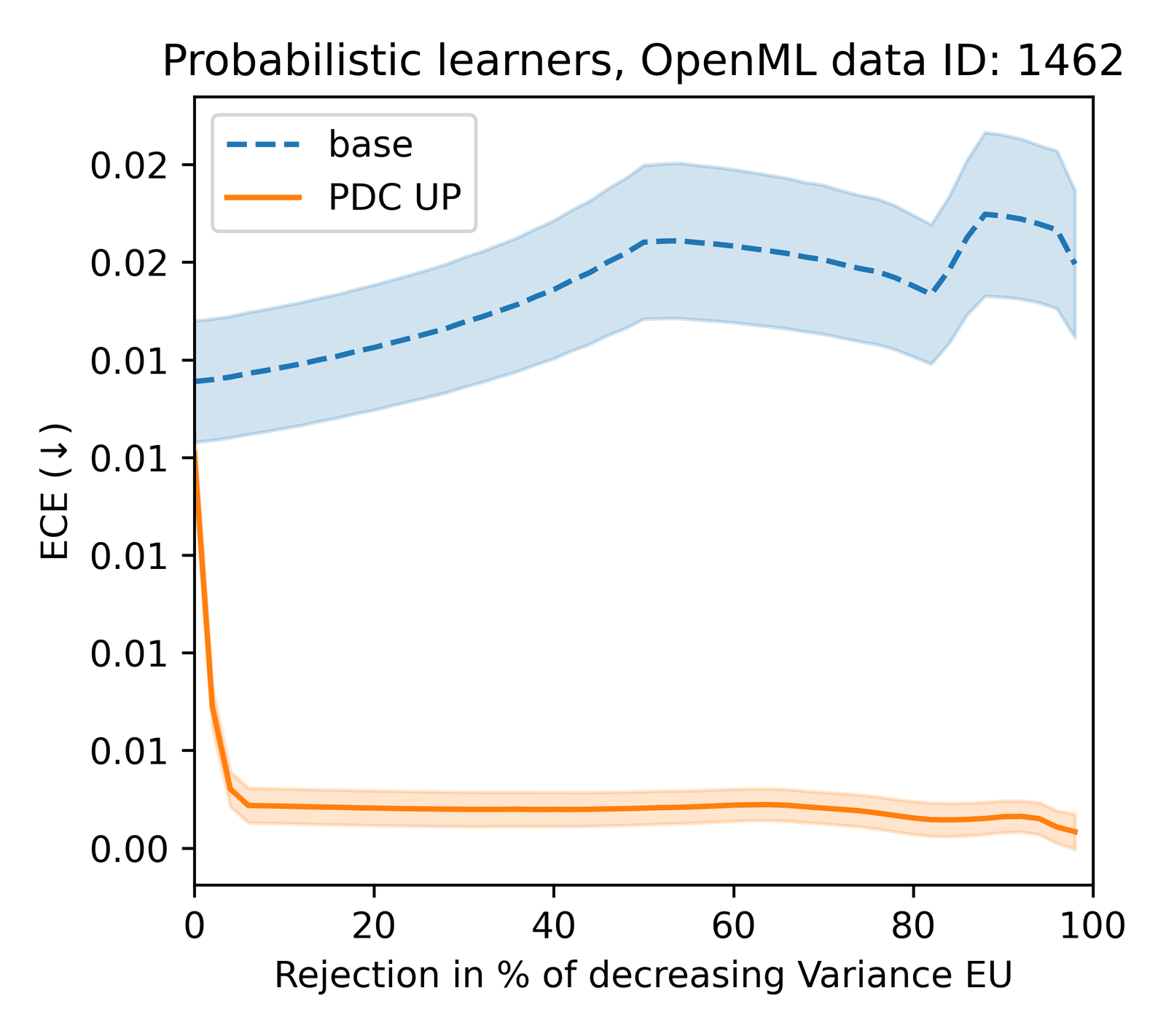}

    \includegraphics[width=0.4\linewidth]{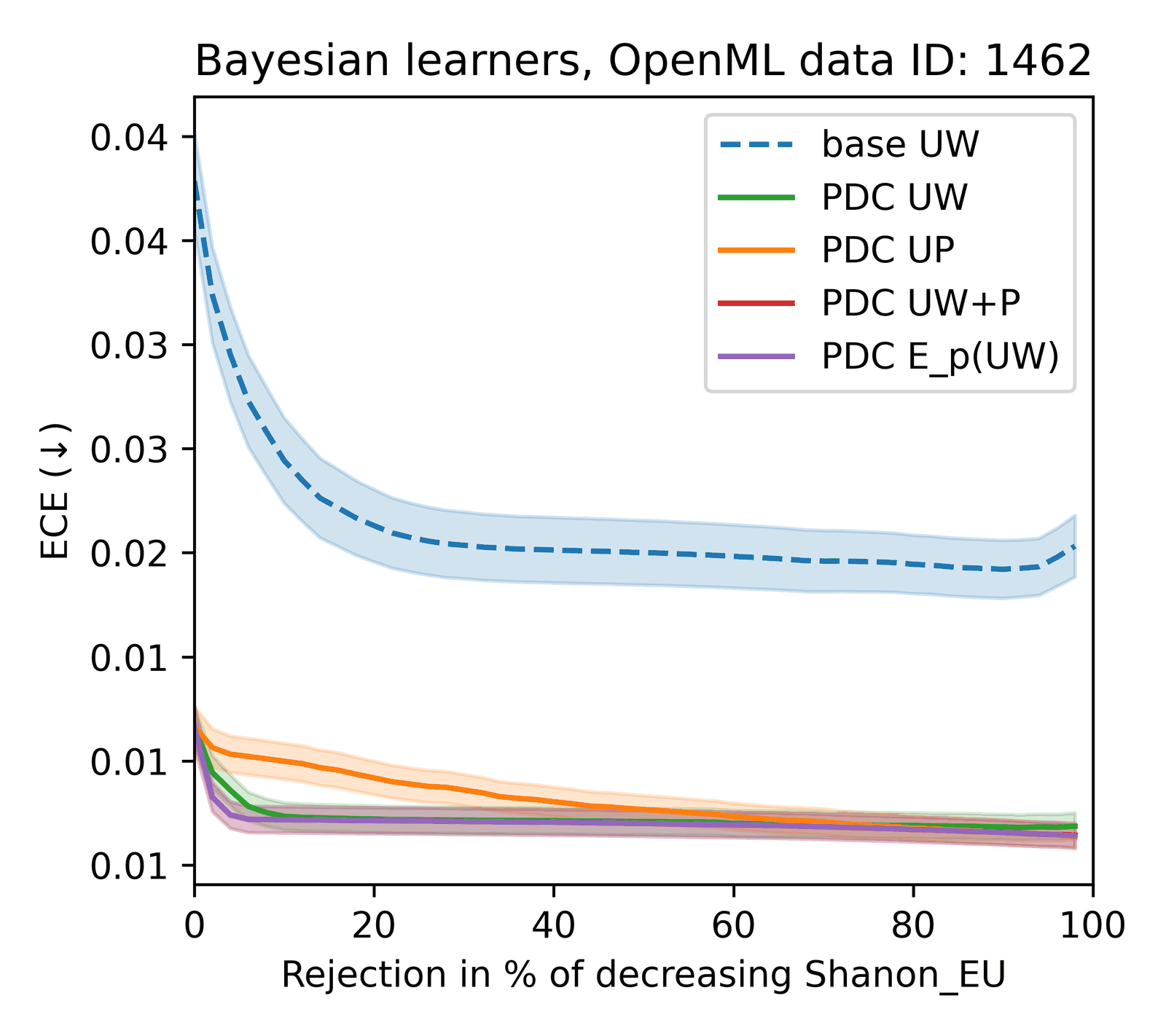}
    \includegraphics[width=0.4\linewidth]{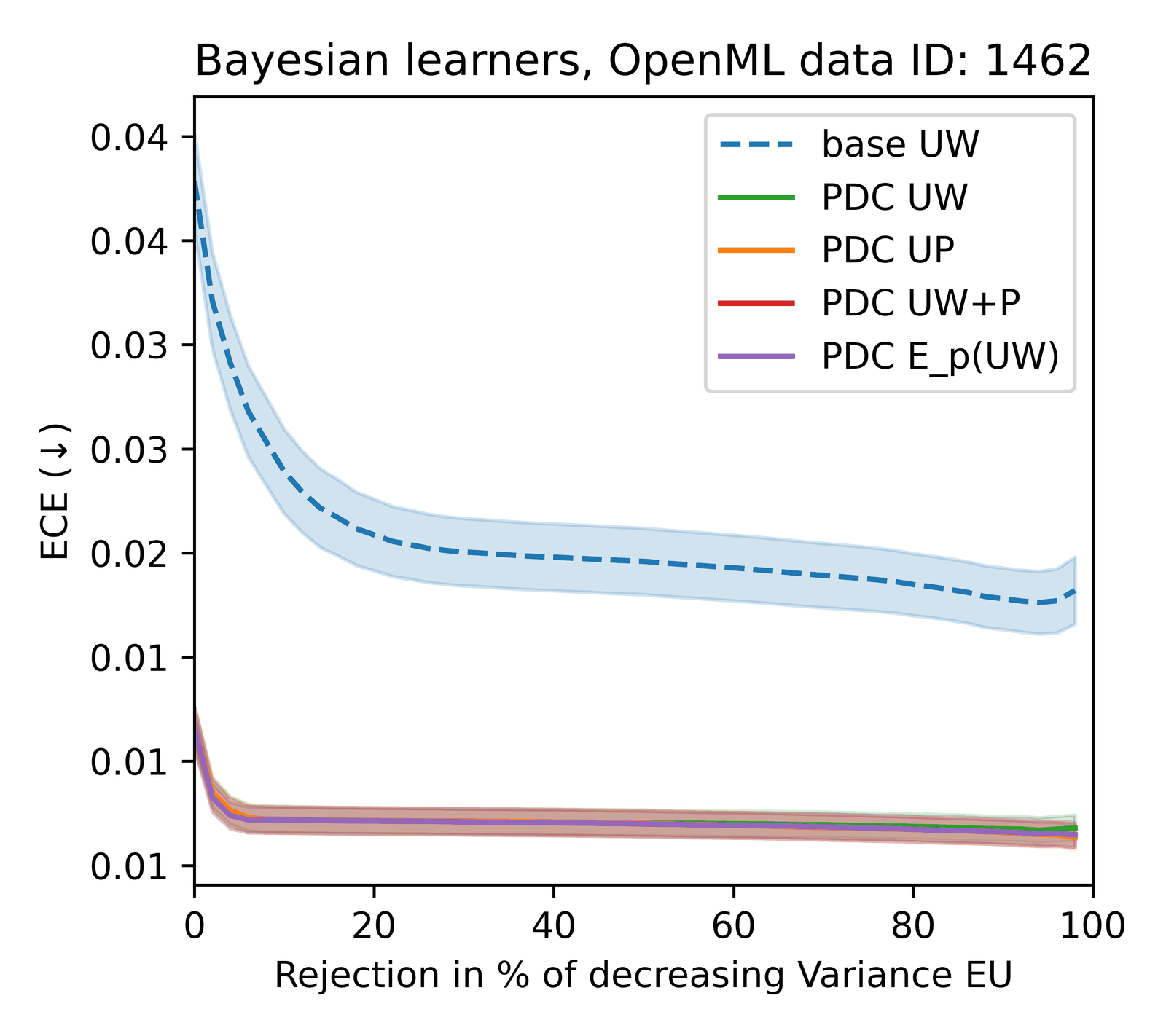}
    \caption{ECE rejection curve for the Dataset 1462.}
\end{figure}
\begin{figure}[h]
    \includegraphics[width=0.4\linewidth]{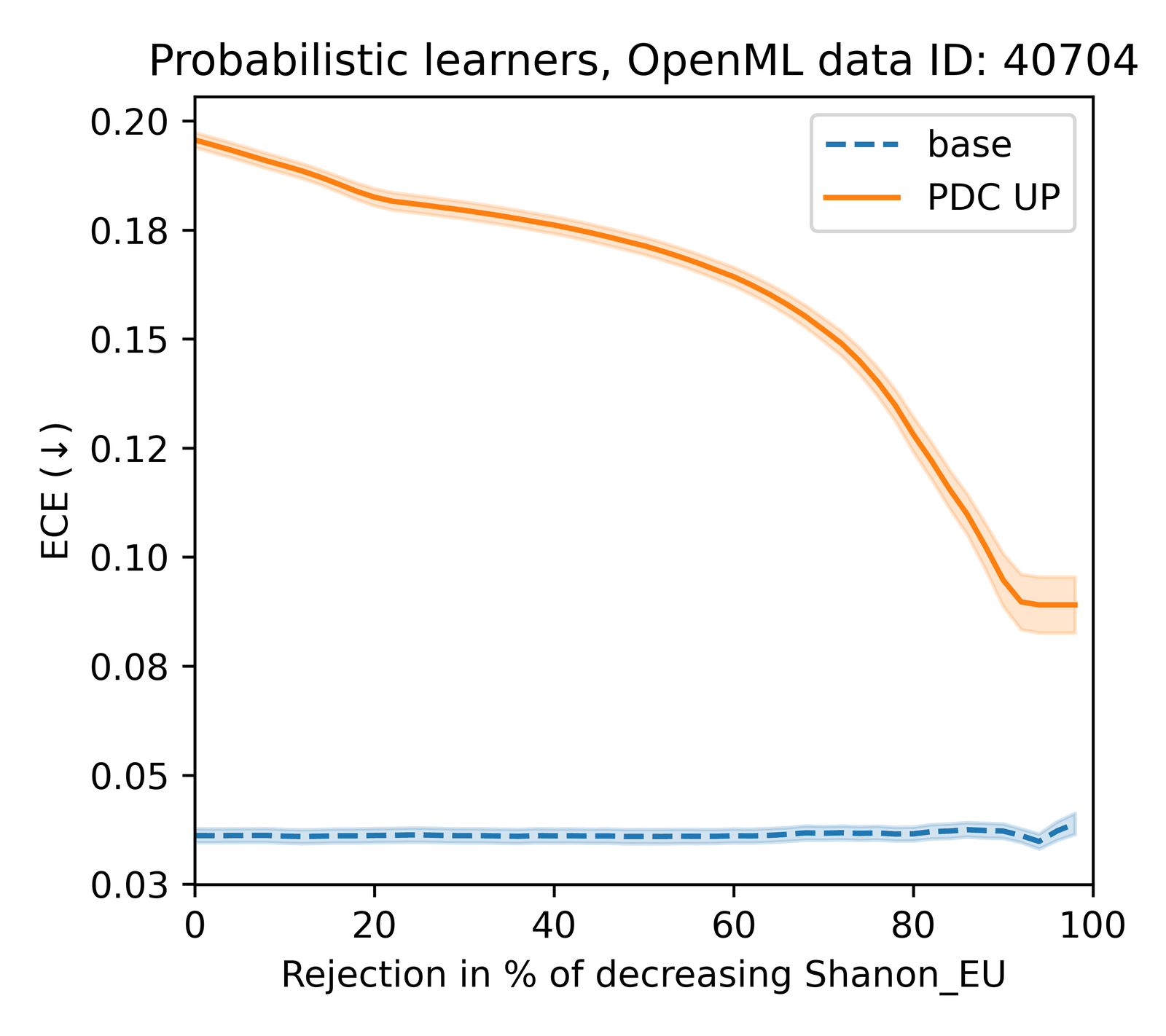}
    \includegraphics[width=0.4\linewidth]{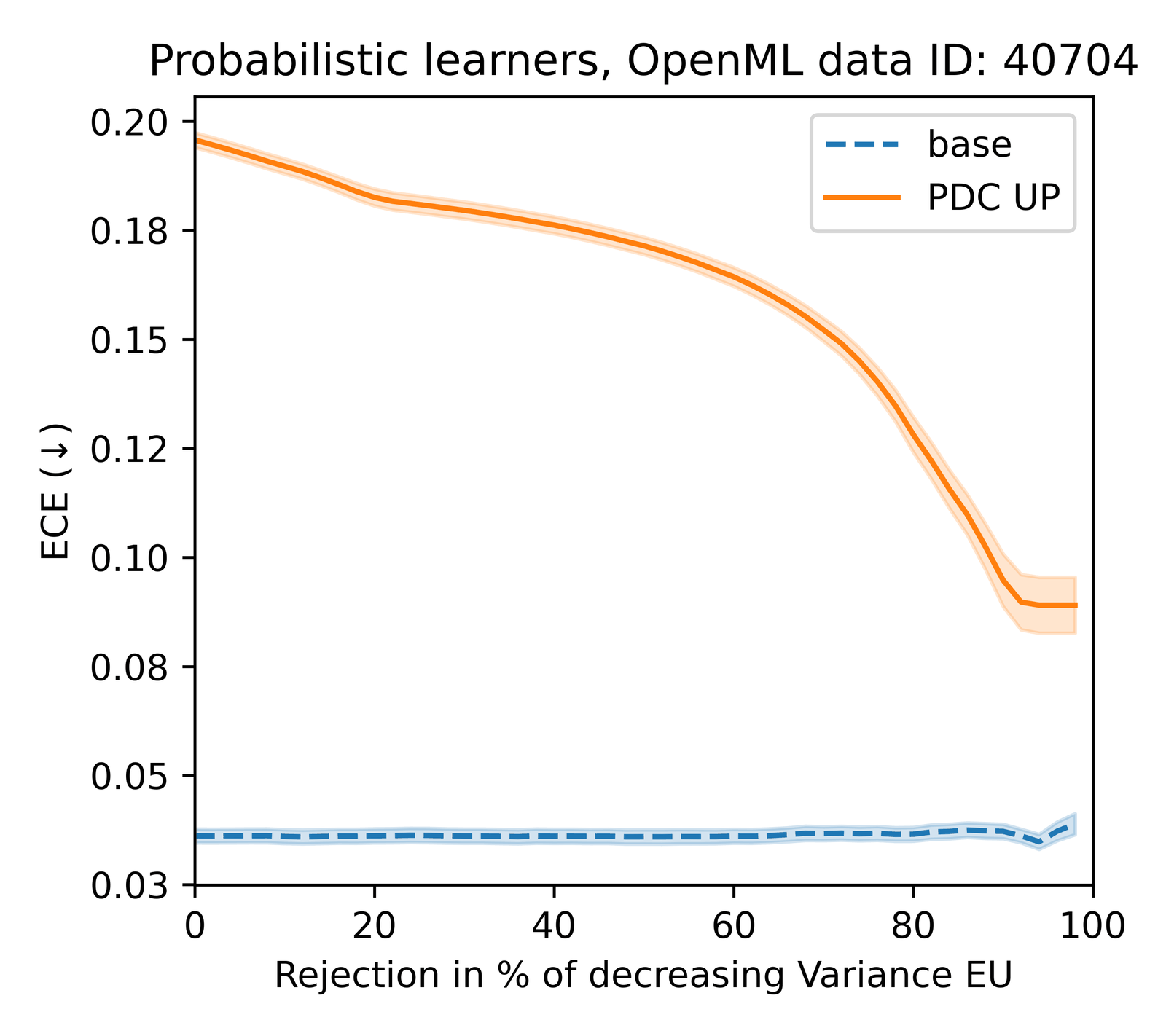}

    \includegraphics[width=0.4\linewidth]{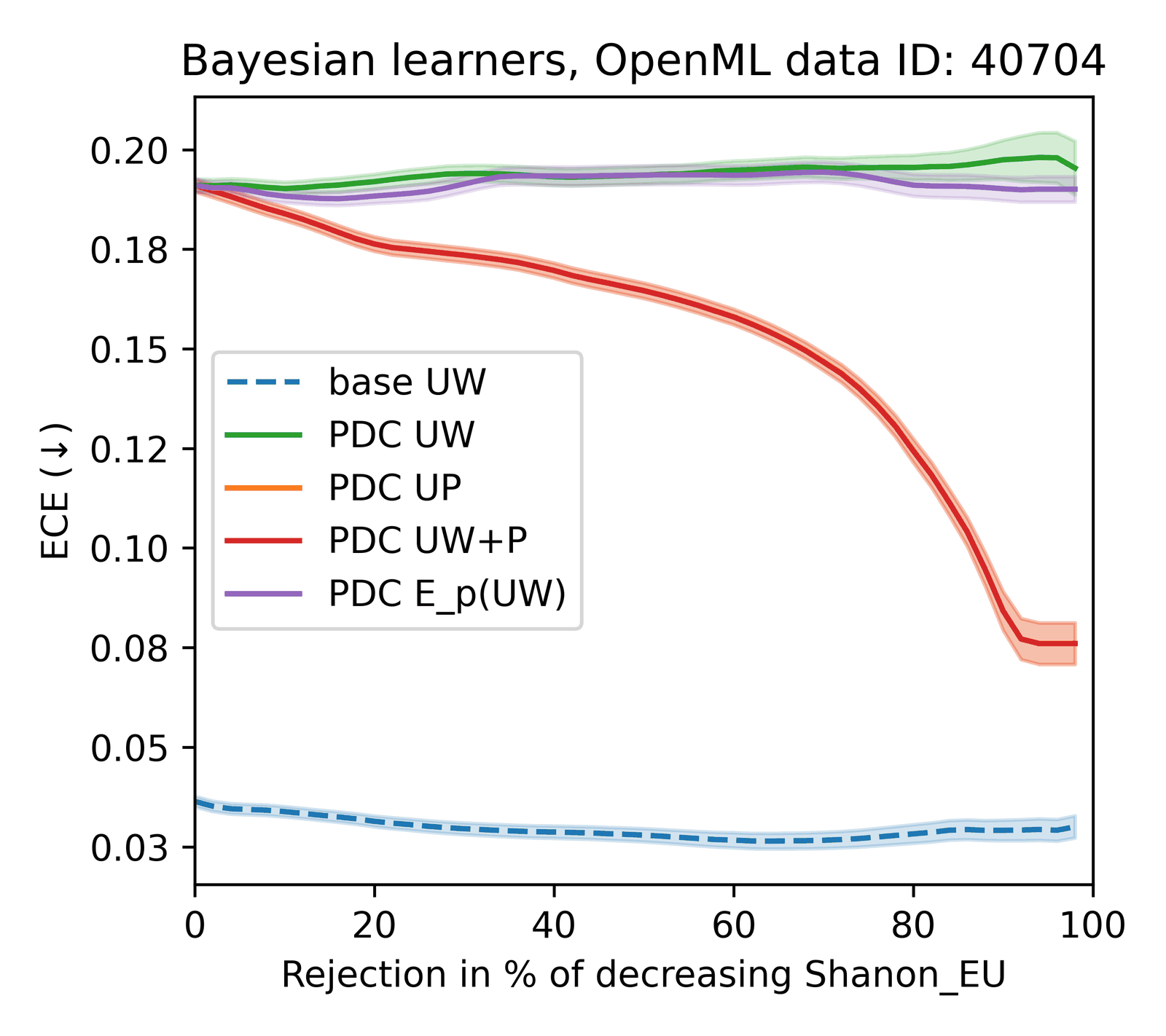}
    \includegraphics[width=0.4\linewidth]{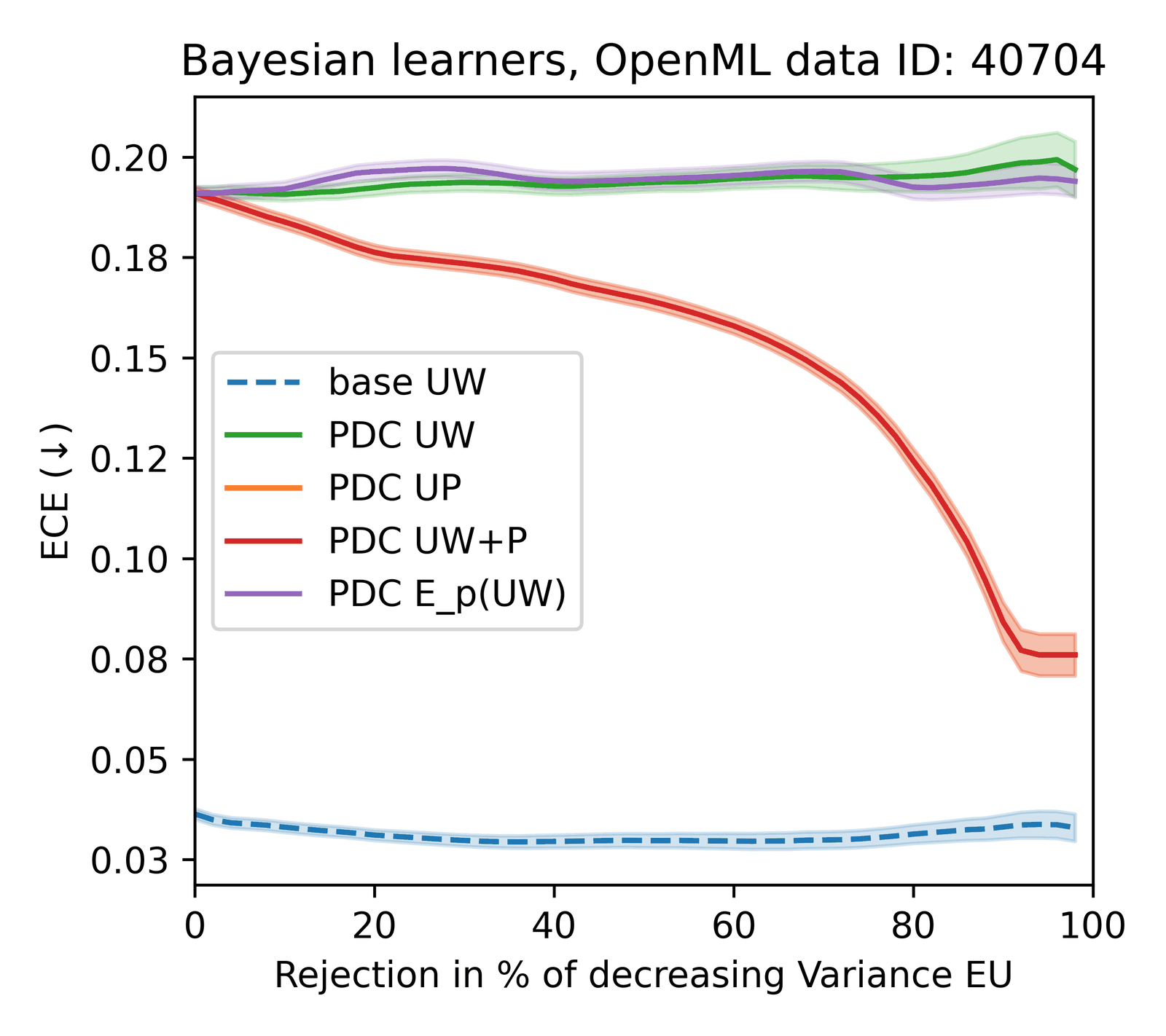}
    \caption{ECE rejection curve for the Dataset 40704.}
\end{figure}
\begin{figure}[h]
    \includegraphics[width=0.4\linewidth]{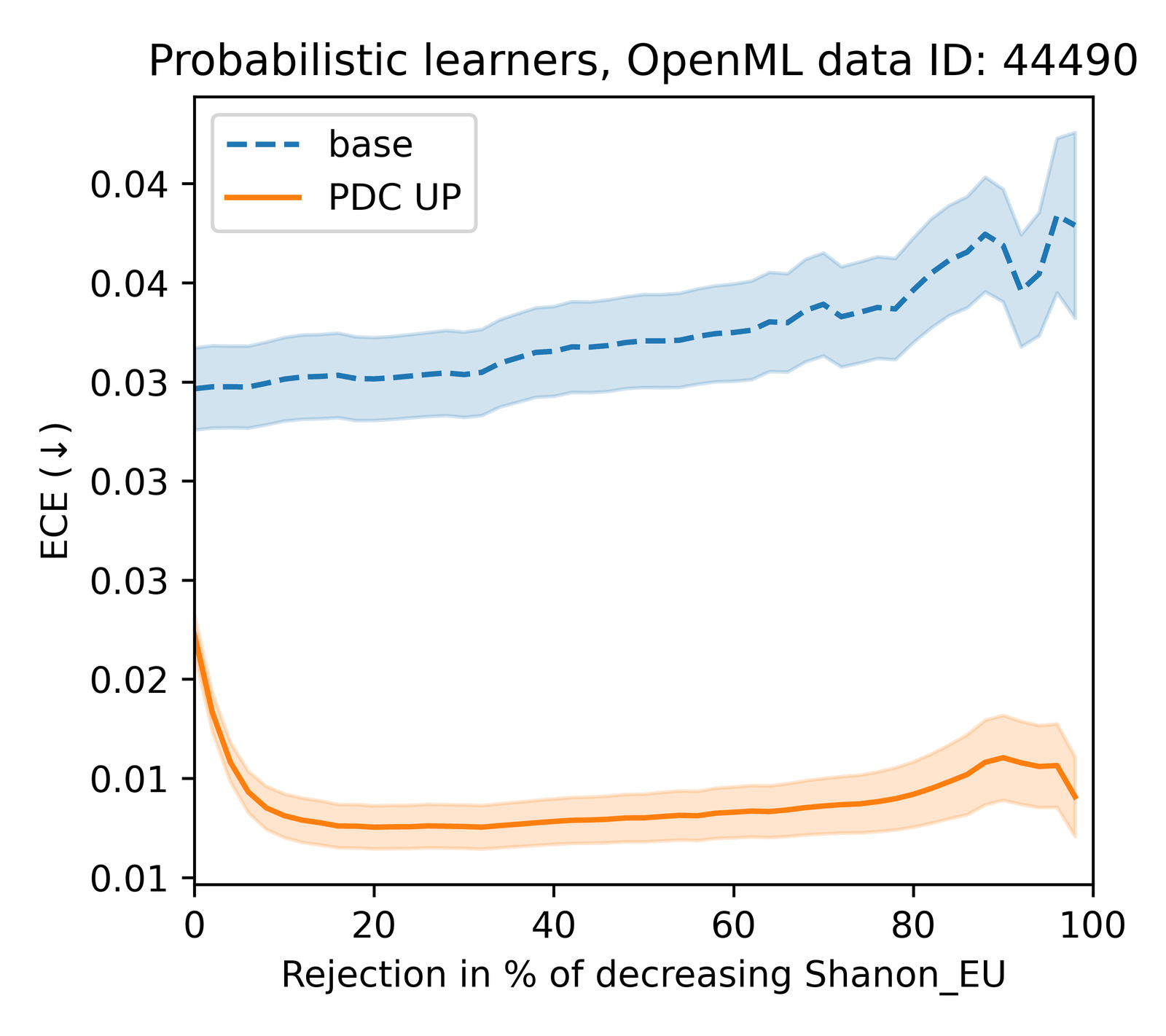}
    \includegraphics[width=0.4\linewidth]{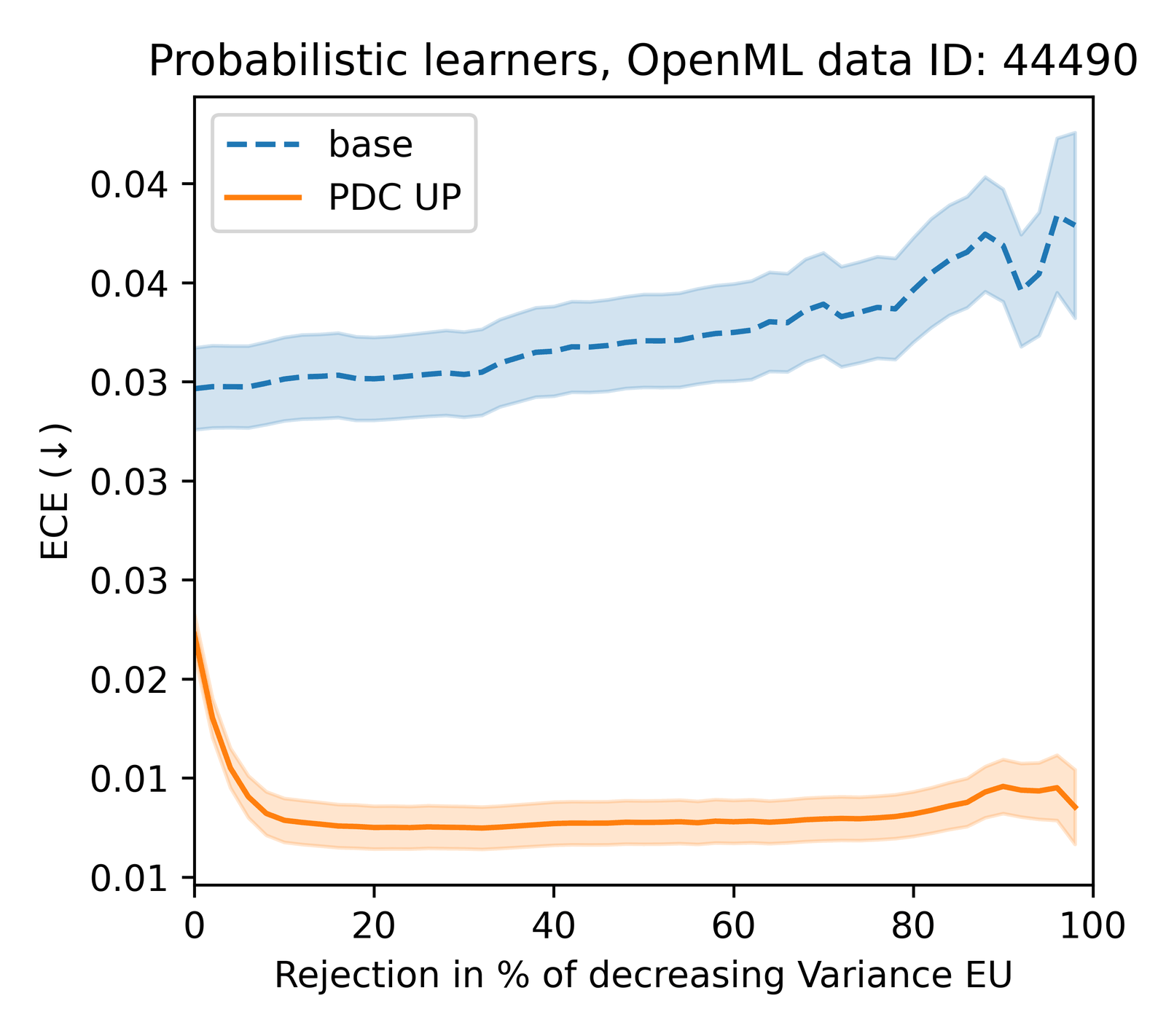}

    \includegraphics[width=0.4\linewidth]{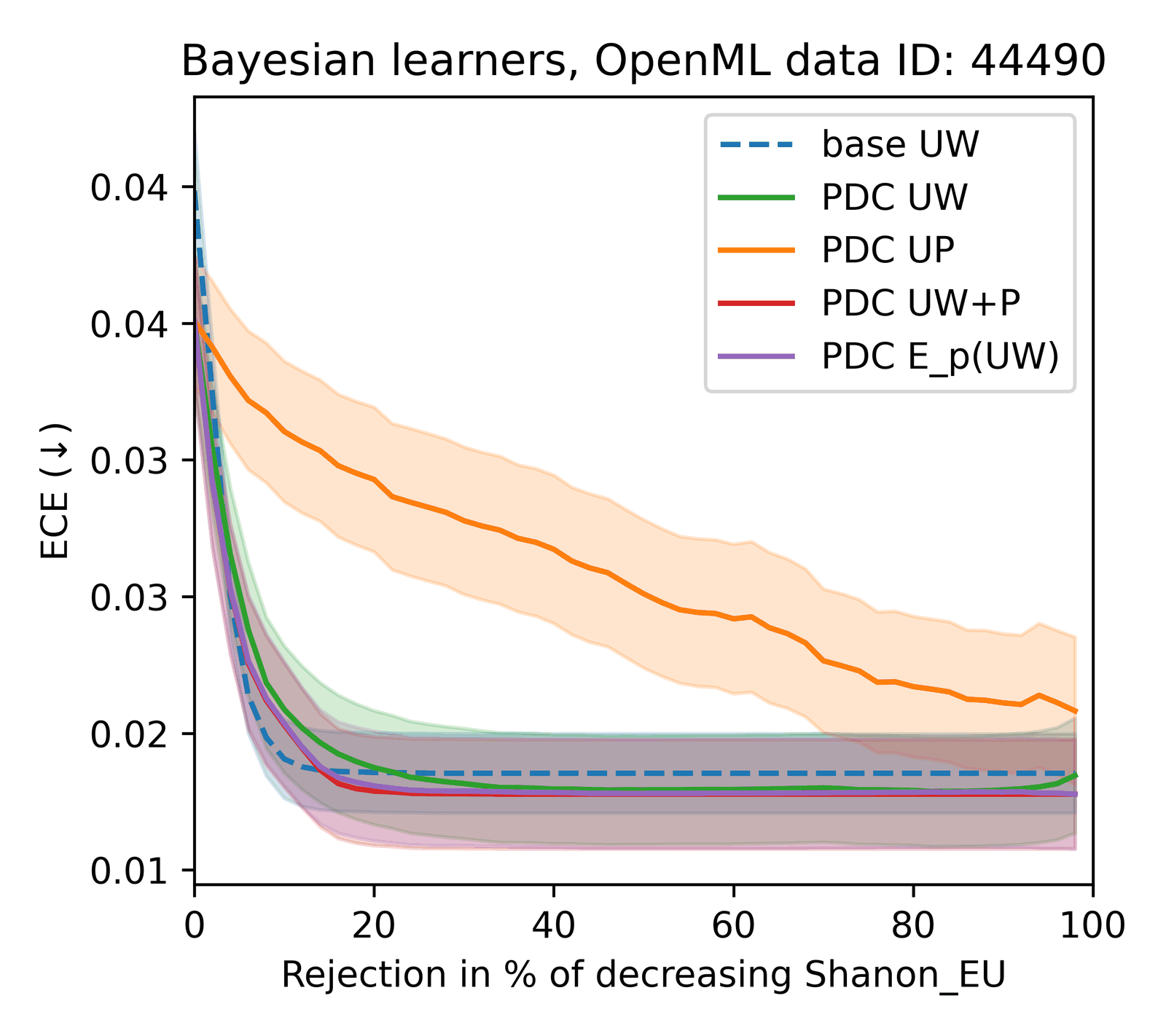}
    \includegraphics[width=0.4\linewidth]{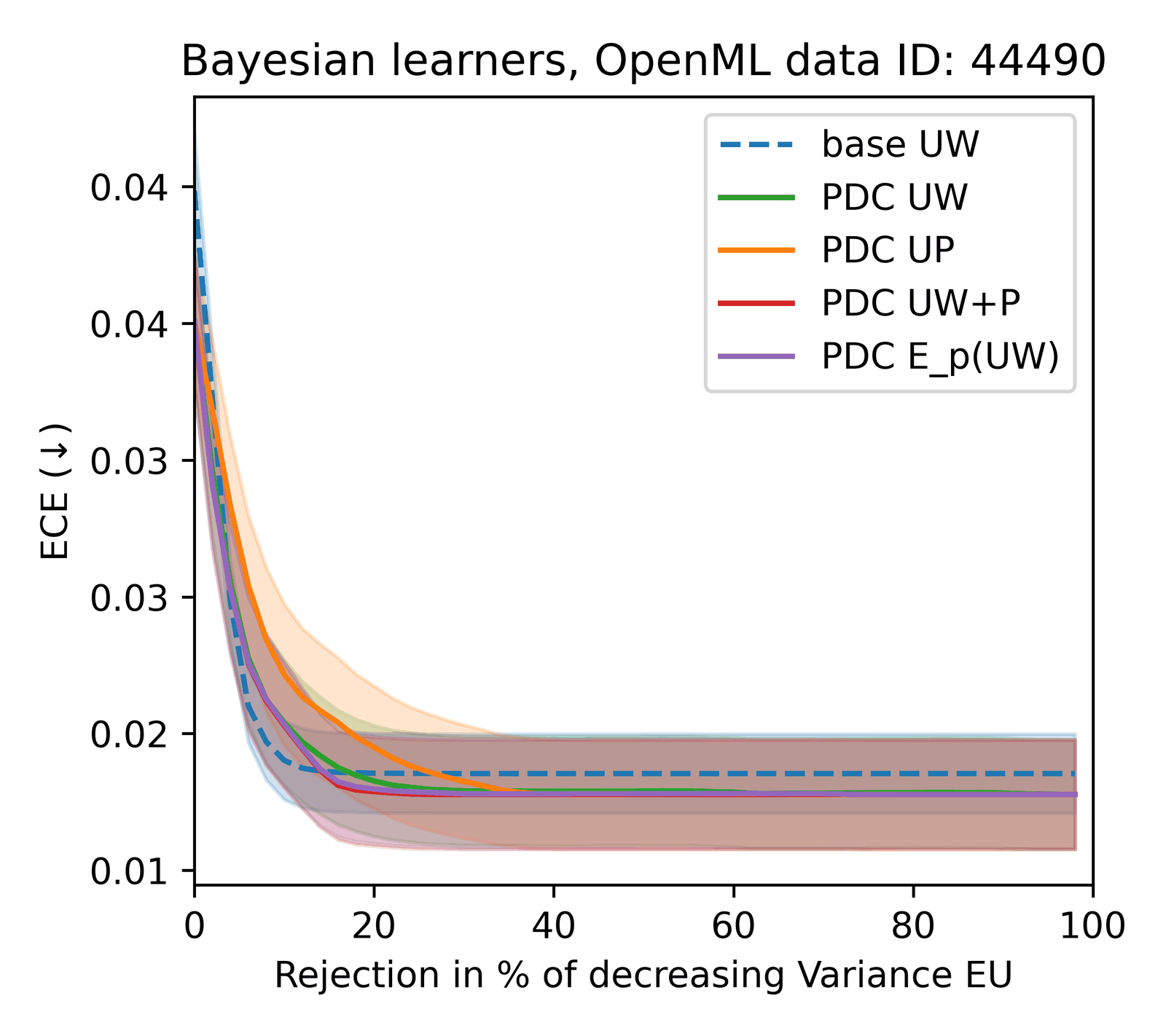}
    \caption{ECE rejection curve for the Dataset 44490.}
\end{figure}

\end{appendices}
\end{document}